\documentclass[10pt,twocolumn,letterpaper]{article}

\usepackage[pagenumbers]{cvpr}

\usepackage{placeins}
\usepackage{bbm}
\usepackage{longtable}
\usepackage{multirow}
\usepackage{fontawesome5}
\usepackage{colortbl}
\usepackage{pifont}
\usepackage{nicefrac}

\newcommand{\xmark}{\ding{55}}
\newcommand{\cmark}{\ding{51}}
\definecolor{plotgreen}{HTML}{006d2c}
\definecolor{plotorange}{HTML}{d94701}

\renewcommand{\paragraph}[1]{\noindent\textbf{#1}}

\makeatletter
\renewcommand\LenToUnit[1]{\strip@pt\dimexpr#1\relax}
\makeatother

\usepackage{bibunits}

\definecolor{cvprblue}{rgb}{0.21,0.49,0.74}
\usepackage[breaklinks,colorlinks,allcolors=cvprblue]{hyperref}

\usepackage{booktabs,amsmath}
\usepackage{etoc}

\title{Ego-METAS: an Egocentric online Multimodal Energy-efficient Temporal Action Segmentation benchmark}

\author{
  Maria Santos-Villafranca\textsuperscript{1} \qquad
  Alejandro Perez-Yus\textsuperscript{1} \\
  Jesus Bermudez-Cameo\textsuperscript{1} \qquad
  Giovanni Maria Farinella\textsuperscript{2} \qquad
  Antonino Furnari\textsuperscript{2} \\[0.6em]
  \makebox[0pt]{\textsuperscript{1}University of Zaragoza - I3A \hfill
  \textsuperscript{2}Department of Mathematics and Computer Science, University of Catania}
}

\begin{document}
\crefname{appendix}{Supp.}{Supp.}
\Crefname{appendix}{Supp.}{Supp.}

\maketitle
\begin{bibunit}
\begin{abstract}
To operate in the physical world, embodied agents must perceive their environment in an ``always-on'' fashion, reading sensors whose energy costs differ by orders of magnitude and that may fail or arrive late. Yet most prior work assumes unlimited energy and reliable inputs. We introduce \textbf{Ego-METAS}, the first \textbf{Ego}centric online \textbf{M}ultimodal \textbf{E}nergy-efficient \textbf{T}emporal \textbf{A}ction \textbf{S}egmentation benchmark: more than 100 hours of untrimmed egocentric video from Ego-Exo4D, CMU-MMAC, and CaptainCook4D with up to 5 synchronized modalities, released with splits, annotations, features, and an analytical energy model retargetable to different devices. We formulate an online temporal action segmentation task where predictions are made under a sensing schedule (which sensors are read, how often, and when) that must fit an energy budget. Across three representative architectures (a temporal convolutional network, a causal Transformer, and a recurrent network), we compare training regimes and sensing strategies. Efficiency comes from choosing which sensors to read and how often, rather than when: reading every sensor at 30\,FPS brings little or no gain with the same training recipe over low-rate schedules of complementary sensors, and periodic schedules selected once on validation data, match or outperform learned policies adapting schedules based on input within the same budget. Only a non-causal oracle gains from read timing, at low budgets. The backbone and its training also matter: a single recurrent model trained once on randomized schedules serves every budget and loses less accuracy when sensors are missing. Ego-METAS thus provides a common basis to measure which sensing decisions are worth learning, and develop systems which can adapt to the energy and sensors available at deployment.
\end{abstract}
\section{Introduction}
\label{sec:introduction}
\begin{figure}
    \centering
    \includegraphics[width=\linewidth]{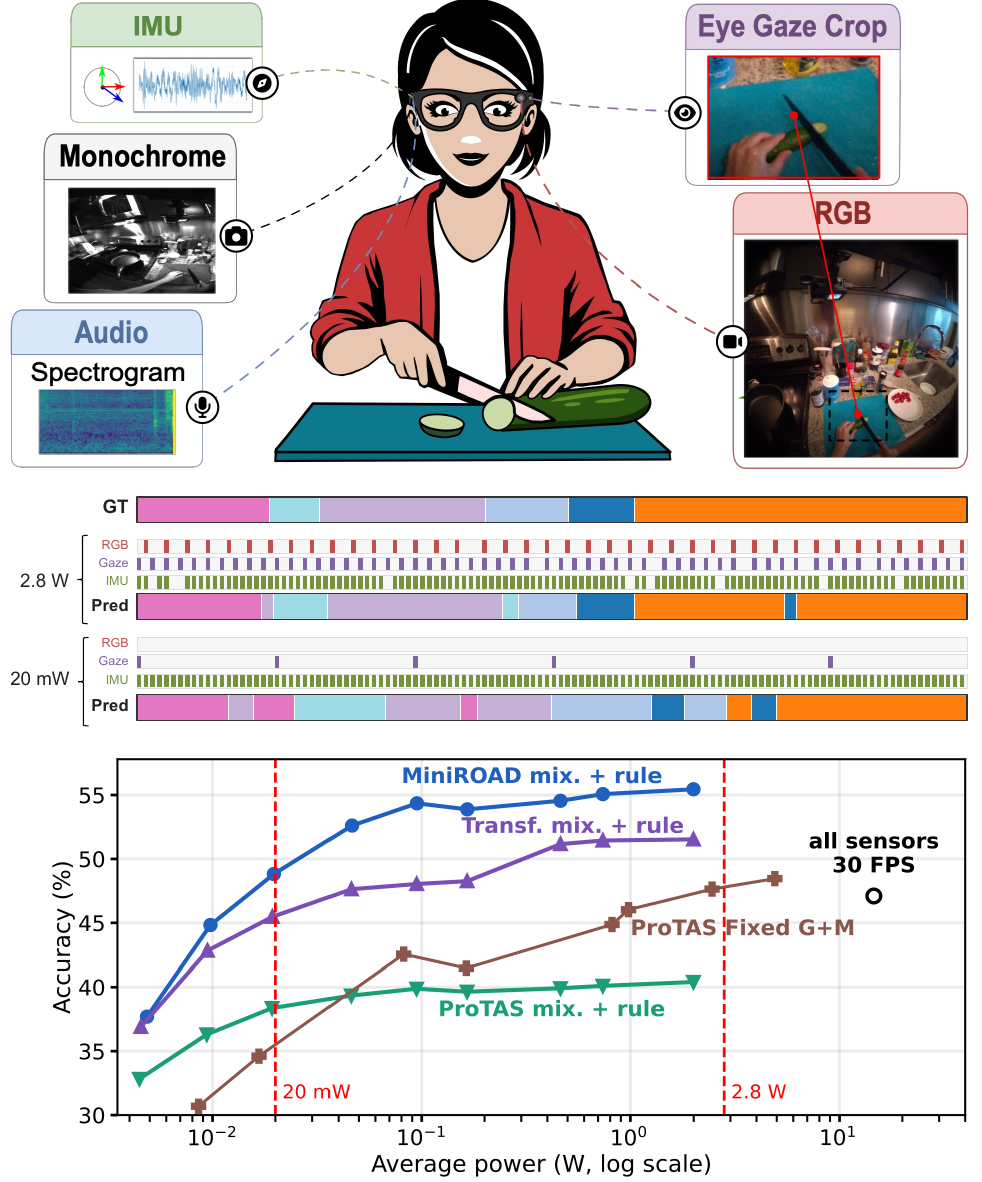}
    \caption{\textbf{The Ego-METAS Benchmark.} Top: wearable devices capture multiple sensor streams. Middle: illustrative sensor reads and predictions under two power budgets. Bottom: accuracy of different baselines against average power on Ego-Exo4D.}
    \label{fig:teaser}
    \vspace{-15pt}
\end{figure}

Continuous perception is essential for intelligent agents operating in the physical world, such as robots or assistants deployed through smart glasses~\cite{plizzari2024outlook}. Yet current vision systems process all modalities exhaustively, assuming that hardware scaling will make energy, computation, and memory abundant.
We argue that perception should budget energy explicitly. As illustrated in \cref{fig:teaser}, wearable devices are multimodal: an assistant can recognize actions online while choosing which sensors to read and how often, given their energy cost and utility.
Energy-aware processing has been studied only in isolated settings, partly because no standardized benchmark targets energy-aware online TAS. Most Temporal Action Segmentation (TAS) work focused on offline uni-modal settings~\cite{ding2024temporal} or exhaustive multimodal processing~\cite{romeo2025multimodal}. Online TAS has received less attention~\cite{shen2024progress,zhong2024onlinetas}, and the few energy-aware works~\cite{panda2021adamml,grauman2024ego,weng2024hcms} use different datasets and protocols, mostly on trimmed clips.

We introduce Ego-METAS: an Egocentric online Multimodal Energy-efficient Temporal Action Segmentation benchmark, sourcing more than 100 hours of untrimmed egocentric video from three datasets recorded with different wearable hardware: {Ego-Exo4D}~\cite{grauman2024ego}{ (Aria glasses)}, {CaptainCook4D}~\cite{peddi2024captaincook4d}{ (GoPro/HoloLens rig)}, and {CMU-MMAC}~\cite{de2009guide}{ (custom multimodal rig)}.
The task is online: a model predicts the action of each frame causally under a sensing schedule (which sensors are read, how often, and when) that must fit an energy budget. We curate up to five modalities (RGB, low-power monochrome images, gaze-centered crops, audio, and IMU) and estimate their acquisition, memory, and processing costs with an analytical energy model {whose hardware constants (sensor power, memory access, {compute}) can be replaced to describe different devices.} On this benchmark, we compare sensing strategies (Fixed schedules, heuristic strategies, and learned policies) on three TAS models: a temporal convolutional network (ProTAS~\cite{shen2024progress}), a causal Transformer, and a recurrent network (MiniROAD~\cite{An_2023_ICCV}). We train each model either as a specialist on one schedule or as a mixture on randomized schedules, so that one model serves different budgets.

We organize the analysis around four questions. (Q1) Which sensing strategies pay off? Low-rate schedules of complementary sensors match all sensors at 30\,FPS (48.0 vs. 47.1 Acc on Ego-Exo4D) at a fraction of the power, while dynamic heuristics bring no systematic gain over Fixed schedules (\cref{tab:all_protas}). (Q2) How much do the TAS model and its training matter? Both matter: the recurrent network {mostly} outperforms the temporal convolutional network, and a model trained on randomized schedules stays close to the specialist (\cref{tab:backbone_training}). (Q3) What does randomized training buy? Robustness to missing sensors (\cref{tab:mismatch}). (Q4) Which sensing decisions are worth learning? Learned policies are no better than periodic reads with the same {reads}, and only a non-causal oracle gains from read timing, at low budgets{, which a causal model imitating it does not confirm} (\cref{tab:learn_where}). These conclusions also hold under the edge-SoC energy model (\cref{tab:edge_small}). Efficiency thus comes from which sensors are read and how often, rather than when.

In summary, we contribute (i) Ego-METAS, with unified splits, cleaned annotations, pre-extracted features, an energy model, and evaluation scripts, to be released upon acceptance; (ii) an analysis of baselines, training regimes, and sensing schedules; and (iii) a practical recipe (randomized training plus a budget rule selecting periodic schedules on validation videos) {with MiniROAD that is} {at most 2.5 Acc points below specialists at 2.8\,W} and better at 20\,mW.

\section{Related Work}
\label{sec:related_work}

\paragraph{Energy-efficient methods:} Deploying vision systems on low-power devices \cite{chen2024survey} typically relies on static network optimizations \cite{goel2020survey}, such as quantization \cite{wang2018training}, pruning \cite{anwar2017structured}, compact architectures \cite{iandola2016squeezenet}, linear-time sequence models \cite{gu2023mamba,beck2024xlstm}, and knowledge distillation \cite{hinton2015distilling}, which EgoDistill \cite{tan2023egodistill} applies to egocentric video and IMU.
These methods reduce the cost of a given model. We study instead which sensors to read, and how often, within an energy budget.

\paragraph{Adaptive frame and modality selection:} For video recognition, dynamic networks~\cite{han2021dynamic} replace uniform sampling~\cite{xu2024pllava} {by selecting} salient clips~\cite{korbar2019scsampler} or frames~\cite{hu2025m}, exiting early~\cite{ghodrati2021frameexit}, or previewing audio before reading frames~\cite{gao2020listen}. Multimodal methods also gate modalities: AdaMML~\cite{panda2021adamml} learns one gate per modality with the Gumbel-Softmax trick~\cite{jang2016categorical}, and HCMS~\cite{weng2024hcms} activates costlier modalities in a cascade conditioned on cheaper ones. Similar gates select features~\cite{su2016leaving}, fusion paths~\cite{xue2023dynamic}, or sensors~\cite{yang2022efficient,cui20252,chowdhury2025egoadapt}. However, most classify trimmed clips and few target an energy budget. Within this family, we adapt AdaMML and HCMS to work on untrimmed streams, to evaluate which sensing decisions are worth learning.

\paragraph{Robustness to missing modalities:} Combining sensors typically improves egocentric recognition~\cite{pramanick2023egovlpv2,zhang2024masked,das2025primus,romeo2025multimodal}, but multiplies sensing and processing costs, while sensors may be unavailable at test time. Prior work handles missing modalities by distillation~\cite{santoscarrion2026karmma} or test-time adaptation~\cite{ramazanova2024test}, and select modalities under domain shift~\cite{marinov2022modselect}. ModDrop~\cite{neverova2016moddrop} drops modalities at random during training, so that one model accepts any subset of inputs, and once-for-all networks~\cite{cai2020once} train one network for many hardware targets. Our mixture training follows the same principle with random sensor rates and phases.

\paragraph{Temporal action segmentation (TAS):} Offline TAS assumes access to the full video for frame-wise classification~\cite{ding2024temporal,lu2024fact,farha2019ms,singhania2023c2f}, while wearable assistants require online TAS, which predicts each frame causally~\cite{shen2024progress,zhong2024onlinetas,ghoddoosian2022weakly}. Online TAS is less explored~\cite{ding2024temporal} and builds on online action detection~\cite{de2016online} with recurrent~\cite{An_2023_ICCV}, Transformer~\cite{xu2021lstr,zhao2022real}, or memory-based models~\cite{zhong2024onlinetas}, which typically read every input at every step. The energy-efficient benchmark of Ego-Exo4D \cite{grauman2024ego} adds an energy budget to online keystep recognition on streaming episodes of arbitrary length, scored with per-step mcAP (see \cref{supp:related_datasets} for a detailed comparison with related datasets). Ego-METAS instead evaluates online TAS on untrimmed videos with long background intervals, from three datasets and devices.

\section{Benchmark}
\label{sec:benchmark}

\subsection{Task definition}
\label{subsec:task_definition}

Let $m\in\{1,\ldots,M\}$ index the available sensors (e.g., RGB, audio, IMU) and $t\in\{1,\ldots,T\}$ the discrete prediction steps. Ego-METAS benchmarks online Temporal Action Segmentation (TAS) from multimodal sensor streams under a predefined power budget $B$. The objective is to maximize online segmentation performance while accounting for sensing, computation, and memory costs.
Sensor usage is represented by a \textbf{sensing schedule}
$\sigma \subseteq \{1,\ldots,M\}\times\{1,\ldots,T\}$, where $(m,t)\in\sigma$ indicates that sensor $m$ is read at step $t$. The schedule therefore determines which sensors are read, how often, and when. A schedule may be periodic, stochastic, heuristic, or generated by a learned policy, in which case decisions can only depend on information available up to the current step.
For every sensing event $(m,t)\in\sigma$, observation $\mathbf{X}^{(m)}_t$ is acquired and processed by a feature extractor $\phi_m$ to obtain a feature vector $\mathbf{x}^{(m)}_t=\phi_m(\mathbf{X}^{(m)}_t)$. A causal TAS model $\Psi$ uses the observations available up to step $t$ to predict the current action class $\hat{y}_t$. 
In practice, every TAS model we benchmark (\cref{sec:exp_setup}) fills in unread steps with a \emph{zero-order hold}: at a step $t$ with $(m,t)\notin\sigma$, it reuses the last feature read, $\mathbf{x}^{(m)}_t=\mathbf{x}^{(m)}_{t'}$ with $t'=\max\{t''\le t:(m,t'')\in\sigma\}$, or $\mathbf{0}$ if $m$ has not been read yet.

\begin{figure}
    \centering
\includegraphics[width=1\linewidth]{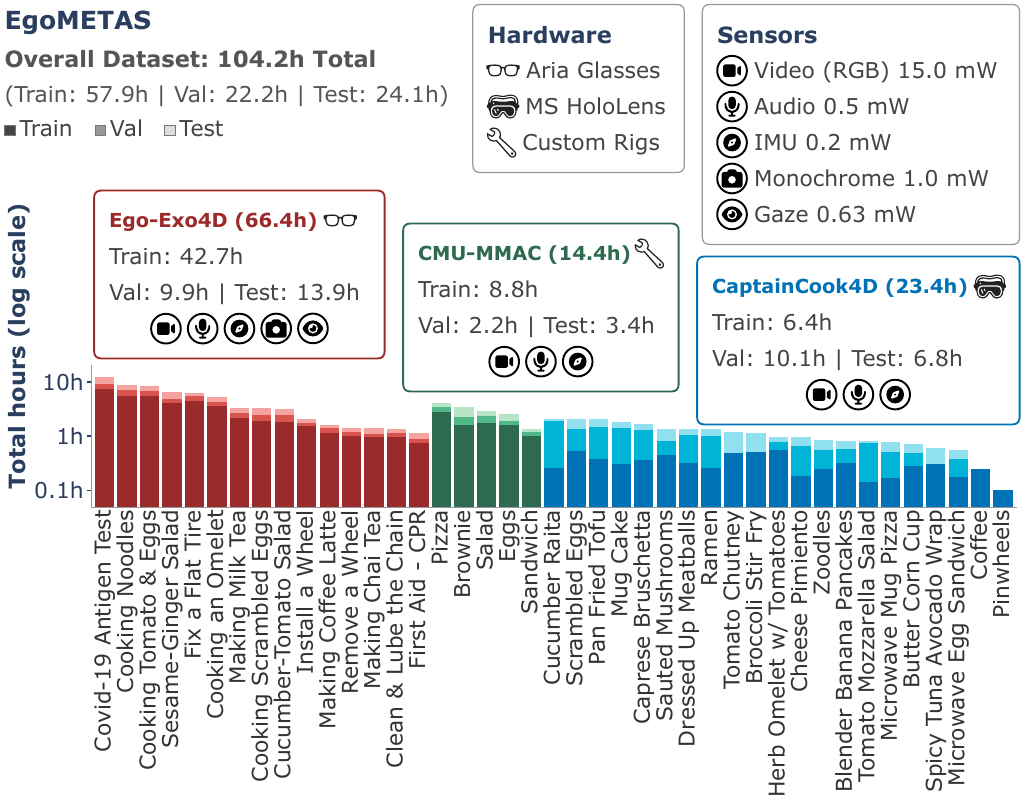}
    \caption{{\textbf{Ego-METAS} comprises 104.2 hours of video and 41 scenarios collected with three wearable devices, encompassing 5 modalities captured with varying energy consumption profiles.}}
    \label{fig:egometas}
\end{figure}

\subsection{Datasets, Modalities and Preparation}
\label{subsec:datasets}

Ego-METAS comprises 104.2 hours of egocentric video across 41 scenarios from three multimodal datasets with diverse sensing platforms and activity domains. The benchmark includes up to five sensor modalities, with train, validation, and test splits (Fig.~\ref{fig:egometas}):

\begin{itemize}
    \item \textbf{Ego-Exo4D~\cite{grauman2024ego}.} 
    This dataset was collected with Aria glasses, and includes RGB, audio, IMU, monochrome video, and gaze. We select 15 scenarios, spanning about 66.4 hours of diverse procedural activities. It is the largest and most diverse subset of the benchmark and the only dataset including activities beyond the cooking domain.
    \item \textbf{CMU-MMAC \cite{de2009guide}.} Recorded using custom multimodal rigs, it captures subjects performing kitchen tasks in a lab setting. We use RGB video, audio, and accelerometer and gyroscope measurements from the IMUs, covering five recipes and 14.4 hours of recordings.
    \item \textbf{CaptainCook4D \cite{peddi2024captaincook4d}.} Recorded using a GoPro and Microsoft HoloLens, this dataset focuses on step-by-step procedural execution and mistake detection. We adopt RGB video, audio, and IMU streams across 21 recipes, for a total of {23.4} hours.
\end{itemize}

\paragraph{Benchmark preparation.}
Integrating the three datasets required substantial processing to obtain a consistent online setting. We revised annotations {to resolve} differences in label granularity, duplicated or overlapping classes, and other inconsistencies that made the original annotations unsuitable for our task. The sensor streams were temporally aligned, including cross-device synchronization where required, and mapped to a common 30 FPS reference timeline following the video stream. 
Dataset selection, comparison with related benchmarks, re-annotation, synchronization, data splits and  statistics are described  in \crefrange{supp:source_datasets}{supp:dataset_stats}. Label mappings and statistics are in \cref{sec:supp_complete_tables}.

\paragraph{Pre-extracted features.}
In addition to the synchronized sensor streams, we will release pre-extracted features for all modalities using frozen feature extractors $\phi_m$ from the literature. A feature vector $\mathbf{x}_t^{(m)}$ is pre-extracted for each modality $m$ at every timestep $t$ (\cref{supp:mod_computation}).

\subsection{Energy computation}
\label{subsec:energy_computation}

We model the energy consumption of a model
as the sum of sensing, computation, and memory costs under a sensing schedule $\sigma$. If the schedule is generated online by a policy $\pi$, we also account for its processing cost.
Following prior work~\cite{grauman2024ego}, we characterize processing cost through multiply-accumulate operations (MACs) and bytes allocated by CUDA operations (PyTorch profiler). For $q\in\{\mathrm{MAC},\mathrm{MEM}\}$, let $f_{\Psi,t}^{q}$, $f_{\pi,t}^{q}$, and $f_{\phi_m}^{q}$ denote the corresponding quantities for the TAS model, online policy, and feature extractor $\phi_m$, respectively. If no online policy is utilized, $f_{\pi,t}^{q}=0$. We then define the total processing requirement at step $t$ as

\begin{equation}
F_t^{q}
=
f_{\Psi,t}^{q}
+
f_{\pi,t}^{q}
+
\sum_{m:(m,t)\in\sigma} f_{\phi_m}^{q}.
\end{equation}
The total energy consumption is then
\begin{equation}
\label{eq:energy_total}
E_{\mathcal{V}}
=
\sum_{t=1}^{T}
\left(
\Delta\tau \sum_{m:(m,t)\in\sigma} P_m
+
\alpha F_t^{\mathrm{MAC}}
+
\beta F_t^{\mathrm{MEM}}
\right).
\end{equation}
Here, $P_m$ is the capture power of sensor $m$ (mW), and $\Delta\tau$ is the duration of one step of our reference timeline ($\Delta\tau=1/30$\,s), i.e., the capture time charged for each read. The constants $\alpha$ and $\beta$ convert MAC operations and memory traffic into energy. 
Profiling only counts operations and memory traffic, whereas hardware assumptions enter through $\alpha$, $\beta$, and $P_m$, so the constants can be changed without re-profiling or re-training.
We adopt the reference constants of the Ego-Exo4D energy benchmark~\cite{grauman2024ego} ($\alpha=4.6$\,pJ/MAC, $\beta=80$\,pJ/byte), which correspond to FP32 arithmetic in 45\,nm technology {\cite{horowitz2014computing}} and to the energy per byte of low-power mobile DRAM traffic.
\cref{tab:edge_small} repeats the main comparisons under the edge-SoC energy model (constants of a recent wearable SoC: 0.4\,pJ/MAC, 60\,pJ/byte), including a $5$\,mW regime.
Since $T\Delta\tau$ is the duration of the multimodal stream in seconds, we compute its average power as $
P_{\mathcal{V}}
=
E_{\mathcal{V}}/(T\Delta\tau)$,
expressed in mW when $E_{\mathcal{V}}$ is expressed in mJ.
A configuration operates within a power budget $B$ when $P_{\mathcal{V}} \le B$. {See \cref{supp:energy_scope,supp:energy_budget,supp:energy_grounding} for scope and limitations, capture powers and profiling, and constants.}

\subsection{Evaluation protocol and metrics}
\label{subsec:evaluation_protocol}

\paragraph{Temporal predictions.}
Models are evaluated on the 30 FPS reference timeline and must produce a prediction at every step. When models operate at lower frequencies, we replicate their predictions causally to the reference framerate.

\paragraph{Operating budgets.}
Following the energy setting of Ego-Exo4D~\cite{grauman2024ego}, we evaluate two regimes: $20$\,mW and $2.8$\,W. These budgets mimic wearables from smart glasses (Ray-Ban Meta~\cite{metarayban}) to headsets (Apple Vision Pro~\cite{applevisionpro}).

\paragraph{Evaluation metrics.}
Following standard TAS literature~\cite{ding2024temporal}, we use Frame Accuracy (Acc), segmental Edit Distance (Edit), and segmental F1 at a temporal Intersection-over-Union threshold of 25\% (F1@25, F1 in the tables). We report them averaged over scenarios, together with the average power $P$. Edit and F1 ignore the background class, whereas Acc counts every frame. Episodic keystep benchmarks such as Ego-Exo4D~\cite{grauman2024ego} use ranking-based metrics such as mean Average Precision (mcAP), unsuited to online TAS (see \mbox{\cref{supp:benchmark_inconsistencies}}).

\section{Models, Training and Sensing Strategies}
\label{sec:baselines}

We evaluate different baseline multimodal energy-efficient schemes. Each of them is defined by three components: the TAS model itself (\cref{sec:models}), the sensing schedules it is trained on, and the sensing strategy that decides which sensors are read at test time (\cref{sec:sens_sched,sec:learned_pol}).

\subsection{TAS models and experimental setup}
\label{sec:models}
\label{sec:exp_setup}
We benchmark three TAS models: ProTAS~\cite{shen2024progress}, a temporal convolutional network; a causal Transformer\footnote{{We use a vanilla Transformer rather than optimized models such as TeSTra}~\cite{zhao2022real}{, which are unsuited to low budgets (see \cref{supp:add_testra}).}}; and MiniROAD~\cite{An_2023_ICCV}, a recurrent network.
Each model is trained either on one fixed sensing schedule (\emph{specialist}) or on a mixture of randomized schedules (\emph{mixture}). In the mixture, every time a training video is fed to the model it is paired with a new randomly sampled schedule, where each sensor is periodic with its own rate (up to 6\,FPS) and phase, over a random subset of sensors, so that one model accepts any schedule at test time (details in \cref{sec:supp_schedules}). {We cap rates at 6\,FPS since training at 30 steps per second is costlier.} To meet a budget, we then select one schedule per model, dataset, and budget maximizing performance on validation videos (\emph{budget rule}, \cref{supp:budget_rule}). We train a separate model for each of the 15 scenarios of Ego-Exo4D~\cite{grauman2024ego} and the 5 scenarios of CMU-MMAC~\cite{de2009guide}. Since CaptainCook4D~\cite{peddi2024captaincook4d} is smaller, we treat all its recipes as a single scenario and train one model on it. Optimizer, epochs, batch size per dataset, and the checkpoint rule are in \cref{subsec:supp_settings}.

\subsection{Heuristic sensing strategies}
\label{sec:sens_sched}
\label{sec:fixed_sched}

\begin{figure}
    \centering
    \includegraphics[width=1\linewidth]{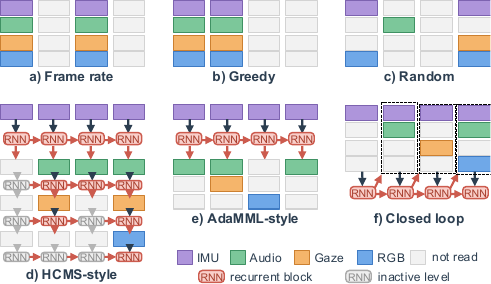}
    \caption{\textbf{Sensing schedules}. a) Fixed-stride reads selected sensors simultaneously at regular intervals.
    b) Greedy uses all modalities until the budget is consumed; c) Random drops individual modalities with a given probability; {d--f) Learned policies use a recurrent network to decide which sensors to read at each step.}
    }
    \label{fig:policies}
\end{figure}

We consider three heuristic sensing strategies inspired by the fixed-stride, greedy, and random baselines of Ego-Exo4D~\cite{grauman2024ego} (\cref{fig:policies}(a)--(c)). These strategies use predefined or stochastic rules, independently of the observed data.

\paragraph{Fixed-stride.}
All selected sensors are read simultaneously at a common fixed sampling rate (\cref{fig:policies}(a)).
Lower rates reduce energy but lose temporal resolution, and short actions may be missed. We evaluate different sensor subsets and sampling rates, denoted \emph{Fixed} in the tables.

\paragraph{Greedy.}
All available sensors are read until the energy budget for the current second is exhausted (\cref{fig:policies}(b)). {The process restarts every second, which avoids bursts followed by long blind periods. At 20\,mW one read exceeds the 1\,s budget, so we use 30\,s windows (\cref{supp:greedy_policy}).}

\paragraph{Random.}
Sensors are randomly dropped at each timestep (\cref{fig:policies}(c)). The TAS model is trained with a dropout rate $\tau_{\mathrm{train}}$, while $\tau_{\mathrm{inf}}$ controls dropout at inference to meet different energy budgets. We also propose a \textit{cost-aware} variant (Random-CA), favoring low-power modalities (\cref{supp:random_policy}).

\subsection{Learned sensing policies}
\label{sec:learned_pol}
We consider two learned policies derived from the literature and a closed-loop policy proposed by us (\cref{fig:policies}(d)--(f)).

\paragraph{AdaMML-style}~\cite{panda2021adamml} (\cref{fig:policies}(e)): at each step, a recurrent network (RNN) reads a cheap always-on input, the IMU, and decides which of the other sensors (RGB, audio, gaze, and monochrome camera) to read at that same step, one binary gate per sensor. The policy does not see what the gated sensors would have returned.

\paragraph{HCMS-style}~\cite{weng2024hcms} (\cref{fig:policies}(d)): sensors are sorted by cost and the cheapest (IMU) is always on. A cascade of gates, one per cost level, decides whether to activate the next more expensive sensor given what cheaper levels have read, so a sensor is read only if cheaper ones are. Each level has its RNN over the features of its sensor and the cheaper ones.

\paragraph{Closed loop} (\cref{fig:policies}(f)), proposed by us: an RNN reads what has actually been sensed so far (the features of the sensors read and its own previous decisions) and decides which sensors to read at the next step, while the IMU is always read. Unlike the two policies above, its decision depends on the content read at $t$, not on a cheap proxy at $t+1$.

\paragraph{Training Details}
Policies are trained to maximize accuracy on top of frozen models trained on mixtures of schedules. Since previous work targeted trimmed segment classification~\cite{panda2021adamml,weng2024hcms}, rather than budget-constrained streaming, these methods cannot tune a policy to a given budget. We tackle this by adding a Lagrangian term to enforce a target budget, with a multiplier updated by dual ascent during training~\cite{nandwani2019primal}. All three policies make binary read decisions, which we train with the straight-through Gumbel-softmax estimator~\cite{jang2016categorical}: each gate gives a hard 0/1 read in the forward pass, while gradients flow through its continuous relaxation. We also experimented with training the policies end-to-end with the TAS model, without Lagrangian optimization, adapting policies to the intended budget by adjusting a threshold over gate predictions. {Policy inputs,  Lagrangian update, and budget calibration are in \cref{supp:learned_pol}.}

\begin{table*}[t]
\caption{{\textbf{Heuristic sensing strategies with ProTAS.} Low-rate multimodal schedules match or exceed 30\,FPS sensing. {Dynamic sensing matches Fixed in Acc only at 2.8\,W (Random-CA on Ego-Exo4D, CaptainCook4D).} For random policies, we report ($\tau_{\mathrm{train}}$/$\tau_{\mathrm{inf}}$).}}
\label{tab:all_protas}
\centering
\scriptsize
\setlength{\tabcolsep}{3.4pt}
\begin{tabular}[t]{@{}l rrr r@{}}
\toprule
\multicolumn{5}{c}{\textbf{Ego-Exo4D}} \\
\midrule
Policy & Acc & Edit & F1 & P \\
\midrule
\multicolumn{5}{@{}l}{\textit{Budget: None} (P in W)} \\
Fixed (30\,FPS R) & 39.9 & \textbf{24.2} & 19.7 & 9.30 \\
Fixed (30\,FPS all) & 47.1 & 23.1 & \textbf{20.4} & 14.6 \\
Fixed (30\,FPS G+M) & \textbf{48.5} & 21.2 & 19.3 & 4.90 \\
\midrule
\multicolumn{5}{@{}l}{\textit{Budget: 2.8\,W} (P in W)} \\
Fixed (6\,FPS R) & 40.6 & 27.1 & 21.4 & 1.86 \\
Fixed (6\,FPS G+M) & 46.0 & 26.5 & 23.8 & 0.98 \\
Fixed (6\,FPS R+A+G) & 48.0 & \textbf{29.2} & \textbf{25.9} & 2.31 \\
\cmidrule(lr){1-5}
Greedy & 47.9 & 28.4 & 23.0 & 2.44 \\
Random (0.9/0.9) & 47.8 & 27.7 & 22.3 & 1.47 \\
Random-CA (0.7/0.7) & \textbf{48.1} & 27.4 & 22.9 & 2.56 \\
\midrule
\multicolumn{5}{@{}l}{\textit{Budget: 20\,mW} (P in mW)} \\
Fixed (0.05\,FPS R) & 24.5 & 14.4 & 10.4 & 16.3 \\
Fixed (0.1\,FPS G+M) & \textbf{34.6} & 22.5 & \textbf{18.1} & 16.7 \\
Fixed (1\,FPS A+I) & 28.0 & 20.0 & 14.1 & 14.0 \\
\cmidrule(lr){1-5}
{Greedy} & 30.7 & 20.5 & 13.5 & 16.1 \\
Random (0.99/0.9987) & 33.4 & \textbf{26.0} & 13.9 & 18.7 \\
Random-CA (0.99/0.9978) & 34.0 & 25.8 & 14.6 & 18.2 \\
\midrule
Blind (no sensors) & 22.3 & 10.4 & 8.4 & 0.4\,mW \\
\bottomrule
\end{tabular}
\hfill
\begin{tabular}[t]{@{}l rrr r@{}}
\toprule
\multicolumn{5}{c}{\textbf{CMU-MMAC}} \\
\midrule
Policy & Acc & Edit & F1 & P \\
\midrule
\multicolumn{5}{@{}l}{\textit{Budget: None} (P in W)} \\
Fixed (30\,FPS R) & 85.0 & \textbf{31.7} & \textbf{36.4} & 9.30 \\
Fixed (30\,FPS all) & 84.0 & 29.6 & 31.8 & 9.72 \\
Fixed (30\,FPS R+A) & \textbf{85.6} & 30.4 & 32.8 & 9.72 \\
\midrule
\multicolumn{5}{@{}l}{\textit{Budget: 2.8\,W} (P in W)} \\
Fixed (6\,FPS R) & \textbf{84.3} & 39.1 & 40.7 & {1.86} \\
Fixed (0.5\,FPS R) & 81.6 & 44.1 & \textbf{44.0} & 0.16 \\
Fixed (1\,FPS R+A) & 83.2 & \textbf{44.8} & 43.3 & 0.32 \\
\cmidrule(lr){1-5}
Greedy & 80.6 & 27.1 & 26.8 & 2.60 \\
Random (0.9/0.9) & 81.0 & 26.0 & 27.4 & 0.98 \\
Random-CA (0.7/0.7) & 80.8 & 26.8 & 26.3 & 1.27 \\
\midrule
\multicolumn{5}{@{}l}{\textit{Budget: 20\,mW} (P in mW)} \\
Fixed (0.05\,FPS R) & 69.4 & 24.1 & 23.8 & 16.0 \\
Fixed (0.05\,FPS {R+A+I}) & \textbf{70.8} & 24.2 & \textbf{26.8} & 16.8 \\
Fixed (1\,FPS A+I) & 67.8 & 26.5 & 20.3 & 14.0 \\
\cmidrule(lr){1-5}
{Greedy} & 66.0 & 19.9 & 14.4 & 11.0 \\
Random (0.97/0.998) & 68.0 & \textbf{30.7} & 20.7 & 17.6 \\
Random-CA (0.99/0.997) & 70.6 & 21.0 & 19.5 & 14.3 \\
\midrule
Blind (no sensors) & 66.3 & 0.0 & 0.0 & 0.4\,mW \\
\bottomrule
\end{tabular}
\hfill
\begin{tabular}[t]{@{}l rrr r@{}}
\toprule
\multicolumn{5}{c}{\textbf{CaptainCook4D}} \\
\midrule
Policy & Acc & Edit & F1 & P \\
\midrule
\multicolumn{5}{@{}l}{\textit{Budget: None} (P in W)} \\
Fixed (30\,FPS R) & 35.6 & \textbf{18.0} & 14.7 & 9.1 \\
Fixed (30\,FPS all) & \textbf{35.9} & 16.9 & \textbf{15.5} & 9.53 \\
Fixed (30\,FPS R+A) & 34.4 & 11.7 & 8.5 & 9.52 \\
\midrule
\multicolumn{5}{@{}l}{\textit{Budget: 2.8\,W} (P in W)} \\
{Fixed (6\,FPS R)} & {42.3} & {26.4} & {18.3} & {1.82} \\
Fixed (1\,FPS R) & 42.6 & 36.4 & \textbf{28.9} & 0.30 \\
Fixed (0.5\,FPS R) & 43.5 & \textbf{38.2} & 23.1 & 0.15 \\
\cmidrule(lr){1-5}
Greedy & 33.8 & 21.0 & 18.2 & 2.54 \\
Random (0.7/0.9) & 31.3 & 18.8 & 15.3 & 0.95 \\
Random-CA (0.97/0.9) & \textbf{44.0} & 26.3 & 20.1 & 0.42 \\
\midrule
\multicolumn{5}{@{}l}{\textit{Budget: 20\,mW} (P in mW)} \\
Fixed (0.05\,FPS R) & 38.3 & 26.1 & 17.8 & 15.3 \\
Fixed (0.05\,FPS R+A) & \textbf{40.3} & 31.9 & 17.7 & 16.0 \\
Fixed (1\,FPS A+I) & 38.9 & 20.0 & 13.4 & 13.7 \\
\cmidrule(lr){1-5}
{Greedy} & 35.2 & 37.2 & 24.9 & 10.9 \\
Random (0.99/0.9985) & 28.6 & \textbf{38.3} & 20.5 & 14.9 \\
Random-CA (0.99/0.9958) & 28.0 & 32.8 & \textbf{20.9} & 17.0 \\
\midrule
Blind (no sensors) & 23.3 & 9.7 & 1.2 & 0.35\,mW \\
\bottomrule
\end{tabular}

\vspace{2pt}
{\footnotesize
\textbf{Sensors}: R = RGB, A = Audio, I = IMU, G = Gaze, M = Monochrome, all = all sensors. P: average power. \textbf{Bold}: best within the budget. 

}
\end{table*}

\section{Experiments}
\label{sec:experiments}
Unless stated otherwise, results use the reference energy model and the budgets of 20\,mW and 2.8\,W.

\subsection{{Heuristic Sensing strategies on Ego-METAS}}
\label{sec:exp_strategies}
Table~\ref{tab:all_protas} {compares heuristic sensing strategies with ProTAS on the three datasets. Configurations at 30\,FPS (Budget: None) are a reference costing several watts.}

\paragraph{Low-rate schedules match dense sensing.} On Ego-Exo4D, {RGB at 6\,FPS is above RGB at 30\,FPS (40.6 vs.\ 39.9 Acc) at one fifth of the power (1.86 vs.\ 9.30\,W).} On CMU-MMAC, reducing RGB from 30 to 6\,FPS costs 0.7 Acc points (84.3 vs.\ 85.0) while power drops from 9.30 to {1.86}\,W. On CaptainCook4D, RGB at 0.5\,FPS is above its 30\,FPS counterpart (43.5 vs.\ 35.6), partly because lower frame rates smooth the predictions (higher Edit and F1).

\paragraph{Multimodal sensing is key.} On Ego-Exo4D, lowering the rate of RGB alone is insufficient (40.6 Acc at 6\,FPS), while adding audio and gaze at the same rate reaches 48.0{, within 0.5 points of Gaze+Monochrome at 30\,FPS (48.5) at less than half its power (2.31 vs.\ 4.90\,W)}. At 20\,mW, Gaze+Monochrome at 0.1\,FPS reaches 34.6 Acc vs.\ 24.5 for RGB at 0.05\,FPS. The best 20\,mW schedule is multimodal on the other datasets too: 4.5 and 17.0 Acc points above blind on CMU-MMAC and CaptainCook4D.

\paragraph{{Dynamic heuristics do not improve on Fixed schedules.}} At 2.8\,W, Random-CA matches the best Fixed schedule on Ego-Exo4D (48.1 vs.\ 48.0 Acc, at 2.56 vs.\ 2.31\,W) and CaptainCook4D (44.0 vs.\ 43.5), but is 3.5 points below it on CMU-MMAC (80.8 vs.\ 84.3). Purely random schedules and Greedy are on par or below Fixed. Random sensing makes no data-dependent decision: it reads every sensor at a reduced rate. At 20\,mW, Random and Random-CA stay at or below the best Fixed schedule (Random-CA: 34.0 vs.\ 34.6 on Ego-Exo4D, 28.0 vs.\ 40.3 on CaptainCook4D), {whereas Greedy is below it (Acc: 30.7 vs.\ 34.6 on Ego-Exo4D, 66.0 vs.\ 70.8 on CMU-MMAC, 35.2 vs.\ 40.3 on CaptainCook4D).} Random does reach a higher Edit than Fixed on all datasets, possibly as sparse reads smooth predictions.
This suggests that the benefit comes from which sensors are read and how often, rather than when, a hypothesis we test in Sec.~\ref{sec:exp_learn}. {A direct test supports it: random reads are  no more accurate than periodic reads with the same reads per sensor and video (MiniROAD, five Ego-Exo4D scenarios; \cref{supp:add_random}).} Even when random sensing spends the 20\,mW budget under the edge-SoC energy model, it stays below the fixed schedule (40.2 vs.\ 42.8 Acc at 18.3 vs.\ 18.7\,mW, \cref{supp:edge_strategies}).

\begin{table*}[t]
\caption{{\textbf{TAS models and training.} MiniROAD and the Transformer beat ProTAS. One mixture  often matches the specialists at 2.8\,W and beats them at 20\,mW; the budget rule adds further gains.}} 
\label{tab:backbone_training}
\centering
\scriptsize
\setlength{\tabcolsep}{4pt}
\begin{tabular}[t]{@{}l rrr@{\hspace{0.3em}}}
\toprule
\multicolumn{4}{c}{\textbf{Ego-Exo4D}} \\
\midrule
Model & Acc & Edit & F1 \\
\midrule
\textit{2.8\,W} & \multicolumn{3}{r@{}}{\makebox[0pt][r]{\textit{Fixed: 6\,FPS R+A+G (P = 2.31\,W)}}} \\
ProTAS spec. & 48.0\makebox[1.3em][l]{$_{\scriptscriptstyle\textcolor{gray}{46.5}}$} & \textbf{29.2}\makebox[1.3em][l]{$_{\scriptscriptstyle\textcolor{gray}{34.5}}$} & \textbf{25.9}\makebox[1.3em][l]{$_{\scriptscriptstyle\textcolor{gray}{28.7}}$} \\
ProTAS mix. & 40.0\makebox[1.3em][l]{$_{\scriptscriptstyle\textcolor{gray}{39.0}}$} & 25.6\makebox[1.3em][l]{$_{\scriptscriptstyle\textcolor{gray}{28.2}}$} & 20.6\makebox[1.3em][l]{$_{\scriptscriptstyle\textcolor{gray}{22.7}}$} \\
ProTAS mix. + rule & 39.8\makebox[1.3em][l]{$_{\scriptscriptstyle\textcolor{gray}{38.8}}$} & 26.9\makebox[1.3em][l]{$_{\scriptscriptstyle\textcolor{gray}{28.5}}$} & 21.5\makebox[1.3em][l]{$_{\scriptscriptstyle\textcolor{gray}{22.9}}$} \\
\textit{\quad P = 1.99\,W} & \multicolumn{3}{r@{}}{\makebox[0pt][r]{\textit{Rule: R$_{3}$A$_{6}$G$_{6}$M$_{6}$I$_{6}$}}} \\
\cmidrule(lr){1-4}
MiniROAD spec. & \textbf{56.3}\makebox[1.3em][l]{$_{\scriptscriptstyle\textcolor{gray}{\mathbf{54.5}}}$} & 25.5\makebox[1.3em][l]{$_{\scriptscriptstyle\textcolor{gray}{\mathbf{37.7}}}$} & 25.7\makebox[1.3em][l]{$_{\scriptscriptstyle\textcolor{gray}{\mathbf{35.1}}}$} \\
MiniROAD mix. & 55.0\makebox[1.3em][l]{$_{\scriptscriptstyle\textcolor{gray}{53.8}}$} & 20.7\makebox[1.3em][l]{$_{\scriptscriptstyle\textcolor{gray}{35.1}}$} & 21.1\makebox[1.3em][l]{$_{\scriptscriptstyle\textcolor{gray}{33.7}}$} \\
MiniROAD mix. + rule & 55.4\makebox[1.3em][l]{$_{\scriptscriptstyle\textcolor{gray}{54.1}}$} & 22.9\makebox[1.3em][l]{$_{\scriptscriptstyle\textcolor{gray}{36.3}}$} & 22.4\makebox[1.3em][l]{$_{\scriptscriptstyle\textcolor{gray}{34.9}}$} \\
\textit{\quad P = 1.99\,W} & \multicolumn{3}{r@{}}{\makebox[0pt][r]{\textit{Rule: R$_{3}$A$_{6}$G$_{6}$M$_{6}$I$_{6}$}}} \\
\cmidrule(lr){1-4}
Transformer spec. & 53.4\makebox[1.3em][l]{$_{\scriptscriptstyle\textcolor{gray}{52.9}}$} & 12.1\makebox[1.3em][l]{$_{\scriptscriptstyle\textcolor{gray}{31.0}}$} & 11.6\makebox[1.3em][l]{$_{\scriptscriptstyle\textcolor{gray}{28.2}}$} \\
Transformer mix. & 52.1\makebox[1.3em][l]{$_{\scriptscriptstyle\textcolor{gray}{52.5}}$} & 9.8\makebox[1.3em][l]{$_{\scriptscriptstyle\textcolor{gray}{25.1}}$} & 9.9\makebox[1.3em][l]{$_{\scriptscriptstyle\textcolor{gray}{24.8}}$} \\
Transformer mix. + rule & 51.5\makebox[1.3em][l]{$_{\scriptscriptstyle\textcolor{gray}{51.3}}$} & 11.6\makebox[1.3em][l]{$_{\scriptscriptstyle\textcolor{gray}{27.1}}$} & 11.5\makebox[1.3em][l]{$_{\scriptscriptstyle\textcolor{gray}{26.9}}$} \\
\textit{\quad P = 1.99\,W} & \multicolumn{3}{r@{}}{\makebox[0pt][r]{\textit{Rule: R$_{3}$A$_{6}$G$_{6}$M$_{6}$I$_{6}$}}} \\
\midrule
\textit{20\,mW} & \multicolumn{3}{r@{}}{\makebox[0pt][r]{\textit{Fixed: 0.1\,FPS G+M (P = 17\,mW)}}} \\
ProTAS spec. & 34.6\makebox[1.3em][l]{$_{\scriptscriptstyle\textcolor{gray}{34.0}}$} & 22.5\makebox[1.3em][l]{$_{\scriptscriptstyle\textcolor{gray}{22.5}}$} & 18.1\makebox[1.3em][l]{$_{\scriptscriptstyle\textcolor{gray}{17.6}}$} \\
ProTAS mix. & 34.7\makebox[1.3em][l]{$_{\scriptscriptstyle\textcolor{gray}{34.1}}$} & 26.3\makebox[1.3em][l]{$_{\scriptscriptstyle\textcolor{gray}{26.5}}$} & 18.5\makebox[1.3em][l]{$_{\scriptscriptstyle\textcolor{gray}{18.2}}$} \\
ProTAS mix. + rule & 37.4\makebox[1.3em][l]{$_{\scriptscriptstyle\textcolor{gray}{36.5}}$} & 26.9\makebox[1.3em][l]{$_{\scriptscriptstyle\textcolor{gray}{27.5}}$} & 20.4\makebox[1.3em][l]{$_{\scriptscriptstyle\textcolor{gray}{20.4}}$} \\
\textit{\quad P = 19\,mW} & \multicolumn{3}{r@{}}{\makebox[0pt][r]{\textit{Rule: A$_{0.3}$G$_{0.2}$I$_{6}$}}} \\
\cmidrule(lr){1-4}
MiniROAD spec. & 40.5\makebox[1.3em][l]{$_{\scriptscriptstyle\textcolor{gray}{39.8}}$} & 33.4\makebox[1.3em][l]{$_{\scriptscriptstyle\textcolor{gray}{33.3}}$} & 25.4\makebox[1.3em][l]{$_{\scriptscriptstyle\textcolor{gray}{25.8}}$} \\
MiniROAD mix. & 41.8\makebox[1.3em][l]{$_{\scriptscriptstyle\textcolor{gray}{40.6}}$} & \textbf{36.7}\makebox[1.3em][l]{$_{\scriptscriptstyle\textcolor{gray}{36.5}}$} & 26.2\makebox[1.3em][l]{$_{\scriptscriptstyle\textcolor{gray}{26.0}}$} \\
MiniROAD mix. + rule & \textbf{48.8}\makebox[1.3em][l]{$_{\scriptscriptstyle\textcolor{gray}{\mathbf{47.1}}}$} & 32.7\makebox[1.3em][l]{$_{\scriptscriptstyle\textcolor{gray}{\mathbf{36.9}}}$} & \textbf{27.4}\makebox[1.3em][l]{$_{\scriptscriptstyle\textcolor{gray}{\mathbf{29.0}}}$} \\
\textit{\quad P = 20\,mW} & \multicolumn{3}{r@{}}{\makebox[0pt][r]{\textit{Rule: G$_{0.3}$I$_{6}$}}} \\
\cmidrule(lr){1-4}
Transformer spec. & 38.7\makebox[1.3em][l]{$_{\scriptscriptstyle\textcolor{gray}{37.9}}$} & 31.8\makebox[1.3em][l]{$_{\scriptscriptstyle\textcolor{gray}{31.5}}$} & 24.9\makebox[1.3em][l]{$_{\scriptscriptstyle\textcolor{gray}{24.2}}$} \\
Transformer mix. & 40.7\makebox[1.3em][l]{$_{\scriptscriptstyle\textcolor{gray}{39.5}}$} & 33.2\makebox[1.3em][l]{$_{\scriptscriptstyle\textcolor{gray}{33.3}}$} & 26.9\makebox[1.3em][l]{$_{\scriptscriptstyle\textcolor{gray}{26.0}}$} \\
Transformer mix. + rule & 45.2\makebox[1.3em][l]{$_{\scriptscriptstyle\textcolor{gray}{43.8}}$} & 24.0\makebox[1.3em][l]{$_{\scriptscriptstyle\textcolor{gray}{33.3}}$} & 18.8\makebox[1.3em][l]{$_{\scriptscriptstyle\textcolor{gray}{25.0}}$} \\
\textit{\quad P = 19\,mW} & \multicolumn{3}{r@{}}{\makebox[0pt][r]{\textit{Rule: A$_{0.3}$G$_{0.2}$I$_{6}$}}} \\
\bottomrule
\end{tabular}
\hfill
\begin{tabular}[t]{@{}l rrr@{\hspace{0.3em}}}
\toprule
\multicolumn{4}{c}{\textbf{CMU-MMAC}} \\
\midrule
Model & Acc & Edit & F1 \\
\midrule
\textit{2.8\,W} & \multicolumn{3}{r@{}}{\makebox[0pt][r]{\textit{Fixed: 6\,FPS R (P = {1.86}\,W)}}} \\
ProTAS spec. & 84.3\makebox[1.3em][l]{$_{\scriptscriptstyle\textcolor{gray}{82.7}}$} & 39.1\makebox[1.3em][l]{$_{\scriptscriptstyle\textcolor{gray}{47.1}}$} & 40.7\makebox[1.3em][l]{$_{\scriptscriptstyle\textcolor{gray}{46.9}}$} \\
ProTAS mix. & 81.9\makebox[1.3em][l]{$_{\scriptscriptstyle\textcolor{gray}{81.0}}$} & 35.5\makebox[1.3em][l]{$_{\scriptscriptstyle\textcolor{gray}{40.4}}$} & 38.9\makebox[1.3em][l]{$_{\scriptscriptstyle\textcolor{gray}{42.3}}$} \\
ProTAS mix. + rule & 80.9\makebox[1.3em][l]{$_{\scriptscriptstyle\textcolor{gray}{79.9}}$} & 35.1\makebox[1.3em][l]{$_{\scriptscriptstyle\textcolor{gray}{37.5}}$} & 36.8\makebox[1.3em][l]{$_{\scriptscriptstyle\textcolor{gray}{38.4}}$} \\
\textit{\quad P = 1.94\,W} & \multicolumn{3}{r@{}}{\makebox[0pt][r]{\textit{Rule: R$_{6}$A$_{6}$I$_{6}$}}} \\
\cmidrule(lr){1-4}
MiniROAD spec. & \textbf{88.3}\makebox[1.3em][l]{$_{\scriptscriptstyle\textcolor{gray}{\mathbf{86.5}}}$} & \textbf{42.5}\makebox[1.3em][l]{$_{\scriptscriptstyle\textcolor{gray}{\mathbf{56.8}}}$} & \textbf{47.4}\makebox[1.3em][l]{$_{\scriptscriptstyle\textcolor{gray}{56.2}}$} \\
MiniROAD mix. & 86.6\makebox[1.3em][l]{$_{\scriptscriptstyle\textcolor{gray}{86.0}}$} & 28.0\makebox[1.3em][l]{$_{\scriptscriptstyle\textcolor{gray}{49.2}}$} & 35.2\makebox[1.3em][l]{$_{\scriptscriptstyle\textcolor{gray}{\mathbf{58.0}}}$} \\
MiniROAD mix. + rule & 85.8\makebox[1.3em][l]{$_{\scriptscriptstyle\textcolor{gray}{84.8}}$} & 30.9\makebox[1.3em][l]{$_{\scriptscriptstyle\textcolor{gray}{50.5}}$} & 35.4\makebox[1.3em][l]{$_{\scriptscriptstyle\textcolor{gray}{55.5}}$} \\
\textit{\quad P = 1.94\,W} & \multicolumn{3}{r@{}}{\makebox[0pt][r]{\textit{Rule: R$_{6}$A$_{6}$I$_{6}$}}} \\
\cmidrule(lr){1-4}
Transformer spec. & 85.5\makebox[1.3em][l]{$_{\scriptscriptstyle\textcolor{gray}{84.8}}$} & 15.3\makebox[1.3em][l]{$_{\scriptscriptstyle\textcolor{gray}{44.2}}$} & 18.2\makebox[1.3em][l]{$_{\scriptscriptstyle\textcolor{gray}{47.6}}$} \\
Transformer mix. & 83.1\makebox[1.3em][l]{$_{\scriptscriptstyle\textcolor{gray}{83.4}}$} & 12.5\makebox[1.3em][l]{$_{\scriptscriptstyle\textcolor{gray}{39.9}}$} & 14.7\makebox[1.3em][l]{$_{\scriptscriptstyle\textcolor{gray}{41.0}}$} \\
Transformer mix. + rule & 83.8\makebox[1.3em][l]{$_{\scriptscriptstyle\textcolor{gray}{83.5}}$} & 15.0\makebox[1.3em][l]{$_{\scriptscriptstyle\textcolor{gray}{39.1}}$} & 16.3\makebox[1.3em][l]{$_{\scriptscriptstyle\textcolor{gray}{42.2}}$} \\
\textit{\quad P = 1.94\,W} & \multicolumn{3}{r@{}}{\makebox[0pt][r]{\textit{Rule: R$_{6}$A$_{6}$I$_{6}$}}} \\
\midrule
\textit{20\,mW} & \multicolumn{3}{r@{}}{\makebox[0pt][r]{\textit{Fixed: 0.05\,FPS R+A+I (P = 17\,mW)}}} \\
ProTAS spec. & 70.8\makebox[1.3em][l]{$_{\scriptscriptstyle\textcolor{gray}{70.2}}$} & 24.2\makebox[1.3em][l]{$_{\scriptscriptstyle\textcolor{gray}{24.2}}$} & 26.8\makebox[1.3em][l]{$_{\scriptscriptstyle\textcolor{gray}{26.8}}$} \\
ProTAS mix. & 70.0\makebox[1.3em][l]{$_{\scriptscriptstyle\textcolor{gray}{69.2}}$} & 37.0\makebox[1.3em][l]{$_{\scriptscriptstyle\textcolor{gray}{38.1}}$} & 26.7\makebox[1.3em][l]{$_{\scriptscriptstyle\textcolor{gray}{26.2}}$} \\
ProTAS mix. + rule & 72.1\makebox[1.3em][l]{$_{\scriptscriptstyle\textcolor{gray}{71.3}}$} & 32.1\makebox[1.3em][l]{$_{\scriptscriptstyle\textcolor{gray}{33.7}}$} & 28.1\makebox[1.3em][l]{$_{\scriptscriptstyle\textcolor{gray}{27.1}}$} \\
\textit{\quad P = 20\,mW} & \multicolumn{3}{r@{}}{\makebox[0pt][r]{\textit{Rule: R$_{0.06}$I$_{6}$}}} \\
\cmidrule(lr){1-4}
MiniROAD spec. & 70.2\makebox[1.3em][l]{$_{\scriptscriptstyle\textcolor{gray}{69.8}}$} & 27.6\makebox[1.3em][l]{$_{\scriptscriptstyle\textcolor{gray}{27.6}}$} & 27.3\makebox[1.3em][l]{$_{\scriptscriptstyle\textcolor{gray}{25.9}}$} \\
MiniROAD mix. & 71.3\makebox[1.3em][l]{$_{\scriptscriptstyle\textcolor{gray}{70.3}}$} & \textbf{40.3}\makebox[1.3em][l]{$_{\scriptscriptstyle\textcolor{gray}{\mathbf{40.9}}}$} & \textbf{31.3}\makebox[1.3em][l]{$_{\scriptscriptstyle\textcolor{gray}{31.6}}$} \\
MiniROAD mix. + rule & 73.2\makebox[1.3em][l]{$_{\scriptscriptstyle\textcolor{gray}{71.8}}$} & 33.6\makebox[1.3em][l]{$_{\scriptscriptstyle\textcolor{gray}{40.0}}$} & 27.9\makebox[1.3em][l]{$_{\scriptscriptstyle\textcolor{gray}{\mathbf{32.2}}}$} \\
\textit{\quad P = 20\,mW} & \multicolumn{3}{r@{}}{\makebox[0pt][r]{\textit{Rule: R$_{0.06}$I$_{6}$}}} \\
\cmidrule(lr){1-4}
Transformer spec. & 72.1\makebox[1.3em][l]{$_{\scriptscriptstyle\textcolor{gray}{71.7}}$} & 26.6\makebox[1.3em][l]{$_{\scriptscriptstyle\textcolor{gray}{26.6}}$} & 30.2\makebox[1.3em][l]{$_{\scriptscriptstyle\textcolor{gray}{30.2}}$} \\
Transformer mix. & 70.1\makebox[1.3em][l]{$_{\scriptscriptstyle\textcolor{gray}{69.1}}$} & 39.8\makebox[1.3em][l]{$_{\scriptscriptstyle\textcolor{gray}{40.3}}$} & 30.0\makebox[1.3em][l]{$_{\scriptscriptstyle\textcolor{gray}{29.1}}$} \\
Transformer mix. + rule & \textbf{74.5}\makebox[1.3em][l]{$_{\scriptscriptstyle\textcolor{gray}{\mathbf{73.3}}}$} & 22.9\makebox[1.3em][l]{$_{\scriptscriptstyle\textcolor{gray}{36.0}}$} & 21.0\makebox[1.3em][l]{$_{\scriptscriptstyle\textcolor{gray}{32.1}}$} \\
\textit{\quad P = 20\,mW} & \multicolumn{3}{r@{}}{\makebox[0pt][r]{\textit{Rule: R$_{0.06}$I$_{6}$}}} \\
\bottomrule
\end{tabular}
\hfill
\begin{tabular}[t]{@{}l rrr@{\hspace{0.3em}}}
\toprule
\multicolumn{4}{c}{\textbf{CaptainCook4D}} \\
\midrule
Model & Acc & Edit & F1 \\
\midrule
\textit{2.8\,W} & \multicolumn{3}{r@{}}{\makebox[0pt][r]{\textit{Fixed: 0.5\,FPS R (P = 0.15\,W)}}} \\
ProTAS spec. & 43.5\makebox[1.3em][l]{$_{\scriptscriptstyle\textcolor{gray}{44.3}}$} & \textbf{38.2}\makebox[1.3em][l]{$_{\scriptscriptstyle\textcolor{gray}{\mathbf{37.5}}}$} & \textbf{23.1}\makebox[1.3em][l]{$_{\scriptscriptstyle\textcolor{gray}{23.0}}$} \\
ProTAS mix. & 44.3\makebox[1.3em][l]{$_{\scriptscriptstyle\textcolor{gray}{44.3}}$} & 33.0\makebox[1.3em][l]{$_{\scriptscriptstyle\textcolor{gray}{34.0}}$} & 21.7\makebox[1.3em][l]{$_{\scriptscriptstyle\textcolor{gray}{22.9}}$} \\
ProTAS mix. + rule & 47.3\makebox[1.3em][l]{$_{\scriptscriptstyle\textcolor{gray}{47.4}}$} & 33.0\makebox[1.3em][l]{$_{\scriptscriptstyle\textcolor{gray}{35.1}}$} & 20.6\makebox[1.3em][l]{$_{\scriptscriptstyle\textcolor{gray}{22.8}}$} \\
\textit{\quad P = 1.91\,W} & \multicolumn{3}{r@{}}{\makebox[0pt][r]{\textit{Rule: R$_{6}$A$_{6}$I$_{6}$}}} \\
\cmidrule(lr){1-4}
MiniROAD spec. & 52.7\makebox[1.3em][l]{$_{\scriptscriptstyle\textcolor{gray}{52.5}}$} & 23.0\makebox[1.3em][l]{$_{\scriptscriptstyle\textcolor{gray}{26.2}}$} & 22.2\makebox[1.3em][l]{$_{\scriptscriptstyle\textcolor{gray}{\mathbf{26.0}}}$} \\
MiniROAD mix. & 52.7\makebox[1.3em][l]{$_{\scriptscriptstyle\textcolor{gray}{52.9}}$} & 20.1\makebox[1.3em][l]{$_{\scriptscriptstyle\textcolor{gray}{23.7}}$} & 19.0\makebox[1.3em][l]{$_{\scriptscriptstyle\textcolor{gray}{22.1}}$} \\
MiniROAD mix. + rule & \textbf{53.9}\makebox[1.3em][l]{$_{\scriptscriptstyle\textcolor{gray}{\mathbf{53.9}}}$} & 11.3\makebox[1.3em][l]{$_{\scriptscriptstyle\textcolor{gray}{23.0}}$} & 10.8\makebox[1.3em][l]{$_{\scriptscriptstyle\textcolor{gray}{20.3}}$} \\
\textit{\quad P = 1.82\,W} & \multicolumn{3}{r@{}}{\makebox[0pt][r]{\textit{Rule: R$_{6}$I$_{6}$}}} \\
\cmidrule(lr){1-4}
Transformer spec. & 47.0\makebox[1.3em][l]{$_{\scriptscriptstyle\textcolor{gray}{47.4}}$} & 16.5\makebox[1.3em][l]{$_{\scriptscriptstyle\textcolor{gray}{19.7}}$} & 13.3\makebox[1.3em][l]{$_{\scriptscriptstyle\textcolor{gray}{16.5}}$} \\
Transformer mix. & 47.3\makebox[1.3em][l]{$_{\scriptscriptstyle\textcolor{gray}{47.4}}$} & 13.5\makebox[1.3em][l]{$_{\scriptscriptstyle\textcolor{gray}{14.3}}$} & 11.5\makebox[1.3em][l]{$_{\scriptscriptstyle\textcolor{gray}{12.5}}$} \\
Transformer mix. + rule & 53.1\makebox[1.3em][l]{$_{\scriptscriptstyle\textcolor{gray}{\mathbf{53.9}}}$} & 3.1\makebox[1.3em][l]{$_{\scriptscriptstyle\textcolor{gray}{11.3}}$} & 2.2\makebox[1.3em][l]{$_{\scriptscriptstyle\textcolor{gray}{9.7}}$} \\
\textit{\quad P = 1.91\,W} & \multicolumn{3}{r@{}}{\makebox[0pt][r]{\textit{Rule: R$_{6}$A$_{6}$I$_{6}$}}} \\
\midrule
\textit{20\,mW} & \multicolumn{3}{r@{}}{\makebox[0pt][r]{\textit{Fixed: 0.05\,FPS R+A (P = 16\,mW)}}} \\
ProTAS spec. & 40.3\makebox[1.3em][l]{$_{\scriptscriptstyle\textcolor{gray}{40.8}}$} & 31.9\makebox[1.3em][l]{$_{\scriptscriptstyle\textcolor{gray}{32.8}}$} & 17.7\makebox[1.3em][l]{$_{\scriptscriptstyle\textcolor{gray}{18.1}}$} \\
ProTAS mix. & 45.0\makebox[1.3em][l]{$_{\scriptscriptstyle\textcolor{gray}{45.1}}$} & 36.4\makebox[1.3em][l]{$_{\scriptscriptstyle\textcolor{gray}{37.7}}$} & 23.7\makebox[1.3em][l]{$_{\scriptscriptstyle\textcolor{gray}{24.4}}$} \\
ProTAS mix. + rule & 44.7\makebox[1.3em][l]{$_{\scriptscriptstyle\textcolor{gray}{44.7}}$} & 34.6\makebox[1.3em][l]{$_{\scriptscriptstyle\textcolor{gray}{36.0}}$} & 20.9\makebox[1.3em][l]{$_{\scriptscriptstyle\textcolor{gray}{21.3}}$} \\
\textit{\quad P = 19\,mW} & \multicolumn{3}{r@{}}{\makebox[0pt][r]{\textit{Rule: R$_{0.04}$A$_{0.3}$I$_{6}$}}} \\
\cmidrule(lr){1-4}
MiniROAD spec. & 43.6\makebox[1.3em][l]{$_{\scriptscriptstyle\textcolor{gray}{43.7}}$} & 35.2\makebox[1.3em][l]{$_{\scriptscriptstyle\textcolor{gray}{35.5}}$} & 22.0\makebox[1.3em][l]{$_{\scriptscriptstyle\textcolor{gray}{22.6}}$} \\
MiniROAD mix. & 47.5\makebox[1.3em][l]{$_{\scriptscriptstyle\textcolor{gray}{47.5}}$} & \textbf{40.6}\makebox[1.3em][l]{$_{\scriptscriptstyle\textcolor{gray}{\mathbf{42.5}}}$} & \textbf{29.0}\makebox[1.3em][l]{$_{\scriptscriptstyle\textcolor{gray}{\mathbf{29.7}}}$} \\
MiniROAD mix. + rule & \textbf{48.2}\makebox[1.3em][l]{$_{\scriptscriptstyle\textcolor{gray}{\mathbf{48.7}}}$} & 31.2\makebox[1.3em][l]{$_{\scriptscriptstyle\textcolor{gray}{36.9}}$} & 20.9\makebox[1.3em][l]{$_{\scriptscriptstyle\textcolor{gray}{24.8}}$} \\
\textit{\quad P = 19\,mW} & \multicolumn{3}{r@{}}{\makebox[0pt][r]{\textit{Rule: R$_{0.06}$I$_{6}$}}} \\
\cmidrule(lr){1-4}
Transformer spec. & 45.6\makebox[1.3em][l]{$_{\scriptscriptstyle\textcolor{gray}{46.1}}$} & 38.1\makebox[1.3em][l]{$_{\scriptscriptstyle\textcolor{gray}{38.2}}$} & 23.5\makebox[1.3em][l]{$_{\scriptscriptstyle\textcolor{gray}{23.8}}$} \\
Transformer mix. & 45.2\makebox[1.3em][l]{$_{\scriptscriptstyle\textcolor{gray}{45.1}}$} & 37.1\makebox[1.3em][l]{$_{\scriptscriptstyle\textcolor{gray}{37.9}}$} & 25.0\makebox[1.3em][l]{$_{\scriptscriptstyle\textcolor{gray}{25.1}}$} \\
Transformer mix. + rule & 47.7\makebox[1.3em][l]{$_{\scriptscriptstyle\textcolor{gray}{48.3}}$} & 13.5\makebox[1.3em][l]{$_{\scriptscriptstyle\textcolor{gray}{21.9}}$} & 10.3\makebox[1.3em][l]{$_{\scriptscriptstyle\textcolor{gray}{15.7}}$} \\
\textit{\quad P = 19\,mW} & \multicolumn{3}{r@{}}{\makebox[0pt][r]{\textit{Rule: R$_{0.04}$A$_{0.3}$I$_{6}$}}} \\
\bottomrule
\end{tabular}

\vspace{2pt}
{\footnotesize
\textbf{Sensors:} R = RGB, A = Audio, I = IMU, G = Gaze, M = Monochrome; subscripts of schedules: FPS. {\textbf{Rule:} schedule selected by the budget rule on validation for each model and budget (reads per second of each sensor); mix.\ + rule is the mixture run on it.} \textbf{Large:} raw \textcolor{gray}{small gray}: causal 2\,s mode filter.\par}

\end{table*}

\subsection{{TAS models and training}}
\label{sec:exp_backbone}
Table~\ref{tab:backbone_training} compares ProTAS, MiniROAD, and the Transformer, each trained {either} as a specialist on the best Fixed schedule of Table~\ref{tab:all_protas} {(one model per budget)} {or} as a single model on a mixture of randomized schedules {(one model for both budgets)}. Gray numbers are obtained with a causal 2\,s mode filter applied to every model.

\paragraph{MiniROAD and the Transformer outperform ProTAS.} As specialists, MiniROAD beats ProTAS in five of six settings ($+8.3$, $+4.0$, and $+9.2$ Acc points at 2.8\,W; $+5.9$, $-0.6$, and $+3.3$ at 20\,mW, on Ego-Exo4D, CMU-MMAC, and CaptainCook4D), and the Transformer in all six ($+5.4$, $+1.2$, $+3.5$ at 2.8\,W; $+4.1$, $+1.3$, $+5.3$ at 20\,mW). The Transformer fragments  predictions, especially at high rates (Edit 12.1 vs.\ 29.2 for ProTAS, Ego-Exo4D at 2.8\,W), {which the filter largely removes (31.0).} On Ego-Exo4D at 2.8\,W, ProTAS  has higher raw Edit and F1 than MiniROAD (29.2 and 25.9 vs.\ 25.5 and 25.7), but with the filter MiniROAD is above it (37.7 and 35.1 vs.\ 34.5 and 28.7). MiniROAD and the Transformer thus produce more fragmented raw predictions than ProTAS, which a simple causal filter corrects. {MiniROAD is larger (16.0M parameters vs.\ 0.98M for ProTAS and 1.23M for the Transformer), but the gain is not due to size: a mixture reduced to 1.2M parameters stays ahead of both}  on Ego-Exo4D (\cref{supp:add_capacity}).

\paragraph{One mixture model can replace the specialists.} At 2.8\,W, the MiniROAD mixture is 1.3 and 1.7 Acc points below its specialist on Ego-Exo4D and CMU-MMAC and equal on CaptainCook4D (52.7), while at 20\,mW it is above it on all three  ($+1.3$, $+1.1$, $+3.9$). The Transformer behaves alike at 2.8\,W ($-1.3$, $-2.4$, $+0.3$) and is mixed at 20\,mW ($+2.0$, $-2.0$, $-0.4$). ProTAS is less consistent (${-8.0}$ on Ego-Exo4D at 2.8\,W, $+4.7$ on CaptainCook4D at 20\,mW). 

\paragraph{The budget rule helps most at low budgets.} The rule, which needs only validation videos, raises the MiniROAD mixture by $+7.0$, $+1.9$, and $+0.7$ points at 20\,mW and stays within 1.2 points of the same mixture on the Fixed schedule at 2.8\,W ($+0.4$, $-0.8$, $+1.2$). It can lower raw Edit and F1 (e.g., CaptainCook4D at 2.8\,W, Edit 11.3 vs.\ 20.1), but the filtered values are close (23.0 vs.\ 23.7). It selects a schedule without per-schedule training, whereas the Fixed schedules of \cref{tab:all_protas} required training several models. The gain holds against Fixed schedules chosen for each mixture on validation (changes of at most 1.2 points, rule still ahead at 20\,mW; \cref{supp:add_fixedref}) and against a {per-sensor} rate grid, CMA-ES, and random read times (most accurate or tied at eight of nine budgets; \cref{supp:add_rule_selection}).

\subsection{Robustness to schedule changes}
\label{sec:exp_mismatch}
{To study robustness regardless of the budget, we fix a single specialist schedule (RGB+Audio+Gaze at 6\,FPS).} Table~\ref{tab:mismatch} reports, for the three models, the difference between the mixture and {this} specialist when both are tested on the same test schedules (the first is the specialist's own).

\paragraph{The mixture stays close to the specialist and is better with missing sensors.} On the specialist's own schedule, the Transformer and MiniROAD mixtures are within 0.1 and 0.4 Acc points of it ($-0.1$ and $+0.4$), and at lower rates they are at or above it (up to $+2.9$). ProTAS generalizes worse: its mixture is 5.8 points below the specialist on the specialist's own schedule (5.0 at 1\,FPS, 1.3 at 0.1\,FPS). With missing sensors, however, the mixture is better for all three models: with audio only it gains $+12.4$ (ProTAS), $+10.0$ (Transformer), and $+15.2$ points (MiniROAD), and the Transformer gains $+11.2$ without RGB. This is expected: the specialist never faces a missing sensor in training, whereas the mixture does. The price is F1 on the specialist's own schedule (MiniROAD $-2.3$, ProTAS $-4.0$). Here the mixture is trained only on randomized schedules of RGB, audio and gaze, the sensors of the specialist.  Absolute values (with Edit) are in \cref{supp:mismatch_abs}, the mixture of all five sensors used in the other tables (larger gap on the specialist's own schedule) in \cref{supp:mismatch_all}, and the comparison with a 20\,mW specialist (all-sensors mixture) in \cref{supp:mismatch_20mw}.

\begin{table}[t]
\caption{\textbf{Effect of training on mixtures.} {The mixture stays close to the specialist on its schedule (MiniROAD, Transformer) and improves all three with missing sensors, most with audio only.}}
\label{tab:mismatch}
\centering
\scriptsize
\setlength{\tabcolsep}{3pt}
\begin{tabular}{@{}l rr rr rr@{}}
\toprule
 & \multicolumn{2}{c}{ProTAS} & \multicolumn{2}{c}{Transformer} & \multicolumn{2}{c}{MiniROAD} \\
\cmidrule(lr){2-3}\cmidrule(lr){4-5}\cmidrule(lr){6-7}
Test schedule (R, A, G) & Acc & F1 & Acc & F1 & Acc & F1 \\
\midrule
\textit{$\Delta$ (mixture $-$ specialist)} & \multicolumn{6}{r@{}}{\textit{Specialist trained on 6\,FPS R+A+G}} \\
Regular, 6\,FPS (training schedule) & \cellcolor{red!70!black!30}$-$5.8 & \cellcolor{red!70!black!12}$-$4.0 & $-$0.1 & $-$0.9 & $+$0.4 & \cellcolor{red!70!black!12}$-$2.3 \\
\midrule
Regular, 1\,FPS & \cellcolor{red!70!black!30}$-$5.0 & \cellcolor{red!70!black!12}$-$3.2 & $+$0.4 & $+$0.2 & $+$1.3 & $-$0.1 \\
Regular, 0.1\,FPS & $-$1.3 & $-$1.6 & $+$0.9 & $+$1.6 & \cellcolor{green!50!black!12}$+$2.9 & \cellcolor{green!50!black!12}$+$2.0 \\
\midrule
Gaze absent & $+$1.7 & $+$0.8 & \cellcolor{green!50!black!12}$+$3.5 & $-$0.3 & \cellcolor{green!50!black!30}$+$7.0 & \cellcolor{green!50!black!12}$+$2.5 \\
RGB absent & \cellcolor{green!50!black!12}$+$3.0 & $+$1.7 & \cellcolor{green!50!black!30}$+$11.2 & $+$1.5 & \cellcolor{green!50!black!30}$+$8.3 & \cellcolor{green!50!black!12}$+$2.4 \\
Audio only & \cellcolor{green!50!black!30}$+$12.4 & \cellcolor{green!50!black!30}$+$7.0 & \cellcolor{green!50!black!30}$+$10.0 & $+$0.8 & \cellcolor{green!50!black!30}$+$15.2 & \cellcolor{green!50!black!30}$+$7.8 \\
Gaze only & \cellcolor{green!50!black!12}$+$2.3 & \cellcolor{green!50!black!12}$+$3.5 & \cellcolor{green!50!black!30}$+$9.3 & \cellcolor{green!50!black!12}$+$2.5 & \cellcolor{green!50!black!30}$+$7.5 & \cellcolor{green!50!black!12}$+$3.3 \\
RGB only & \cellcolor{green!50!black!12}$+$2.1 & $+$0.5 & \cellcolor{green!50!black!12}$+$3.2 & $+$0.9 & \cellcolor{green!50!black!30}$+$5.1 & $+$1.5 \\
\bottomrule
\end{tabular}
\par\vspace{2pt}
{\footnotesize R = RGB, A = Audio, G = Gaze. $\Delta$: mixture minus specialist tested on the same schedule. Shades: light for $|\Delta|\geq 2$, strong for $|\Delta|\geq 5$ points.}

\end{table}

\subsection{Learned Policies}
\label{sec:exp_learn}
Table~\ref{tab:learn_where} tests the learned policies introduced in \cref{sec:learned_pol} on a frozen MiniROAD mixture and investigates what is worth learning.
Fixed-stride schedules settle which sensors to read and how often, whereas learned policies tackle \textit{when} {to read.} ``Periodic'' below refers to the periodic schedules of the budget rule, where each sensor is read at its own rate.

\paragraph{Learned policies do not beat periodic reads with the same reads.}
At 20\,mW, the AdaMML-style and HCMS-style policies are 8.8 and 8.6 Acc points below the budget rule (40.0 and 40.2 vs.\ 48.8), and the closed-loop policy is 2.8 below it (46.0). {Periodic reads with the same counts beat the three policies by} $+7.4$, $+6.9$, and $+1.7$ points (``periodic, same reads''), so the learned timing does not help. At 2.8\,W, the policies are within 1.4 points of the rule (54.0--55.3 vs.\ 55.4) but spend more (2.25--2.80 vs.\ 1.99\,W), and periodic reads with their counts are as good or better (54.4--55.4). {This holds with a label-free objective or without the time-since-last-read input (\cref{supp:learned_variants}), and {when trained end-to-end with ProTAS}, still below the best Fixed schedule (\cref{supp:learned_e2e}).}

\paragraph{Timing has headroom only for a non-causal oracle.} We thus asked whether random sensing or reading at true action boundaries improves over periodic. 
With the same reads, timing does not matter at 2.8\,W (random, true boundaries, and the oracle are within 0.1 Acc points of periodic). At 20\,mW, random timing loses 1.8 points and reading at the true action boundaries does not help ($-0.3$). An oracle that optimizes the read times in 5\,s windows on each test video gains $+5.9$ points, but it is non-causal and fitted to the model on the same videos, so its gain is optimistic and not an upper bound, and a causal policy trained to imitate it does not recover it (47.3, $-1.5$ vs.\ periodic; at 2.8\,W {55.4, $0.0$}). Timing thus has limited headroom, mostly at low budgets
The oracle search and the causal policy imitating it are described in \cref{supp:read_timing}, and the results with other window lengths and models, and the learnability of the oracle's reads, in \cref{supp:add_policies}.

\begin{table}[t]
\caption{\textbf{What is worth learning.} {Ego-Exo4D, frozen MiniROAD mixture.} (a) Learned policies vs.\  periodic reads with the same reads per sensor; $^{*}$above the budget in some runs (on average only for the closed loop at 20\,mW). (b) Same reads, only timing changes: {only the non-causal oracle gains, at 20\,mW.} {The imitating row is the mean of two causal-policy seeds and two thresholds (\cref{supp:causal_router}).} Bold: best within budget in (a), best causal schedule in (b).
}
\label{tab:learn_where}
\centering
\scriptsize
\setlength{\tabcolsep}{4pt}
\begin{tabular}{@{}l rrr rrr@{}}
\toprule
 & \multicolumn{3}{c}{20\,mW} & \multicolumn{3}{c}{2.8\,W} \\
\cmidrule(lr){2-4}\cmidrule(lr){5-7}
 & Acc & F1 & P (mW) & Acc & F1 & P (W) \\
\midrule
\multicolumn{7}{@{}l}{\textit{(a) Learned policies, {target} budget}} \\
Budget rule (periodic) & \textbf{48.8} & \textbf{27.4} & 19.7$^{\ \ }$ & 55.4 & 22.4 & 1.99$^{\ \ }$ \\
AdaMML-style & 40.0 & 21.1 & 18.5$^{*}$ & 55.3 & 23.1 & 2.79$^{*}$ \\
\quad periodic, same reads & 47.4 & 26.5 & 18.5$^{*}$ & 55.3 & 22.5 & 2.79$^{*}$ \\
HCMS-style & 40.2 & 21.9 & 19.3$^{*}$ & 54.0 & 21.1 & 2.25$^{*}$ \\
\quad periodic, same reads & 47.1 & 27.1 & 19.3$^{*}$ & 54.4 & 21.6 & 2.25$^{*}$ \\
Closed loop & 46.0 & 25.8 & 21.7$^{*}$ & 55.2 & 23.5 & 2.80$^{*}$ \\
\quad periodic, same reads & 47.7 & 26.4 & 21.7$^{*}$ & 55.4 & 23.0 & 2.80$^{*}$ \\
\midrule
\multicolumn{7}{@{}l}{\textit{(b) Read timing, same reads}} \\
Periodic & \textbf{48.8} & \textbf{27.4} & 19.7 & \textbf{{55.4}} & {22.4} & {1.99} \\
Random & 47.0 & 25.2 & 19.7 & {55.4} & \textbf{{23.0}} & {1.99} \\
At true action boundaries & 48.5 & 29.3 & 19.7 & {55.4} & {22.8} & {1.99} \\
\cmidrule(lr){1-7}
Oracle (non-causal, optim.) & 54.7 & 33.4 & 19.7 & {55.5} & {22.5} & {1.99} \\
\cmidrule(lr){1-7}
Learned, imitating the oracle (causal) & 47.3 & 26.6 & 19.7 & {55.4} & {22.6} & {1.99} \\
\bottomrule
\end{tabular}

\end{table}

\subsection{Experiments under a different energy model}
\label{sec:exp_edge}
We repeat the main comparisons on Ego-Exo4D under an edge-SoC energy model reflecting more modern hardware (\cref{tab:edge_small}), adding a more restrictive 5\,mW budget. Edge-SoC versions of Tables~\ref{tab:all_protas}, \ref{tab:backbone_training}, and \ref{tab:learn_where} are in \cref{supp:add_edge}.

\begin{table}[t]
\caption{\textbf{The conclusions hold under the edge-SoC energy model.} Frozen MiniROAD mixture. Periodic schedules: G$_{0.29}$I$_{6}$ at 5\,mW, G$_{0.86}$I$_{6}$M$_{0.18}$ at 20\,mW, R$_{6}$A$_{6}$G$_{6}$M$_{6}$I$_{6}$ at 2.8\,W. Specialist: best Fixed configuration within budget (G$_{0.1}$M$_{0.1}$ at 5\,mW, G$_{0.43}$M$_{0.43}$ at 20\,mW, A$_{6}$I$_{6}$G$_{6}$M$_{6}$ at 2.8\,W).}
\label{tab:edge_small}
\centering
\scriptsize
\setlength{\tabcolsep}{4pt}
\begin{tabular}{@{}l rrr@{}}
\toprule
Ego-Exo4D, edge-SoC (Accuracy) & 5\,mW & 20\,mW & 2.8\,W \\
\midrule
Periodic (budget-rule reads), MiniROAD mix. & 48.7 & 51.7 & 55.5 \\
Specialist, Fixed & 40.5 & 49.6 & 56.2 \\
Closed loop & 46.4 & 50.3 & 55.5 \\
\quad periodic, same reads & 46.6 & 51.7 & 55.5 \\
\textit{Same reads as periodic:} & & & \\
\quad Random & 46.2 & 50.7 & 55.5 \\
\quad Oracle (non-causal) & 54.7 & 54.6 & 55.5 \\
\quad Causal policy imitating the oracle & 47.5 & 50.2 & 55.5 \\
\bottomrule
\end{tabular}

\end{table}

\paragraph{Same conclusions with the edge-SoC energy model.} The periodic reads of the budget rule give the best accuracy among the causal strategies at 5 and 20\,mW (48.7 and 51.7 Acc). They are 8.2 and 2.1 points above the Fixed G+M specialist (48.7 vs.\ 40.5 and 51.7 vs.\ 49.6). Random reads with the same counts lose 2.5 and 1.0 points, and the causal policy imitating the oracle loses 1.2 and 1.5. The closed loop is 0.2 and 1.4 points below periodic reads with its own counts at 5 and 20\,mW. Only the non-causal oracle improves on periodic reads ($+$6.0 and $+$2.9 points at 5 and 20\,mW). At 2.8\,W the budget does not bind under this model (the rule reads every sensor at 6\,{FPS}, 735\,mW), so all strategies coincide at 55.5 Acc, 0.7 points below the specialist reading A+I+G+M at 6\,FPS (56.2).

\paragraph{{Statistical significance.}} Paired bootstrap intervals over test videos (\cref{supp:stats}) confirm the gains of the budget rule and of the oracle at 20\,mW, and that no learned policy improves on periodic reads. The 2.8\,W differences between specialist, mixture, and rule are small and should be read as close.

\subsection{Qualitative examples and Trade-Offs}
\label{sec:qualitative_examples}

\cref{supp:energy_accuracy} gives accuracy--power curves like \cref{fig:teaser} for the three datasets and both energy models: the budget rule is the most accurate option.
\cref{fig:qualitative_sat} shows a test video of Ego-Exo4D (CMU-MMAC: \cref{fig:supp_qualitative_cmu} in the supplement), processed by one MiniROAD mixture model under the budget-rule schedules at 2.8\,W and 20\,mW (\cref{tab:backbone_training}). At 2.8\,W, models read every sensor (on Ego-Exo4D, RGB at 3\,FPS, others at 6\,FPS) and their predictions are closer to the ground truth (58.2 accuracy). At 20\,mW, they read the IMU at every step and gaze every 3.3\,s:  predictions are more fragmented and accuracy drops to 49.4, but most segments are found.

\begin{figure}[t]
    \centering
    \includegraphics[width=\linewidth]{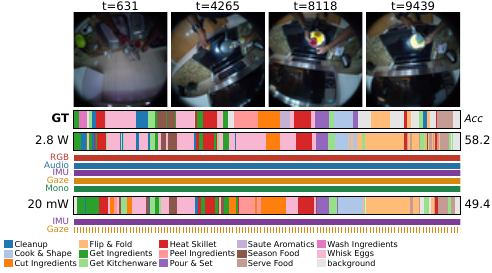}
    \caption{{\textbf{Qualitative result on Ego-Exo4D.} One MiniROAD model, two budgets.}}
    \label{fig:qualitative_sat}
    \vspace{-10pt}
\end{figure}

\section{Conclusions}
\label{sec:conclusions}
We introduced Ego-METAS, the first benchmark for online, energy-efficient multimodal temporal action segmentation, on three egocentric datasets. Efficiency comes from which sensors are read and how often rather than when (only a non-causal oracle gains from timing): low-rate multimodal periodic schedules selected on validation match dense sensing and learned policies, and a recurrent model trained on randomized schedules serves every budget. Together, randomized training and the budget rule form a practical recipe {with MiniROAD} that is at most 2.5 Acc points below specialists at 2.8\,W and improves on them at 20\,mW.

\section{Acknowledgments}
This work was supported by projects PID2024-158322OB-I00 , (MCIN/AEI/10.13039/501100011033/ FEDER, UE), project JIUZ2024-IyA-07 and Aragon Government DGA T45-23R. We thank CINECA under the ISCRA initiative, for HPC resources and support.
\end{bibunit}

\clearpage\begin{bibunit}

\makeatletter
\let\suppOrigNatLink\hyper@natlinkstart
\let\suppOrigNatAnchor\hyper@natanchorstart
\renewcommand*{\hyper@natlinkstart}[1]{\suppOrigNatLink{supp.#1}}
\renewcommand*{\hyper@natanchorstart}[1]{\suppOrigNatAnchor{supp.#1}}
\makeatother

\clearpage
\setcounter{page}{1}
\maketitlesupplementary

\appendix
\crefalias{section}{appendix}
\crefalias{subsection}{appendix}
\crefalias{subsubsection}{appendix}
\renewcommand{\thetable}{S\arabic{table}}
\renewcommand{\thefigure}{S\arabic{figure}}
\setcounter{table}{0}\setcounter{figure}{0}
\etocsetlocaltop.toc{part}
\etocsetnexttocdepth{2}
\etocsettocstyle{\noindent\textbf{Contents}\par\small}{\par\medskip}
\localtableofcontents

\section{Benchmark construction}
\label{supp:benchmark_construction}

\subsection{Source Datasets and Selection}
\label{supp:source_datasets}
Ego-METAS combines three existing multimodal egocentric datasets: Ego-Exo4D~\cite{grauman2024ego}, CMU-MMAC~\cite{de2009guide}, and CaptainCook4D~\cite{peddi2024captaincook4d}. We describe below the source datasets and the selection of recordings used in our benchmark.

\paragraph{Ego-Exo4D.}
Ego-Exo4D is a multimodal egocentric dataset (RGB, audio, IMU, monochrome video, and gaze, with synchronized exocentric views) including videos of procedural activities and related keystep annotations. {Ego-Exo4D also defines a benchmark for energy-efficient online multimodal keystep recognition, which builds on its keystep recognition benchmark. We start from the keystep recognition annotations, revise them, and apply a re-annotation pipeline to adapt them to our requirements (\cref{supp:reannotation_egoexo}).}
Due to insufficient data, we excluded the scenarios ``Cooking Pasta'' and ``Cooking Sushi Rolls''. The former contained only 4 videos (3 required in the training set to cover all classes and 1 remaining for testing, leaving no validation set), while the latter consisted of a single video and could not be split across training, validation, and test sets. Removing these 5 videos, along with those missing audio or IMU data, resulted in a final dataset of 773 videos. These remaining videos—split into 493 training, 115 validation, and 165 test videos—are reported in \cref{tab:scenario_detailed_stats}.
The resulting dataset covers 15 scenarios and a total duration of 66.4 hours. See \cref{supp:dataset_stats} for splits and detailed statistics.

\paragraph{CMU-MMAC.}
We obtained CMU-MMAC from the official project website and use the annotations of EgoProceL~\cite{bansal2022myview} with the splits of FACT~\cite{lu2024fact}.
For final data preparation, we isolated the egocentric videos for this study, excluding the third-person perspectives as they contain redundant information and deviate from a true wearable device setting. Videos lacking audio or IMU data were also removed.
The resulting dataset covers five recipes and a total duration of 14.4 hours. See \cref{supp:dataset_stats} for splits and detailed statistics.

\paragraph{CaptainCook4D.}
This dataset was taken from the official webpage of \cite{peddi2024captaincook4d}.
We keep only the recordings that {contain} RGB video, audio and IMU information. Therefore, a considerable amount of videos were discarded.
RGB video and audio data were acquired from a GoPro camera system, whereas IMU data were collected with a HoloLens. \cref{supp:synchronization} describes how we synchronize them.
The resulting dataset contains 78 videos across 21 recipes, for a total duration of {23.4 hours}. See \cref{supp:dataset_stats} for detailed statistics.

\subsection{Related benchmarks and datasets}
\label{supp:related_datasets}

\cref{tab:supp_related_datasets2} places Ego-METAS next to related datasets and to the energy-efficient keystep recognition benchmark of Ego-Exo4D~\cite{grauman2024ego}. We list the egocentric datasets with procedural activities and step or action labels, the datasets that are not fully procedural but are widely used in egocentric vision, and the third-person or mixed-view procedural datasets that serve as standard benchmarks for temporal action segmentation. Each cell reports what the source itself states. Properties that a source does not mention are marked as absent.

The closest setting is the Ego-Exo4D energy keystep task: it evaluates online keystep recognition on streaming episodes under budgets of 20\,mW and 2.8\,W, with RGB and audio from one capture device, and scores per-frame mcAP. Ego-METAS differs in the label sets (keysteps merged per scenario, \cref{supp:reannotation}), the metrics (Acc, Edit, F1@25 and power), the sources (three datasets and three capture platforms), the sensors (IMU, monochrome and gaze next to RGB and audio, all aligned to a common 30 FPS timeline) and the analytical energy model (\cref{supp:energy}). Other datasets are richer in other respects: HoloAssist and CaptainCook4D add depth and hand or head pose, Ego4D and Ego-Exo4D are larger, and Assembly101 and Breakfast come with established temporal action segmentation benchmarks. None of the listed datasets defines a power budget besides the Ego-Exo4D energy keystep task.

\begin{table*}[t]                       
        \centering
        \caption{Ego-METAS and related datasets and benchmarks. \cmark{}: stated by the source; \xmark{}: not listed by 
        the 
        source; $\circ$: partial (subset or third-party only). \emph{Ego}: egocentric streams in the release ($^+$:         exocentric views are also released). \emph{Proc.}: procedural (task-oriented, ordered steps toward a goal).  
        \emph{TAS}: native step or action labels with start and end times in the official release. \emph{Scale}:
        recipes,   
        tasks, goals or scenarios as named by the source. \emph{Mono}: monochrome SLAM camera. Hours as stated by each source, except for datasets marked $^a$, for which we report only the hours of egocentric video although they also release other views ($^+$). 
        $^a$Egocentric hours only. $^b$Partial in the official release; full 5-recipe labels derived from EgoProceL.
        $^c$Original labels, which Ego-METAS merges and splits anew (\cref{supp:reannotation}). $^d$Hours of step labels in a 3670\,h collection. $^e$27.6K ego segments (87h). 
        $^f$Ego-Exo4D only.
        }
        \label{tab:supp_related_datasets2}
        \scriptsize
        \setlength{\tabcolsep}{4pt}
        \begin{tabular}{@{}lrlcccccccc@{}}             
        \toprule                        
         & Hours & Scale & Ego & Proc. & TAS & RGB & Aud. & IMU & Gaze & Mono\\ 
        \midrule
        \multicolumn{11}{@{}l}{\textit{Procedural activities}}\\                  EgoProceL~\cite{bansal2022myview} & 62 & 16 tasks & \cmark & \cmark & \cmark & \cmark &
        $\circ$ & 
        $\circ$ & $\circ$ & \xmark\\                  
        EgoPER~\cite{lee2024egoper} & 28 & 5 tasks & \cmark & \cmark & \cmark & \cmark & \cmark & \xmark & \cmark &     
        \xmark\\
        EGTEA Gaze+~\cite{li2021eye} & 28 & 7 recipes & \cmark & \cmark & \cmark & \cmark & \cmark & \xmark & \cmark &  
        \xmark\\
        CMU-MMAC~\cite{de2009guide} & 16.4$^a$ & 5 recipes & \cmark$^+$ & \cmark &   
        $\circ$$^b$ & \cmark & \cmark & \cmark & \xmark & \xmark\\ CaptainCook4D~\cite{peddi2024captaincook4d} & {94.5} & {24 recipes} & \cmark & \cmark & \cmark$^c$ & \cmark &   
        \cmark                          
        & \cmark & \xmark & \xmark\\    HoloAssist~\cite{wang2023holoassist} & 166 & 20 tasks & \cmark & \cmark & \cmark & \cmark & \cmark & \cmark &   
        \cmark                         
        & \xmark\\
        HD-EPIC~\cite{perrett2025hd} & 41.3 & 69 recipes & \cmark & \cmark & \cmark & \cmark & \cmark & \cmark & \cmark & \cmark\\
        Ego4D Goal-Step~\cite{song2023ego4dgoalstep} & 430$^d$ & 86 goals & \cmark & \cmark & \cmark & \cmark & $\circ$ 
        &   
        $\circ$ & $\circ$ & \xmark\\
        Assembly101~\cite{sener2022assembly101} & 167$^a$ & 101 toys & \cmark$^+$ & \cmark & \cmark & \cmark & \xmark &     
        \xmark &                        
        \xmark & \cmark\\               Breakfast~\cite{kuehne2014language} & 77 & 10 activities & \xmark & \cmark & \cmark & \cmark &
        \xmark & \xmark & \xmark & \xmark\\
        50Salads~\cite{stein2013combining} & $\sim$4 & 1 task & \xmark & \cmark & \cmark & \cmark & \xmark & $\circ$ &      
        \xmark & 
        \xmark\\
        \midrule
        \multicolumn{11}{@{}l}{\textit{Not fully procedural}}\\
        EPIC-KITCHENS-100~\cite{damen2022rescaling} & 100 & -- & \cmark & \xmark & \cmark & \cmark & \cmark & $\circ$ &   
        \xmark & \xmark\\
        Ego4D~\cite{grauman2022ego4d} & 3670 & -- & \cmark & $\circ$ & $\circ$ & \cmark & $\circ$ & $\circ$ & $\circ$ & \xmark\\
        Ego-Exo4D (full)~\cite{grauman2024ego} & 221.3$^a$ & 43 tasks & \cmark$^+$ & $\circ$ & $\circ$$^c$ & \cmark & \cmark & \cmark & \cmark & \cmark\\
        Aria Everyday Activities~\cite{lv2024aea} & {$>$7} & -- & \cmark & \xmark & \xmark & \cmark & \cmark & \cmark 
        &   
        \cmark & \cmark\\
        \midrule
        \multicolumn{11}{@{}l}{\textit{Energy-aware benchmark}}\\
        Ego-Exo4D energy keystep~\cite{grauman2024ego} & 87h segments$^e$ & 17 scenarios & 
           
        \cmark & \cmark & \cmark & \cmark & \cmark & \xmark & \xmark & \xmark\\             
        \midrule
        \multicolumn{11}{@{}l}{\textit{Ours}}\\
        \textbf{Ego-METAS} & {104.2} & 41 scenarios & \cmark & \cmark & \cmark & \cmark & \cmark & \cmark & $\circ$$^f$ 
        & 
        $\circ$$^f$\\
        \bottomrule
        \end{tabular}
    \end{table*}

\subsection{Re-annotation and Label Processing}
\label{supp:reannotation}

Ego-METAS builds upon the initial Energy-efficient multimodal keystep recognition benchmark introduced in Ego-Exo4D~\cite{grauman2024ego}. The provided annotations present several important limitations for online Temporal Action Segmentation (TAS). First, the initial benchmark utilizes a trimmed subset of annotations extracted from the Keystep benchmark, excluding a substantial number of segments, particularly those belonging to background class. While this design choice may be reasonable for keystep prediction in pre-segmented clips, in METAS we aim to evaluate continuous perception in long clips, requiring models to handle long background transitions between keysteps. For long procedural videos in the wild, it is crucial to learn to distinguish between true background and foreground, which we found a profound limitation of the existing annotations.
Second, we observed that the original dataset has some inconsistencies in the annotations, including duplicated classes (see \cref{fig:dup-classes}). Similar problems were observed for CaptainCook4D. Taking all these into account, motivated us to perform the re-annotation of the datasets for Ego-METAS, which we describe here:

\begin{figure}
    \centering
    \includegraphics[width=1.0\linewidth]{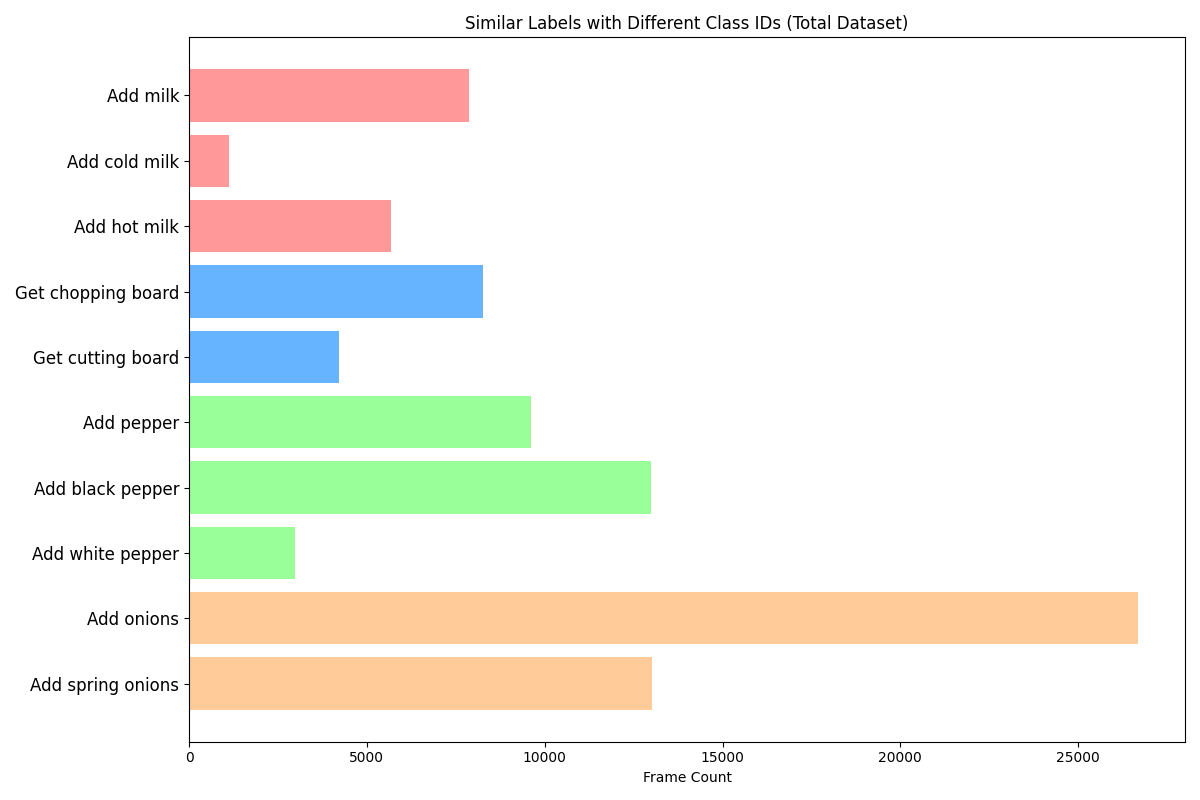}
    \caption{Examples of duplicated classes}
    \label{fig:dup-classes}
\end{figure}

\begin{figure}
  \centering
    \centering
    \includegraphics[width=1.0\linewidth]{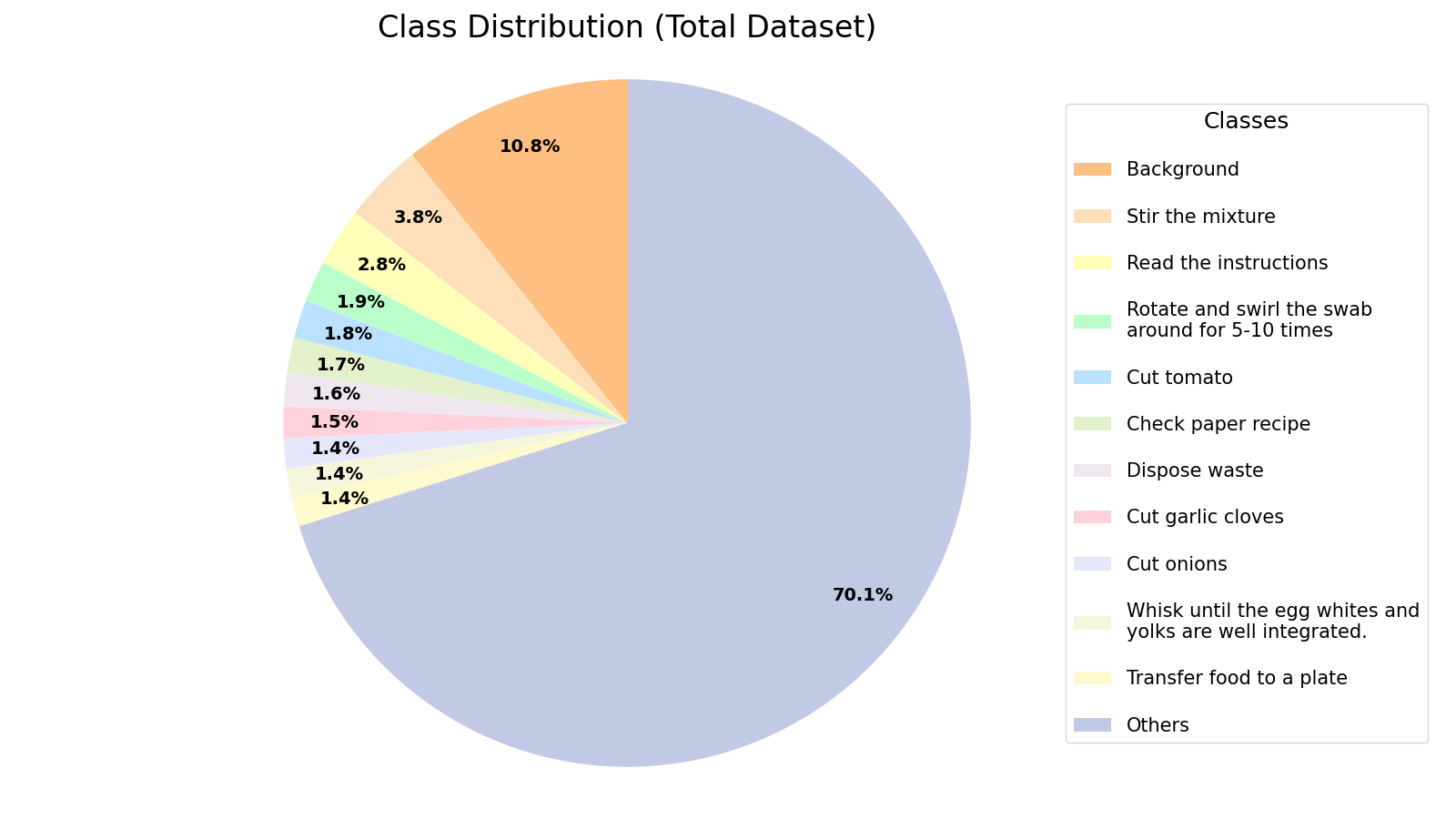}
    \caption{Original Dataset Labels}
    \label{fig:egoexo4d-617-stats}
\end{figure}

\paragraph{Ego-Exo4D.} 
\label{supp:reannotation_egoexo}
We first performed a detailed analysis of the structure of the dataset, including number of classes, distribution, frequency of appearance, and segment length. We covered both the full annotations and the subset of classes for the Keystep recognition benchmark. 
Following prior works~\cite{shen2024progress}, we determined the most appropriate strategy for this benchmark was to train a model per scenario and average the results to obtain the global dataset metrics.
Each model thus only handles the keysteps of one scenario, which reduces the risk of collapsing to the \textit{``Background''} class.

Even opting for this training strategy, the class granularity was too fine for the limited number of videos per scenario, presenting extreme long-tail variability across its 617 fine-grained keysteps (after pruning videos lacking all modalities). Many classes were semantically redundant, and \textit{``Background''} remained the dominant class for the majority of frames (as illustrated in \cref{fig:egoexo4d-617-stats}), effectively making this recognition task excessively difficult. In the original annotations, only one scenario exhibited sufficient balance between class counts and data samples (``First Aid - CPR'').

To address these challenges, we established a systematic class consolidation and filtering pipeline. We first extracted empirical statistics per class (measuring instance frequency, segment duration distributions, and video coverage per scenario). Given the large number of initial classes, we fed this statistical report along with semantic descriptions into an LLM~\cite{team2023gemini} to seed initial cluster proposals based on frequency of appearance and semantic coherence. These seed groupings were subsequently reviewed by 5 annotators, who gave feedback and iterated on the LLM proposals until none had further objections. The prompt used is the following: ``Given this current taxonomy [taxonomy of scenario] and stats per action [number of steps and \% over whole scenario per action], generate a CSV mapping that merges 'Splintered' and 'Invisible' classes into a consolidated list (e.g., mapping 40 classes down to a robust 18–20) to achieve a more balanced class distribution.''

In sum, this re-annotation stage comprised two complementary operations:
\begin{enumerate}
    \item \textbf{Semantic consolidation}: Reconciling inconsistent and redundant actions by grouping fine-grained synonyms into coherent procedural steps (e.g., merging \textit{``Add milk''} with \textit{``Add cold milk''}, or \textit{``Get chopping board''} with \textit{``Get cutting board''}) (see \cref{fig:dup-classes}).
    \item \textbf{Background mapping}: Merging unlearnable, sporadic, or task-inconsistent classes into \textit{``Background''} (52 classes across all scenarios).
\end{enumerate}
The decision to map these 52 classes to \textit{``Background''} followed strict empirical and procedural criteria:

\begin{itemize}[leftmargin=*,itemsep=2pt,topsep=2pt]
    \item \textbf{Low-frequency and tail-occurrence filtering}: 
    Our empirical duration and occurrence reports revealed that several classes appeared only 1--3 times across an entire scenario's video collection. Such severe data starvation precludes robust learning and evaluation; these rare outliers were therefore mapped to \textit{``Background''}.
    
    \item \textbf{Incompatible testing modalities and transient steps}: 
    For example, in the Covid-19 test (12 classes mapped to \textit{``Background''}), the dataset features standard nasal-swab test kits. A few outlier recordings utilized alternate saliva-based kits containing actions such as \textit{``Collect saliva sample''}, \textit{``Spit into the testing tube''}, and \textit{``Slowly extract the test tube from the mouth''}. These were filtered out because saliva collection is an incompatible test modality that does not occur during nasal swabbing. In addition, preparatory actions such as \textit{``Blow nose''} exhibited near-zero duration (often 1--2 seconds), appeared sporadically, and were omitted by the vast majority of participants. This rationale was utilized in other scenarios.
    \item \textbf{Passive waiting and secondary glances}: 
    Secondary non-manipulation interactions (e.g., glancing at a smartphone, inspecting paper recipes) and passive incubation intervals (e.g., \textit{``Wait for 15 minutes''}, setting external timers) represent inactive waiting periods rather than active physical object manipulations.
\end{itemize}

Subsequently human supervision was needed for final corrections and verifications.
Final reannotations can be found from \cref{tab:new_label_mapping_clean_chain_617} to \cref{tab:new_label_mapping_making_chai_tea_617}.

\paragraph{CaptainCook4D.}
\label{supp:reannotation_captain}
Annotations encompass 324 recipe-specific keysteps distributed across 21 diverse cooking activities.
Similarly to Ego-Exo4D, background class was predominant, representing 18\% of the dataset, again causing the training models to collapse. We therefore apply the same procedure as for Ego-Exo4D: groupings proposed by the LLM and reviewed by the same five annotators until they agreed on every class. 
To facilitate cross-recipe knowledge transfer, all 324 fine-grained recipe steps were clustered into 11 macro procedural classes, where the \textit{``Background''} class was reserved for the original  \textit{``Background''} class.
The new annotations can be found at \cref{tab:class_mapping_captain}.

\subsection{Temporal Synchronization and Preprocessing}
\label{supp:synchronization}

Because none of the original datasets were natively prepared for real-time energy-aware streaming, we implemented a multi-stage temporal synchronization process across modalities. All sensor streams were aligned to the common 30 FPS reference timeline used in Ego-METAS.

\paragraph{Ego-Exo4D.}
We aligned raw Meta Aria sensor telemetry (IMU, SLAM monochrome cameras, gaze) with the released audio and video streams. High-frequency modalities (IMU, gaze) were mapped to the primary temporal grid using timestamp-based nearest-neighbor interpolation.

\paragraph{CMU-MMAC.}
Since the audio and IMU streams were not originally synchronized with the video and annotations, we aligned them following the methodology of \cite{nakamura2021sensor}, applying the precise temporal offsets established in that work to synchronize the IMU, RGB video, and audio streams.

\paragraph{CaptainCook4D.}
As GoPro (RGB and audio) and HoloLens (IMU) streams lacked hardware clock synchronization, we estimated an initial offset via classical visual feature matching between GoPro and HoloLens video feeds, followed by frame-level manual verification and refinement.

\subsection{Pre-extracted features}
\label{supp:mod_computation}

\noindent
For each modality, we extract the feature as follows:

\paragraph{RGB.} We employ DINOv3 \cite{simeoni2025dinov3} (ViT-B) as feature extractor, which as default takes $448\times 448$ input images at each timestep $t$. Feature extraction is applied at each timestep $t$ in the RGB video stream as $\mathbf{x}_t^{(RGB)}=\phi_{RGB}(\mathbf{X}_t^{(RGB)})$. {For the energy model, we count the multiply-accumulates (MACs) of the extractor with \texttt{thop} on the official DINOv3 implementation.}

\paragraph{Monochrome.} 
We use the same extractor as for RGB, \ie DINOv3 (ViT-B), applied to the monochrome camera stream, but with low-resolution $256\times 256$ input images, to serve as an energy-efficient alternative modality: $\mathbf{x}_t^{(Mono)}=\phi_{Mono}(\mathbf{X}_t^{(Mono)})$.

\paragraph{IMU.} We encode six channels of one IMU (triaxial accelerometer in m/s$^2$ and gyroscope in rad/s) with PRIMUS~\cite{das2025primus}, pre-trained on 5\,s windows of the head-mounted IMU of Ego-Exo4D.
At each step $t$, we take the 150 samples of the 30\,Hz signal before $t$ (5\,s, zero-padded at the start of the video), resample them linearly to 250 samples, the input length of PRIMUS, and keep its 512-dimensional embedding.
We use the left IMU of the Aria glasses in Ego-Exo4D, the IMU of the HoloLens~2 headset in CaptainCook4D, and a wired IMU on the right arm in CMU-MMAC.
Unlike the other two, the CMU-MMAC IMU is worn on the body rather than on the head, as in the pre-training data of PRIMUS.

\paragraph{Audio.} We use SSAST \cite{gong2022ssast} (Small variant, 384-d embeddings) audio features as follows: for each timestep $t$, we take a clip segment corresponding to the last second of audio up to $t$. This clip is converted into a Mel spectrogram, which is the required input for the SSAST backbone: $\mathbf{x}_t^{(Audio)}=\phi_{Audio}(\text{spectrogram}(\mathbf{X}_t^{(Audio)}))$. We use the SSAST version where the spectrogram is split into a sequence of 16 × 16 patches.  Notice that, while SSAST is typically pre-trained on 10-second clips, we reduce clip length down to just 1 second to reduce compute and memory, and allow the policy to save energy in a more granular way.

\paragraph{Gaze.} For gaze modality, instead of using the 2D on the image location as a signal, we cropped a $192\times 192$ bounding box around the 2D gaze position. That allows us to use local visual information which could be relevant for action recognition since this crop represents the region of the image the user is interacting with or the user’s interest. This crop is also encoded with DINOv3. {The crop is taken from the RGB image; if the gaze is close to the border, we shift the box inward so that it keeps its size, without padding.}

\noindent Energy estimates, including sensor reads and feature extraction are in \cref{tab:supp_per_read}.

\subsection{Dataset Splits and Statistics}
\label{supp:dataset_stats}
\label{supp:data_stats}
\begin{figure*}[hp]
  \centering
  \includegraphics[width=\linewidth]{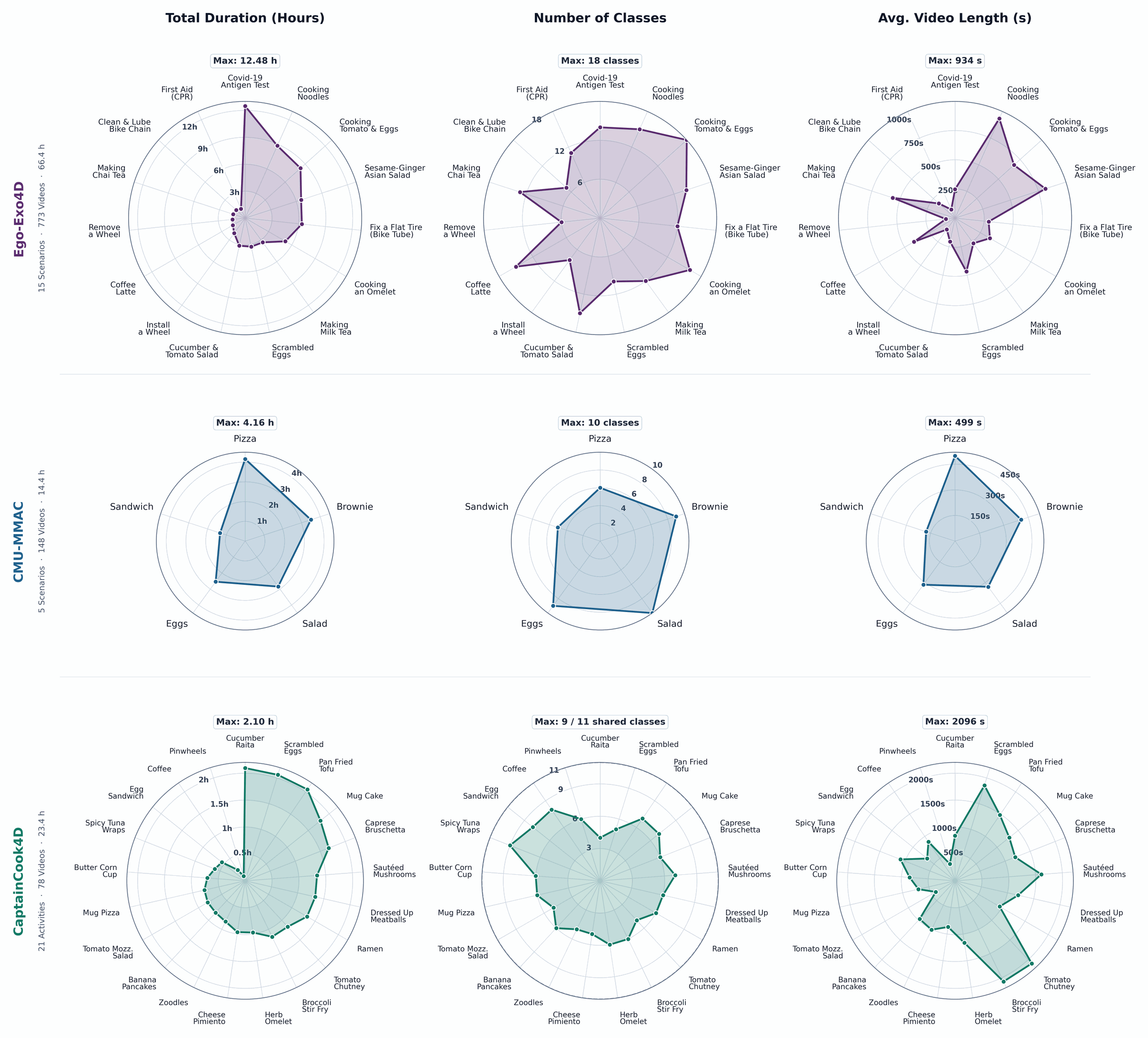}
  \caption{\textbf{Statistics of each scenario in the three datasets.} Each row is a dataset (Ego-Exo4D, CMU-MMAC, CaptainCook4D) and each polygon has one vertex per scenario. Columns report the total duration (hours), the number of action classes, and the average video length (seconds) of the scenario. The outer ring of each polygon is the maximum value over the scenarios of that dataset (printed in the panel title), so the axes are not comparable across panels.}
  \label{fig:polygon}
\end{figure*}

In this section, we outline the final details and statistics for the released multi-dataset. 
\cref{fig:polygon} details the total duration (in hours), number of classes, and average video length of every scenario for all three datasets after the re-annotation process (\cref{supp:reannotation}). Below we describe the splits and statistics of each dataset.

\paragraph{Ego-Exo4D.} Since the test set annotations from Ego-Exo4D were never publicly released, we decided to use the validation set from their benchmark as our test set and split in a 80-20\% the training set to obtain a validation set.
Furthermore, our analysis revealed that, for some videos, the actors were narrating in different languages the steps of the procedure, and other videos have loud music in the background preventing the audio from being informative. Although initial results did not show significant impact in the results, we have documented the names in which participants narrate the steps (see \cref{tab:scenario_coverage}). To ensure the narrations do not help models to actually predict the correct steps, we made sure that none of these videos are on the test set by swapping the compromised ones with another uncompromised from another split. In addition, we enforce that the training set has all available classes in that split.

\begin{table}[t]
\centering
\caption{Actor narration per scenario.}
\resizebox{\columnwidth}{!}{%
\begin{tabular}{lccc}
\toprule
\textbf{Scenario} & \textbf{Found} & \textbf{Total} & \textbf{Percentage (\%)} \\
\hline
First Aid - CPR & 26 & 51 & 50.98 \\
Clean and Lubricate the Chain & 6 & 26 & 23.08 \\
Install a Wheel & 9 & 62 & 14.52 \\
Remove a Wheel & 12 & 64 & 18.75 \\
Fix a Flat Tire & 24 & 79 & 30.38 \\
Covid-19 Rapid Antigen Test & 29 & 182 & 15.93 \\
Cooking Noodles & 7 & 34 & 20.59 \\
Making Coffee latte & 2 & 14 & 14.29 \\
Making Cucumber \& Tomato Salad & 14 & 55 & 25.45 \\
Cooking Scrambled Eggs & 7 & 25 & 28.00 \\
Cooking an Omelet & 9 & 54 & 16.67 \\
Making Milk Tea & 16 & 45 & 35.56 \\
Making Sesame-Ginger Asian Salad & 7 & 29 & 24.14 \\
Cooking Tomato \& Eggs & 5 & 44 & 11.36 \\
Making Chai Tea & 2 & 9 & 22.22 \\
\bottomrule
\end{tabular}%
}
\label{tab:scenario_coverage}
\end{table}

The shortest video has 589 frames (19.6 seconds) and the largest has 62,392 frames (34 minutes, and 39 seconds), yielding a total duration of 66 hours, 24 minutes, and 43.8 seconds. See \cref{tab:scenario_detailed_stats} for additional per-scenario statistics, including global and independent statistics per split.
Regarding number of classes, the dataset has an average of 12.4 classes per scenario. \cref{tab:cls_stats_Covid-19_Rapid_Antigen_Test,tab:cls_stats_Clean_and_Lubricate_the_Chain,tab:cls_stats_Cooking_Scrambled_Eggs,tab:cls_stats_Cooking_Noodles,tab:cls_stats_Cooking_an_Omelet,tab:cls_stats_Cooking_Tomato_and_Eggs,tab:cls_stats_Fix_a_Flat_Tire_-_Replace_a_Bike_Tube,tab:cls_stats_Install_a_Wheel,tab:cls_stats_Making_Chai_Tea,tab:cls_stats_Making_Coffee_latte,tab:cls_stats_Making_Cucumber_and_Tomato_Salad,tab:cls_stats_Making_Milk_Tea,tab:cls_stats_Making_Sesame-Ginger_Asian_Salad,tab:cls_stats_Remove_a_Wheel,tab:cls_stats_First_Aid_-_CPR} show additional per-class statistics for all scenarios independently.

\paragraph{CMU-MMAC.} We built upon the framework provided by \cite{lu2024fact}, further partitioning their training set into a 80-20\% split to create a validation set. See \cref{tab:cmu_stats} for additional per-scenario statistics, including splits.

Videos lacking audio or IMU data were also removed, resulting in a total dataset duration of 14 hours, 22 minutes, and 48 seconds. Regarding video duration, shortest video has 99.1 seconds, whereas longer one has 850.2 seconds, yielding generally shorter video durations compared to Ego-Exo4D. Moreover, average number of classes per scenario is 7.8. See \cref{tab:classes_1,tab:classes_2,tab:classes_3,tab:classes_4,tab:classes_5} for further per-class statistics for all scenarios independently.

\paragraph{CaptainCook4D.} After removing the videos without audio or IMU, none of the three official splits has enough videos for training and evaluation (see \cref{tab:split_comparison}).
For that reason, we do not train on a per-scenario basis, and instead the full dataset is trained jointly.
Thus, we create a new split for this benchmark (\cref{tab:captain_stats}), following a greedy class-coverage criterion applied. Specifically, we first iteratively select videos that maximize the number of covered recipe-level step classes until 100\% of all classes across all recipes are present in the training set; this assigns one video per recipe (21 videos) to training. Next, we prioritize test-set coverage by greedily selecting videos that add uncovered classes, which assigns one test video to every recipe with at least two recordings. Finally, all remaining recordings are assigned to the validation set. The resulting split contains 21, 38, and 19 videos (26.9\%, 48.7\%, and 24.4\% of the 78 videos) for training, validation, and test, corresponding to 6.4, 10.1, and 6.8 hours (27.6\%, 43.2\%, and 29.2\% of the 23.4-hour total duration). Because few recordings remain per recipe after filtering out videos without synchronized audio or IMU (1 to 9 videos per recipe), the split cannot be uniformly stratified: \textit{Coffee} and \textit{Pinwheels} have a single video, which is assigned to training, while \textit{Tomato Chutney}, \textit{Broccoli Stir Fry}, and \textit{Spicy Tuna Avocado Wraps} have two videos each (one for training and one for test, with no validation video). We prefer this split to the official ones (\cref{tab:split_comparison}) because all three official splits leave one or two recipes with zero training videos, whereas our split guarantees that every recipe and all 11 action classes are seen during training, every recipe with at least two videos (19/21) is evaluated at test time, and all 11 action classes are covered across training, validation, and test.

As shown in \cref{tab:captain_stats}, the CaptainCook4D subset yields a total of 78 videos and {23.4 hours}, where shortest video lasts 246.9 seconds, whereas the longest is 2467.9 seconds long (41 minutes and 8 seconds), longer than the longest Ego-Exo4D video (34 minutes and 39 seconds). Regarding number of classes, CaptainCook4D has 6.1 unique classes per activity.

\begin{table*}[t]
\centering
\caption{Ego-Exo4D \cite{grauman2024ego} per scenario statistics}
\resizebox{\textwidth}{!}{
\begin{tabular}{lccccccccc}
\toprule
& \multicolumn{3}{c}{N. Videos (Narrations) / duration (min)}& & &\\
\cmidrule(lr){2-4}
Scenario & Train & Val & Test & Classes & Total Time (h) & Avg Len (s) & Min Len (s) & Max Len (s) \\
\midrule
Covid-19 Rapid Antigen Test & 110 (16)/443.20 & 27 (13)/113.96 & 45 (0)/191.56 & 14 & 12.48 & 246.8 $\pm$ 83.2 & 78.6 & 594.3 \\
Clean and Lubricate the Chain & 17 (5)/57.88 & 4 (1)/9.40 & 5 (0)/13.77 & 7 & 1.35 & 187.0 $\pm$ 73.9 & 57.8 & 366.8 \\
Cooking Scrambled Eggs & 16 (5)/112.65 & 3 (2)/31.24 & 6 (0)/51.36 & 10 & 3.25 & 468.6 $\pm$ 244.0 & 62.6 & 1000.6 \\
Cooking Noodles & 22 (3)/324.94 & 5 (4)/91.43 & 7 (0)/112.91 & 15 & 8.82 & 934.0 $\pm$ 407.3 & 333.1 & 1704.2 \\
Cooking an Omelet & 35 (6)/219.40 & 8 (3)/35.64 & 11 (0)/55.68 & 16 & 5.18 & 345.2 $\pm$ 231.2 & 109.8 & 1245.0 \\
Cooking Tomato \& Eggs & 28 (5)/330.92 & 7 (0)/72.66 & 9 (0)/94.97 & 18 & 8.31 & 679.8 $\pm$ 411.2 & 43.1 & 2079.7 \\
Fix a Flat Tire - Replace a Bike Tube & 55 (16)/269.38 & 13 (8)/64.21 & 11 (0)/47.91 & 12 & 6.36 & 289.7 $\pm$ 137.3 & 28.3 & 669.2 \\
Install a Wheel & 41 (8)/91.33 & 10 (1)/15.06 & 11 (0)/18.30 & 8 & 2.08 & 120.7 $\pm$ 103.1 & 24.3 & 621.6 \\
Making Chai Tea & 6 (2)/57.42 & 1 (0)/8.84 & 2 (0)/18.14 & 13 & 1.41 & 562.7 $\pm$ 135.4 & 441.8 & 922.2 \\
Making Coffee latte & 10 (1)/69.46 & 2 (1)/15.53 & 2 (0)/9.86 & 15 & 1.58 & 406.5 $\pm$ 227.0 & 29.1 & 987.9 \\
Making Cucumber \& Tomato Salad & 34 (8)/109.89 & 8 (6)/35.35 & 13 (0)/43.61 & 15 & 3.15 & 206.0 $\pm$ 138.4 & 37.2 & 509.6 \\
Making Milk Tea & 27 (10)/131.28 & 6 (6)/27.25 & 12 (0)/41.95 & 12 & 3.34 & 267.3 $\pm$ 173.1 & 74.8 & 732.5 \\
Making Sesame-Ginger Asian Salad & 18 (6)/240.90 & 4 (1)/50.31 & 7 (0)/102.49 & 14 & 6.56 & 814.6 $\pm$ 340.4 & 207.1 & 1383.3 \\
Remove a Wheel & 43 (9)/59.14 & 10 (3)/11.21 & 11 (0)/14.59 & 6 & 1.42 & 79.6 $\pm$ 62.0 & 19.6 & 322.6 \\
First Aid - CPR & 31 (19)/44.00 & 7 (7)/9.39 & 13 (0)/14.57 & 11 & 1.13 & 80.0 $\pm$ 23.4 & 45.5 & 184.9 \\
Total & 493 (119) & 115 (56) & 165 (0) & 186 & 66.42 & & & & \\
\bottomrule
\end{tabular}
}
\label{tab:scenario_detailed_stats}
\end{table*}

\subsection{Evaluation Metrics and Limitations of mcAP}
\label{supp:benchmark_inconsistencies}

The main evaluation metric chosen in the previous Ego-Exo4D energy-efficient benchmark~\cite{grauman2024ego} was the mean calibrated Average Precision (mcAP), proposed initially by \citet{de2016online}. While commonly used in online action detection, it is suboptimal for online TAS, for the reasons below.

For a given action class, the calibrated Average Precision (cAP) is defined as
\begin{equation}
\label{eq:cap_supp}
\mathrm{cAP} = \frac{1}{P} \sum_k \frac{w \cdot TP(k)}{w \cdot TP(k) + FP(k)} I(k),
\end{equation}
with
\begin{equation}
w = \frac{\text{\# negative frames}}{\text{\# positive frames}},
\end{equation}
where $TP(k)$ and $FP(k)$ are the cumulative true and false positives at rank $k$, $I(k)$ is an indicator function equal to 1 if the $k$-th prediction is a true positive and 0 otherwise, and $P$ is the total number of positive frames in the dataset. Notice that it can be rewritten as
\begin{equation}
\label{eq:cap_rewritten_supp}
\mathrm{cAP} = \frac{1}{P} \sum_k \frac{TP(k)}{TP(k) + \frac{FP(k)}{w}} I(k).
\end{equation}

As the proportion of background frames (negative frames) increases, $w$ grows accordingly, so that even a large number of false positives, divided by $w$, barely lowers the score.
Consequently, the reported performance in datasets with lots of background, such as Ego-Exo4D, can be artificially inflated, potentially approaching 100\%, despite limited true recognition capability.
For example, for a keystep that covers 1\% of the frames of a video, a model that detects half of these frames and raises false alarms on 1\% of the remaining ones has an actual precision of 34\% (two of every three detections are false alarms), but a calibrated precision of 98\%.

When training on the full dataset, models tend to collapse to predict the background class, as this circumstance produces deceptively high metric scores while failing to capture relevant actions. This issue becomes more pronounced when training across all scenarios. As shown in the original dataset label distribution (\cref{fig:egoexo4d-617-stats}), the background class dominates the data distribution, clearly separating itself from the remaining classes. As a result, the metric is biased toward background predictions, masking the model's actual ability to recognize meaningful actions.

To address this issue, our benchmark uses other metrics as primary metrics: Frame Accuracy (Acc), segmental Edit Distance (Edit), and segmental F1 scores.

\subsection{Release, Licences, and Personal Data}
\label{supp:release_licences}
Ego-METAS re-uses three existing datasets and does not include new recordings: consent and ethics approval are those of the original datasets. We release the annotations, the splits, the pre-extracted features, the energy model, and the scripts on Hugging Face. In respect of the individual dataset licences, we do not redistribute the source videos or audio, which contain faces and speech of the participants. Instead, the codebase documents how to obtain them from the original sources under their licences, and provides the splits and the metadata needed to align the raw streams with our annotations and features.

\FloatBarrier
\section{Energy model}
\label{supp:energy}

This section describes the energy model behind the power values of the paper (\cref{eq:energy_total}). We first explain why the model is analytical and not tied to a specific device (\cref{supp:energy_scope}). We then give the technical details of how we count MAC operations and memory and which capture powers we use (\cref{supp:energy_budget}). Finally, we report the constants of the two energy models used in the paper, \emph{reference} and \emph{edge-SoC}, and the energy of one read of each sensor under both models (\cref{supp:energy_grounding}).

\subsection{Design choices and scope}
\label{supp:energy_scope}
{The energy model is analytical and, by design, not tied to a specific device. We make this choice for four reasons. First, the per-operation energy of commercial wearable processors is rarely published, and their hardware is closed (e.g., the product page of the Snapdragon AR1 Gen~1 describes its process node only as ``advanced''~\cite{qualcomm2023ar1}), so a device-specific model could not be always checked or reproduced. Second, a benchmark tied to one commercial device would depend on a product that changes every few years and would bind the community to its vendor. Third, an analytical model only needs the profiled MAC and memory counts and a few constants: it runs without the target hardware, and every reported power can be recomputed. Fourth, the benchmark compares sensing schedules and models at the same budget; these trade-offs, rather than the absolute power of a given product, are what the energy model must capture. Since $\alpha$, $\beta$, and $P_m$ are parameters of \cref{eq:energy_total}, the benchmark can be retargeted to a specific device by replacing them, without re-profiling or re-training the models, and the main conclusions hold under both the reference and the edge-SoC constants on Ego-Exo4D (\cref{sec:exp_edge}, \cref{supp:add_edge}). The model also makes three simplifications, which affect mainly the absolute power of every schedule. It does not include static power (leakage, idle sensors, and the always-on parts of the device), nor the image signal processing between the sensor and the feature extractor. In addition, the memory term sums the memory allocated by each operation of a forward pass, as reported by the profiler (mostly activations). It is therefore a proxy for DRAM traffic: it does not include the reads of the model weights, and it counts activations that an accelerator may keep on chip. Finally, features are pre-extracted with frozen extractors, and feature extraction dominates the cost of a read (309.5 of 310.0\,mJ for RGB, \cref{tab:supp_per_read}), so a lighter extractor would change it; jointly optimizing the sensing policy and the extractor is left to future work.}

\subsection{Energy computation and profiling}
\label{supp:energy_budget}

For convenience, we report \cref{eq:energy_total} from the main paper below:
\[
E_{\mathcal{V}} = \sum_{t=1}^{T}\left(\Delta\tau \sum_{m:(m,t)\in\sigma} P_m + \alpha F_t^{\mathrm{MAC}} + \beta F_t^{\mathrm{MEM}}\right).
\]
\cref{eq:energy_total} separates quantities that we count from constants that describe the hardware. The MAC counts $F_t^{\mathrm{MAC}}$ depend only on the model and the input length, and the memory counts $F_t^{\mathrm{MEM}}$ depend on the model and, to a lesser extent, on its implementation (operator fusion, memory allocator). To estimate them, we profile the forward pass of each model on a desktop GPU: THOP for the MACs and the PyTorch profiler for the memory, as detailed below. We use the GPU only to count operations and bytes, and no energy value comes from its power draw. The counts are therefore independent of the power characteristics of the profiling hardware. The hardware enters only through three groups of constants: the energy per MAC $\alpha$, the energy per byte of memory traffic $\beta$, and the capture power of each sensor $P_m$. The constants $\alpha$ and $\beta$ for the reference and edge-SoC energy models are discussed in \cref{supp:energy_grounding}.
The pre-extracted features run on a 30\,FPS timeline, so $\Delta\tau=1/30$\,s in \cref{eq:energy_total}.

\paragraph{Capture.}
We adopt $P_{\text{RGB}}=15$\,mW and $P_{\text{audio}}=0.5$\,mW from~\cite{grauman2024ego}, and set $P_{\text{mono}} = 1$\,mW~\cite{cottini2011cmos}, $P_{\text{IMU}} = 0.2$\,mW~\cite{mouser_st_lis2du12}, and $P_{\text{gaze}} = 0.63$\,mW~\cite{bonazzi2024retina}. They come from sensor datasheets and published measurements. Since they do not depend on computation, they are the same in both energy models. {These are the powers of a sensor that streams continuously. In \cref{eq:energy_total}, each read is charged the capture energy $P_m\Delta\tau$ with $\Delta\tau=1/30$\,s, whatever the read rate (\cref{tab:supp_per_read}; e.g., 0.50\,mJ per RGB read), so $P_m$ is the capture power of a schedule only when the sensor is read at 30\,FPS. We therefore report capture costs as energy per read.} A gaze read is charged $P_{\text{gaze}}$ only: we assume that it behaves like a read of a lower-resolution sensor, without capturing a full RGB frame. {The gaze feature is extracted from a crop of the RGB image around the gaze point (\cref{supp:mod_computation}); we assume that the crop is not obtained by capturing the full frame and cropping it in software, but by on-sensor processing that extracts the region of interest and transmits only its pixels, so that the transfer of the full frame is not paid}~\cite{gomez2022distributed}.

\paragraph{Computation.} We count MACs with the THOP profiler~\cite{thop}. For each feature extractor $\phi_m$, $f^{\mathrm{MAC}}_{\phi_m}$ is the count of one forward pass. For the TAS models, the MAC count grows linearly with the input length $L$ but does not pass through the origin; at the beginning of training we pass two inputs of different lengths through the model and fit
\[
f^{\mathrm{MAC}}_{\Psi}(L) = m_{\mathrm{frame}}\, L + n_{\mathrm{fixed}},
\]
where $m_{\mathrm{frame}}$ is the cost of each additional step and $n_{\mathrm{fixed}}$ the cost that does not depend on the input length.

\paragraph{Memory.} For the memory term, the PyTorch profiler reports the bytes allocated by each operation of a forward pass, and we sum them over the operations. This count does not scale linearly with the input length, so we profile every video. The memory term is a proxy for DRAM traffic.

\subsection{Reference and edge-SoC constants}
\label{supp:energy_grounding}

The constants $\alpha$ and $\beta$ also set the cost of the feature extractors and therefore dominate the power of most schedules.

{\paragraph{{Reference constants}.} The reference model uses $\alpha=4.6$\,pJ/MAC and $\beta=80$\,pJ/byte, the values adopted by the energy-efficient benchmark of Ego-Exo4D~\cite{grauman2024ego}. The value of $\alpha$ equals the energy of a 32-bit floating-point multiplication (3.7\,pJ) plus a 32-bit floating-point addition (0.9\,pJ) in 45\,nm CMOS~\cite{horowitz2014computing,sze2020evaluate}. It thus describes an unquantized network executed with general-purpose floating-point units, in a technology several node generations older than the processors of current smart glasses, which run quantized networks on dedicated accelerators (e.g., the Snapdragon AR1 Gen~1 of Ray-Ban Meta, built on an advanced process node with a Hexagon NPU~\cite{qualcomm2023ar1}). {We read $\beta$ (10\,pJ/bit) as an energy per byte of DRAM traffic.  It is {half  the energy of the off-chip DRAM in the same 45\,nm table (640\,pJ per 32-bit read, i.e., 160\,pJ per byte)~\cite{horowitz2014computing,sze2020evaluate}} and slightly above that of current low-power mobile DRAM (about 7\,pJ/bit) \cite{pawlowski2019prospects}, so it corresponds to mobile DRAM with a small margin. {On a wearable device, DRAM traffic cannot be avoided: the weights of the feature extractors (e.g., 86M parameters for the ViT-B backbone of DINOv3~\cite{simeoni2025dinov3}, i.e., tens to hundreds of megabytes depending on the numerical precision) exceed the on-chip memory of mobile neural accelerators, so they are read from DRAM at every frame. The memory term approximates this traffic only roughly.}} Both constants are therefore conservative for current wearable devices: {they run quantized networks, and an 8-bit MAC costs 0.23\,pJ (multiplication 0.2\,pJ, addition 0.03\,pJ) already in the same 45\,nm table}~\cite{horowitz2014computing,sze2020evaluate}{, 20 times less than $\alpha$, with further savings at newer technology nodes.} The absolute powers of the reference model are upper estimates for an optimized device, and the 20\,mW and 2.8\,W budgets~\cite{grauman2024ego} are correspondingly strict.}

{\paragraph{Edge-SoC constants.} The second energy model reflects more recent wearable hardware, which runs quantized networks on dedicated neural accelerators built in 4--5\,nm technology (e.g., the Snapdragon AR1 Gen~1 of Ray-Ban Meta~\cite{qualcomm2023ar1}) and stores the model weights in low-power mobile DRAM. We call it the \emph{edge-SoC} model and set $\alpha=0.4$\,pJ/MAC and $\beta=60$\,pJ/byte, while the capture powers $P_m$ do not change. For computation, an 8-bit MAC already costs 0.23\,pJ in 45\,nm technology~\cite{horowitz2014computing,sze2020evaluate}, and less at newer nodes. {We set $\alpha$ to about twice this value. The factor of two is our assumption: it accounts for costs that the model does not count separately, mainly moving data inside the chip, while the arithmetic itself is cheaper at 4--5\,nm than at 45\,nm. The resulting $\alpha$ is about 11 times lower than in the reference model.} For memory, $\beta$ (7.5\,pJ/bit) is close to the energy of LPDDR5 DRAM, about 7\,pJ/bit~\cite{pawlowski2019prospects}, and 25\% lower than in the reference model. Since the profiled MAC and memory counts do not depend on the energy model, switching model only rescales the compute and memory terms of \cref{eq:energy_total}: no model is re-profiled or re-trained, and the accuracy of every schedule is unchanged. The feature extractors become cheaper, so the same budget buys more reads. For instance, reading RGB, audio, and gaze at 6\,FPS costs 2.31\,W under the reference model and 0.58\,W under the edge-SoC model (Table~\ref{tab:all_protas}, \cref{tab:supp_edge_t1}). On Ego-Exo4D, at 20\,mW the budget rule reads gaze at 0.3\,FPS under the reference model (\cref{tab:backbone_training}) and gaze at 0.86\,FPS plus the monochrome camera at 0.18\,FPS under the edge-SoC model (\cref{tab:edge_small}); at 2.8\,W every sensor read at 6\,FPS fits the budget (735\,mW, \cref{tab:supp_edge_t4}). Budgets tighter than 20\,mW thus become meaningful, while 2.8\,W no longer constrains the schedule: at 5\,mW the budget rule reaches 48.7 accuracy on Ego-Exo4D (\cref{tab:edge_small}), 0.1 points below the reference model at 20\,mW (48.8, \cref{tab:backbone_training}). \cref{sec:exp_edge} and \cref{supp:add_edge} repeat the main comparisons under this model.}

\paragraph{Energy per read.} \Cref{tab:supp_per_read} gives the energy of one read of each sensor on Ego-Exo4D, i.e., of capturing one frame and processing it with the feature extractor, under the two models. The capture energy is the same in 
both models; the energy of feature extraction (MACs and memory) is what the model changes. The power of a periodic schedule is approximately the sum, over its sensors, of read rate times energy per read, and it is dominated by feature extraction rather than by sensor capture: a feature extractor runs a large network at every read, while {capturing one frame costs at most 0.50\,mJ (RGB)}. For instance, reading RGB, audio, and gaze at 6\,FPS gives $6\times(310.0+13.8+60.5)$\,mJ/s $\approx 2.31$\,W under the reference model, as in Table~\ref{tab:all_protas}, and capture accounts for only about 3\,mW of it.
\begin{table}[t]
\caption{\textbf{Energy per read} of each sensor on Ego-Exo4D (one read = one captured frame processed by its feature extractor): capture energy ($P_m\Delta\tau$, \cref{supp:energy_budget}), equal under both models, and feature-extraction energy (MACs and memory) under the reference and edge-SoC energy models. The energy of a read is the sum of the two columns.}
\centering
\scriptsize
\begin{tabular}{lccc}\toprule
 & Capture & \multicolumn{2}{c}{Feature extraction}\\\cmidrule(lr){3-4}
Sensor & (mJ) & Reference (mJ) & Edge-SoC (mJ)\\\midrule
RGB & 0.50 & 309.5 & 77.3\\
Audio & 0.02 & 13.8 & 3.2\\
IMU & 0.007 & 0.14 & 0.02\\
Gaze & 0.02 & 60.5 & 15.7\\
Monochrome camera & 0.03 & 102.6 & 25.8\\\bottomrule
\end{tabular}
\label{tab:supp_per_read}
\end{table}

\begin{table*}[t]
\caption{\textbf{Energy of the TAS models and of the policy networks per call} on Ego-Exo4D under the reference and edge-SoC energy models (MAC plus memory energy; the memory term is the profiled bytes times $\beta$). The backbones take all five sensors as input, so their cost is an upper bound for models that take fewer sensors. All values are per streamed step, except the RGB projection and HCMS cascade rows (per RGB read); the Transformer uses a context of 256 steps.}
\label{tab:supp_models}
\centering
\scriptsize
\begin{tabular}{llcccc}\toprule
Component & Called & MACs & Bytes & Reference ($\mu$J) & Edge-SoC ($\mu$J)\\\midrule
ProTAS & per step & $9.7\times10^{5}$ & $8.9\times10^{4}$ & 11.6 & 5.72\\
MiniROAD (GRU) & per step & $1.6\times10^{7}$ & $3.8\times10^{4}$ & {76.7} & 8.70\\
Causal Transformer & per step & $1.6\times10^{6}$ & $1.7\times10^{5}$ & 20.8 & 10.6\\\midrule
Policy, IMU only & per step & $1.1\times10^{5}$ & $1.2\times10^{3}$ & 0.58 & 0.11\\
Policy, closed loop & per step & $7.4\times10^{4}$ & $1.2\times10^{3}$ & 0.43 & 0.10\\
\quad RGB projection & per RGB read & $2.5\times10^{4}$ & $1.3\times10^{2}$ & 0.12 & 0.02\\
Policy, HCMS cascade (RGB projection) & per RGB read & $9.8\times10^{4}$ & $1.0\times10^{3}$ & 0.53 & 0.10\\\bottomrule
\end{tabular}

\end{table*}

\paragraph{Energy of the models and policies.} The total energy of a schedule also includes the TAS model $\Psi$ and, for learned policies, the policy network $\pi$ (\cref{eq:energy_total}). Unlike the sensors, they do not run per read: they run at every step of the model timeline, whatever the number of sensors read in that step, so their power is the step rate times their energy per step (6 steps per second in the Ego-Exo4D experiments of the main paper, 30 when the model runs at 30\,FPS). \Cref{tab:supp_models} gives their cost per step or per read, in MACs, bytes, and energy. At 6 steps per second, {with all five sensors as input (an upper bound, see below), MiniROAD draws 0.46\,mW under the reference model, i.e., 2.3\% of a 20\,mW budget (five times more at 30 steps per second); ProTAS and the causal Transformer draw 70 and 125\,$\mu$W, respectively.} The policy networks cost less than 0.1\,mW ({0.43--0.58\,$\mu$J per step, i.e., 2.6--3.5\,$\mu$W at 6 steps per second, }\cref{tab:supp_models}). The power of a schedule is therefore the sum of the sensor terms, which scale with the read rate of each sensor, and of a small term that depends only on the model and the step rate. Since each sensor has its own rate, the reads of different sensors fall on different steps in asynchronous schedules, and the energy of a step varies along the video, while the model term stays constant. All reported powers are averages over the whole video.

{The cost of a backbone depends on the number of input sensors, because the first layer scales with the sum of the feature dimensions (3200 for all five sensors). We report the cost with all five sensors as input, which is an upper bound for models that take fewer sensors: for instance, ProTAS needs $8.9\times10^{5}$ MACs per step with RGB, audio, and gaze as input, against $9.7\times10^{5}$ with all five. A mixture takes the features of all five sensors whatever the schedule, so its cost does not change with the schedule.}

\FloatBarrier
\section{Training and sensing-strategy details}
\label{supp:impl}
This section gives the details needed to reproduce the models and sensing strategies of \cref{sec:baselines}. We first report the training settings (\cref{subsec:supp_settings}). We then describe Random and Random-CA (\cref{supp:random_policy}), the generator of the randomized schedules on which the mixtures are trained (\cref{sec:supp_schedules}), the budget rule that selects a schedule for a given budget (\cref{supp:budget_rule}), the learned policies (\cref{supp:learned_pol}), and the oracle search and the causal policy for read timing (\cref{supp:read_timing}). The corresponding results are in \cref{supp:add_exp}: the comparison of methods for choosing the schedule in \cref{supp:add_backbones}, random versus periodic reads {and the Greedy window} in \cref{supp:add_random}, the training variants of the learned policies in \cref{supp:learned_results}, and read timing in \cref{supp:add_policies}.
\subsection{Experimental settings}
\label{subsec:supp_settings}
To ensure reproducibility, we provide the specific training configurations used for our baselines in all three datasets. We train per scenario on Ego-Exo4D \cite{grauman2024ego} and CMU-MMAC \cite{de2009guide}, since joint training causes background class to strongly dominate and can lead to collapse. However, for CaptainCook4D \cite{peddi2024captaincook4d}, we perform global training across all scenarios simultaneously as there are not enough data to be trained per scenario (details on annotations in \cref{supp:reannotation}).
For the energy computation, we follow the formulation and energy values in \cref{subsec:energy_computation} using the profiler implementation described in \cref{supp:energy_budget}.

All baselines are implemented in PyTorch and trained with the Adam optimizer. We use a weight decay of $10^{-4}$, an initial learning rate of $5\cdot10^{-4}$, and a cosine annealing learning-rate schedule. For non-learned policies, we train for 50 epochs {on every dataset and scenario,} following the training protocol of ProTAS~\cite{shen2024progress}{, with batch size 2, 1, and 2 on Ego-Exo4D, CMU-MMAC, and CaptainCook4D, respectively}. Every baseline is evaluated with the checkpoint of highest validation accuracy. {Each non-learned policy is trained with its own sensing rule active: the models of Random and Random-CA are trained with dropout rate $\tau_{\mathrm{train}}$, and the model of Greedy is trained with its per-second budget rule, so that it sees the Greedy read pattern during training.}

\paragraph{TAS models.} MiniROAD~\cite{An_2023_ICCV} projects the input features to 2048 dimensions (linear layer, layer normalization, ReLU, dropout 0.2) and feeds them to a one-layer unidirectional GRU with 1024 hidden units and a linear classification head. The causal Transformer has 2 layers with 4 heads and width 128 (dropout 0.3), one token per step computed from the sample-and-hold features, and each prediction attends to at most the past 100\,s. ProTAS~\cite{shen2024progress} is used in its causal version, with 4 stages of 10 dilated residual layers and 64 feature maps.

\paragraph{Specialists and mixtures.} MiniROAD~\cite{An_2023_ICCV}, the causal Transformer and our ProTAS mixtures use the same optimizer, learning rate, weight decay and schedule; MiniROAD is trained on whole videos with cross-entropy on every frame plus a truncated-MSE smoothing term of weight 0.15. On Ego-Exo4D, every model of the paper, specialist or mixture, is trained for 50 epochs with batch size 2, so that all rows share one recipe.

\paragraph{Checkpoint.} All test results are obtained with the checkpoint of highest validation accuracy, as for the baselines; mixtures select it on validation with all sensors read at every step of the 6\,Hz clock. Since some Ego-Exo4D scenarios have only one or two validation videos, this checkpoint can come from the first epochs: scenarios 9 and 10 (6 and 10 training videos) are data-limited for every model.

\subsection{Random and Random-CA}
\label{supp:random_policy}

Random just performs a random dropout of each modality per frame. However, the cost-aware version (Random-CA), takes into account the cost of each modality, enforcing more expensive ones to be dropped more times than the cheaper ones while maintaining the dropout percentage $\tau$ of Random.
The percentage of times the average of the modalities should be on is $p_{\text{on}} = 1.0 - \tau$.

For each modality $m \in \{1, \ldots, M\}$ with associated cost $c_m$, we compute log-normalized costs $\log c_m$. And define the cost range as $\Delta \log c=\max_m \log c_m - \min_m \log c_m$

For each modality, we compute relative expensiveness:
\begin{equation}
e_m = \frac{\log c_m - \min \log c}{\Delta \log c}
\end{equation}

This is transformed into relative selection weights using a weighted interpolation:
\begin{equation}
w_m = \lambda_{\min} + (1 - \lambda_{\min})(1 - e_m)
\end{equation}
where $\lambda_{\min} = 0.25$ ensures all modalities remain sampleable even for expensive modalities.

The weights are then normalized by their mean, $\tilde{w}_m = w_m / (\frac{1}{M}\sum_{k} w_k)$, so that the mean read probability is $p_{\text{on}}$, and, finally, the read probability of each modality at each step is $p_m = \min(1,\, p_{\text{on}}\, \tilde{w}_m)$.
Random and Random-CA are trained with one drop rate $\tau_{\text{train}}$ and evaluated with different $\tau_{\text{inf}}$ to meet each budget; the values are given in parentheses in Table~\ref{tab:all_protas} ($\tau_{\text{train}}/\tau_{\text{inf}}$).

\subsection{Greedy}
\label{supp:greedy_policy}
Greedy resets its budget every second, so that the device does not read in bursts and then stays blind for long. In the reference energy model at 20\,mW, a single read of any sensor set costs more than the budget of one second, so we count the budget over 30\,s windows in the 20\,mW rows of Table~\ref{tab:all_protas}; at 2.8\,W we keep the 1\,s window. The comparison with Blind is in \cref{supp:add_greedy_window}.

\subsection{{Randomized training schedules}}
\label{sec:supp_schedules}
Every time a training video is fed to a model trained on the mixture (\cref{sec:baselines}), it is paired with a new random schedule drawn by the generator described below; at test time the model is run with the fixed schedule under evaluation.

\paragraph{Notation.} The model runs at a base rate of 6\,FPS: a video of $T$ steps lasts $T/6$ seconds. With $M$ sensors (five on Ego-Exo4D: RGB, audio, IMU, gaze, monochrome camera; three on CMU-MMAC and CaptainCook4D: RGB, audio, IMU), a schedule is a set of reads $\sigma\subseteq\{1,\dots,M\}\times\{1,\dots,T\}$, where $(m,t)\in\sigma$ means that sensor $m$ is read at step $t$, as in \cref{sec:benchmark}. A read rate of $f$\,FPS corresponds to a read period of $\Pi=6/f$ steps. The generator draws $\Pi$ from
\begin{eqnarray}
&\mathcal{P}=\{1,2,3,6,12,30,60\}\ \text{steps} &\\ \nonumber &\;\Longleftrightarrow\;& \nonumber \\ &\{6,3,2,1,0.5,0.2,0.1\}\ \text{FPS}.&
\end{eqnarray}
A schedule is built in two stages: a \emph{base schedule} (\cref{sec:supp_base}), followed by two \emph{sensor restrictions} (\cref{sec:supp_pert}).

\subsubsection{Stage 1: base schedule}
\label{sec:supp_base}

Each sensor $m$ is read periodically, independently of the other sensors. A period $\Pi_m$ is drawn uniformly from $\mathcal{P}$ and an offset $o_m$ uniformly from $\{0,\dots,\Pi_m-1\}$; the sensor is read at the steps $t$ such that $(t+o_m) \bmod \Pi_m = 0$. For example, with $\Pi_m=6$ (1\,FPS) and $o_m=2$, the sensor is read at steps $4, 10, 16, \dots$, \ie once per second starting $2/3$\,s into the video. If $\Pi_m=1$, the sensor is read at every step. Sensors therefore run at different rates and are not synchronized, with a random offset $o_m$ so that reads do not always fall on the same steps.

\subsubsection{Stage 2: sensor restrictions}
\label{sec:supp_pert}

The base schedule is then restricted to a subset of the sensors, so that the model also learns to work when some sensors are absent for the whole video. Two restrictions are applied in the following order, each independently of the other:
\begin{enumerate}
\item \textbf{Random sensor subset ($p=0.4$).} A number $k$ is drawn uniformly in $\{1,\dots,M\}$, and only $k$ sensors, chosen uniformly without replacement, are kept for the whole video; the others are never read.
\item \textbf{Device profile ($p=0.6$).} The schedule is restricted to the sensors of a device profile, drawn from \cref{tab:supp_devices}; sensors outside the profile are never read. Profiles are intersected with the sensors available in each dataset (\eg audio+IMU+gaze becomes audio+IMU on CMU-MMAC and CaptainCook4D).
\end{enumerate}

\begin{table}[h]\centering\small
\caption{\textbf{Device profiles} used to restrict training schedules.}
\label{tab:supp_devices}
\setlength{\tabcolsep}{4pt}
\begin{tabular}{@{}llc@{}}\toprule
Profile & Sensors kept & Probability\\\midrule
Glasses without scene camera & audio, IMU, gaze & 25\%\\
Action camera & RGB, audio, IMU & 15\%\\
Watch / phone & audio, IMU & 15\%\\
Eye-tracking + mono & gaze, monochrome & 15\%\\
Glasses with scene camera & RGB, audio, gaze & 15\%\\
Full device & all sensors & 15\%\\\bottomrule
\end{tabular}

\end{table}

\subsubsection{Example schedules}
\label{sec:supp_examples}

\begin{figure*}[t]
    \centering
    \includegraphics[width=1\linewidth]{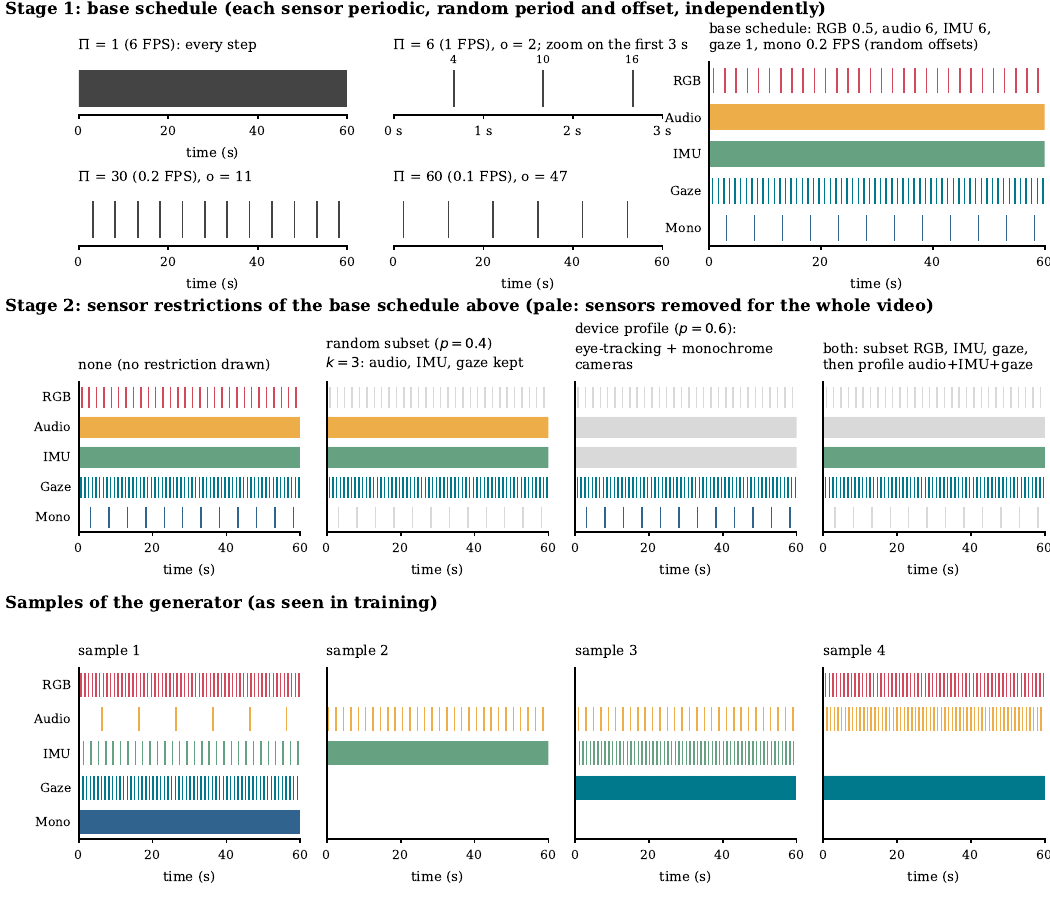}
    \caption{\textbf{Randomized training schedules} (60\,s at 6\,FPS). Top: periodic streams with random period $\Pi$ and offset $o$ (left) and the base schedule they form (right) (Stage~1). Middle: restrictions of this base schedule, with the removed sensors in gray (Stage~2). Bottom: four schedules drawn by the generator, as seen in training.}
    \label{fig:supp_schedules}
\end{figure*}

\paragraph{Examples.} \cref{fig:supp_schedules} shows examples from each stage of the generation process. In the base schedule (Stage~1), each sensor is read at its own periodic rate with a random offset, \eg RGB at 0.5\,FPS, gaze at 1\,FPS, the monochrome camera at 0.2\,FPS, and audio and IMU at 6\,FPS. The restrictions of Stage~2 are applied to this schedule: a random sensor subset (here audio, IMU, and gaze), a device profile (here the eye-tracking and monochrome cameras), or both (here the subset RGB, IMU, and gaze, then the profile audio, IMU, and gaze, which leaves IMU and gaze). The bottom row shows four schedules produced by the generator during training.

\subsection{{Budget rule}}
\label{supp:budget_rule}

{A model trained on randomized schedules (\cref{sec:supp_schedules}) accepts any schedule. Hence, meeting a budget $B$ reduces to choosing which sensors to read and how often. We build a pool of candidate schedules that spend the budget: IMU, nearly free, is always read at 6\,FPS; for each non-empty subset of the other sensors, the remaining budget is split evenly across them (for pairs also 1/4--3/4) and each sensor is read periodically at the rate its share pays for (at most 6\,FPS, any leftover going to the other sensors; a sensor is read every $\Pi$ steps of 1/6\,s, with $\Pi$ an integer, so we round the period up, which rounds the rate down and keeps the schedule within the budget), so cheaper sensors are read more often. Candidates built for a ladder of lower budgets (5, 10, 20, 35, 50, 100, 200, 500, 1000, and 2800\,mW, up to $B$) are also included, so that the rule can spend less than $B$ when that is more accurate. Every candidate is evaluated once on the validation videos with the trained model, and we select the one whose mean power on the validation videos is within the budget and with the highest mean validation accuracy over the scenarios of the dataset, so that one schedule serves the whole dataset (for each model). Since the candidates do not depend on $B$, the same evaluations serve every budget (229 candidates for budgets up to 2.8\,W on Ego-Exo4D, 81 up to 20\,mW, and 39 on CMU-MMAC and CaptainCook4D). Only the selected schedule is evaluated on test.}

Other ways of choosing the schedule are compared with the rule in \cref{supp:add_backbones} (\cref{tab:supp_rule_selection}).

\subsection{Learned policies}
\label{supp:learned_pol}

The learned policies of Table~\ref{tab:learn_where} decide at every step (1/6\,s) which sensors to read. All of them are trained on top of a frozen TAS model (the MiniROAD mixture) and always read the IMU. We describe their architectures, then the objective and the budget constraint; training variants are in \cref{supp:learned_results}.

\subsubsection{Architectures}
All three policies run at the step rate of the TAS model (6 steps per second), always read the IMU, and make one binary read decision per step for each of the other four sensors (RGB, audio, gaze, and monochrome camera). The TAS model is the frozen MiniROAD mixture, which receives each sensor through sample-and-hold: the feature of its last read, or zeros before the first read. The read decisions act only on this input. Each policy normalizes its input features with the per-dimension mean and standard deviation of its training videos. The time since the last read of a sensor enters as $\log(1+s)$, with $s$ in seconds and $s=60$ if the sensor has not been read yet. In training, every decision is a binary Gumbel sample with a straight-through estimator: the forward pass uses the hard 0/1 read, and the backward pass uses the gradient of its relaxation.

\paragraph{AdaMML-style.} At step $t$ the policy receives the IMU feature of step $t$, the log time since the last read of each of the four sensors, and its own four decisions at $t-1$. It never sees the features of the other sensors. These inputs are concatenated into one vector (512 dimensions for the IMU feature, 4 for the times, and 4 for the decisions on Ego-Exo4D). A linear layer with GELU maps this input to 96 dimensions, a GRU cell with 96 hidden units updates the state, and a linear head outputs one read logit per sensor. The decisions set the reads of the same step $t$.

\paragraph{HCMS-style.} The sensors are ordered by energy per read: IMU, audio, gaze, monochrome camera, and RGB. A linear layer with ReLU projects each sensor's feature to 128 dimensions. Each sensor is one level of a cascade, with its own LSTM cell (128 units) whose input is the concatenation of the projections of its own sensor and of all cheaper ones. The IMU level is always active. Each other level decides whether to read its sensor with a gate MLP (a linear layer to 128 units, a ReLU, and a linear layer to two logits, whose difference is the read logit), which takes as input the projections of the cheaper sensors read at $t$ and the hidden and cell states of the level at $t-1$. A sensor is read only if all cheaper ones are, and the projections of unread sensors are set to zero. An active level updates its LSTM state; an inactive level takes the state of the level below. The decisions are taken level by level and set the reads of step $t$. Unlike the other two policies, this one has no time-since-last-read or previous-decision input, as in the original HCMS design, where the LSTM states carry the read history.

\paragraph{Closed loop.} At step $t$ the policy receives what was actually read at $t$: each sensor's feature through its own linear projection to 32 dimensions (zeros if not read), the five read indicators, the log time since the last read of each of the five sensors, and its own four decisions at $t-1$. A linear layer with GELU maps this input to 96 dimensions, a GRU cell with 96 hidden units follows, and a linear head outputs four read logits, which set the reads of step $t+1$. No scheduled sensor is read at the first step. In training, the straight-through gates also multiply the policy's own projected inputs, so the gradient flows through what it chose to read.

\subsubsection{Objective and budget} The final policies maximize accuracy: the loss is the cross-entropy (CE), against the labels, of the frozen TAS model's prediction with the policy's reads. The power constraint is enforced with a Lagrange multiplier $\lambda$, so that the full loss is
\[
\mathcal{L}=\mathrm{CE}+\lambda\,\frac{P-B}{B},
\]
where $P$ is the mean power of the policy's reads in the mini-batch and $B$ the budget left after the IMU. After every step, $\lambda$ is updated by dual ascent,
\[
\lambda\leftarrow\lambda+\eta\,\frac{P-B}{B},
\]
with $\lambda_0=0$ and $\eta=0.005$. The update is two-sided, so that the policy spends the budget rather than undercutting it. 
We train with Adam (learning rate $10^{-3}$) for 3600 iterations (about 300 epochs) on mini-batches of 8 random crops of 540 steps taken from all the training videos of the scenario, with the Gumbel temperature decreasing linearly from 1 to 0.3 (held at 0.3 for the last iterations), and keep the checkpoint with the best validation accuracy.
At test time the policies are deterministic: the read decision is still binary, but a sensor is read when its logit plus a shift is positive, without Gumbel noise. The shift acts as a threshold on the logit: a larger shift makes the policy read more (spending more power), a smaller one makes it more conservative. Since a deterministic policy would otherwise not necessarily spend the budget, we use one shift per policy, shared by all sensors, and find it by binary search on the whole training videos of the policy, increasing the shift until the mean power on those videos matches the budget; on the test videos the power can differ from the budget and exceed it.

\subsection{Read timing: oracle search and {causal policy}}
\label{supp:read_timing}

Part (b) of Table~\ref{tab:learn_where} measures how much accuracy the timing of the reads can add when the number of reads is fixed. It uses two components, described here: an oracle search that finds read times with knowledge of the future (\cref{supp:oracle_search}), and a causal policy trained to imitate it (\cref{supp:causal_router}).

\subsubsection{Oracle search}
\label{supp:oracle_search}
We study read timing for one scheduled sensor (gaze at 20\,mW, RGB at 2.8\,W) while the other sensors keep their periodic schedule, and with the number of reads of the scheduled sensor fixed to that of periodic sampling. The oracle optimizes each video in consecutive windows of $k$ seconds ($k\in\{2,5,20\}$), in order. In each window the number of reads is kept, and every read is moved in turn to the best of a random set of free steps of the window (up to 12 candidates, one pass), where the best step minimizes $\mathrm{KL}(p_{\text{all}}\,\|\,p_{\text{sched}})$ up to the end of the window. Here $p_{\text{all}}$ is the prediction of the same model with all sensors read at every step and $p_{\text{sched}}$ the prediction with the schedule under search. The search uses no labels but sees up to $k$ seconds of the future, so it is non-causal and tailored to the model and the video it is run on: its gain is optimistic and not an upper bound. The \emph{random} rows place the same number of reads at random steps, and the \emph{true boundaries} row places them at the true action boundaries (it uses labels). The oracle therefore measures how much a model could gain from timing alone if the future were known, which is why the main paper reads its gain as headroom and not as a reachable target.

\subsubsection{Causal router imitating the oracle}
\label{supp:causal_router}
The causal policy of the ``imitating'' row of \cref{tab:learn_where} is a GRU (hidden size 96) that decides at step $t$ whether to read the scheduled sensor at step $t+1$. Its input at step $t$ is the feature of every sensor read at $t$ (through a linear projection to 32 dimensions; zeros for the scheduled sensor if it was not read), the read indicator of the scheduled sensor, the logarithm of the time since its last read, and its previous decision. We train it with teacher forcing on the 5\,s oracle schedules of the training videos, with a binary cross-entropy weighted by $(1-p)/p$, where $p$ is the read rate of the oracle. We train for 100 epochs, keeping the epoch with the lowest validation teacher-forcing loss. The row of \cref{tab:learn_where} is the mean over four runs per scenario (two training seeds of the causal policy, each deployed with the two thresholds described below), then over the 15 scenarios. The oracle reads at few steps ($p$ is small), so without weighting the causal policy could reach a low loss by never reading; weighting the read steps by $(1-p)/p$ gives them the same total weight as the steps without a read. At test time each 5\,s block has a quota of reads equal to the periodic reads of the block: the causal policy reads when its score exceeds a threshold, and the remaining reads are forced onto the last steps of the block if the score never crosses it. The threshold is the $(1-q/k)$ quantile of the teacher-forced scores on the training videos, where $q/k$ is the periodic number of reads per step (a block of $k$ steps has a quota of $q$ reads): the causal policy then reads on the fraction $q/k$ of the training steps with the highest scores, which matches the quota on average; a second threshold computed on whole training videos is also used, and the row of \cref{tab:learn_where} is the mean over two causal-policy seeds and the two thresholds. The two thresholds are two ways of calibrating the quota: the causal policy is trained on short excerpts, but at test time it runs on whole videos, where the distribution of its scores can differ, so we do not know which calibration is better. Averaging over the thresholds and over the seeds keeps the row from depending on one arbitrary choice; each of the four runs is below periodic reads. Every block thus has exactly the number of reads of periodic sampling and every schedule has the same energy, so the causal policy of Table~\ref{tab:learn_where} differs from periodic reads only in timing.

\FloatBarrier
\section{{Additional experiments}}
\label{supp:add_exp}
The subsections follow the order of the experiments in the main paper: heuristic sensing strategies (\cref{supp:add_random}), TAS models and training (\cref{supp:add_backbones}), robustness to schedule changes (\cref{supp:add_mismatch}), learned policies and read timing (\cref{supp:learned_results,supp:add_policies}), the edge-SoC energy model (\cref{supp:add_edge}), and the energy-accuracy trade-off curves with additional qualitative examples (\cref{supp:energy_accuracy}).

\subsection{{Heuristic sensing strategies: Random reads and Greedy window}}
\label{supp:add_random}

{This section completes the heuristic sensing strategies of Table~\ref{tab:all_protas}: whether Random reads gain from reading at random times (\cref{supp:add_random_periodic}) and how Greedy compares with Blind under different budget windows (\cref{supp:add_greedy_window}).}

{\subsubsection{Random reads versus periodic reads}
\label{supp:add_random_periodic}}
The main-paper results (Table~\ref{tab:all_protas}) suggest that Random sensing does not gain from reading at random times: it reads every sensor at a reduced rate, which resembles a low-rate multimodal schedule, so its accuracy would come from which sensors are read and how often rather than from when. Here we test this directly. \cref{tab:supp_random} compares iid random reads with periodic reads that have the same number of reads per sensor and video, so that only the timing of the reads differs. We run the test with MiniROAD and ProTAS on Ego-Exo4D, on five scenarios (1, 4, 11, 13, and 14), with three random draws. For MiniROAD we use two settings: Random applied at test time to the schedule mixture, and a separate model trained with Random reads and tested with them. For ProTAS we only apply Random at test time to the schedule mixture. Random reads are not better than periodic ones. With Random 0.9, they lose 0.30 Acc with the mixture and 0.57 with the Random-trained model; with the cost-aware variant (Random-CA 0.7) the difference is within 0.05 points (on the 6\,Hz clock, Random-CA reads four of the five sensors at every step, so only the timing of the RGB reads differs). In this test, then, the timing of the reads does not explain the accuracy of Random; the sensors and rates it reads are the same as the periodic reads. For ProTAS, the result is the same: $-0.16$ Acc with Random 0.9 and $-0.06$ with Random-CA 0.7.
\begin{table*}[t]
\caption{\textbf{Random reads in depth} (MiniROAD architecture and, in the last two rows, the ProTAS mixture; Ego-Exo4D, scenarios 1, 4, 11, 13, 14; reference model; mean power 1.46\,W for Random 0.9 and 2.42\,W for Random-CA 0.7{; Table~\ref{tab:all_protas} uses all 15 scenarios and gives 1.47\,W for Random and 2.56\,W for Random-CA}). Periodic reads have the same number of reads per sensor and video at regular times. $\Delta$Acc: Random minus periodic. $\Delta$Acc is computed before rounding, so it can differ from the difference of the rounded Acc values shown.}
\label{tab:supp_random}
\centering
\scriptsize
\setlength{\tabcolsep}{4pt}
\begin{tabular}{@{}ll rrr rrr r@{}}
\toprule
Model & Reads & \multicolumn{3}{c}{Random reads} & \multicolumn{3}{c}{Periodic, same reads} & $\Delta$Acc \\
 & & Acc & Edit & F1 & Acc & Edit & F1 & \\
\midrule
Mixture, test-time & Random 0.9 & 59.6 & 23.3 & 24.5 & 59.9 & 22.1 & 23.5 & -0.30 \\
Mixture, test-time & Random-CA 0.7 & 60.1 & 20.7 & 22.3 & 60.1 & 20.6 & 22.4 & -0.05 \\
Random-trained & Random 0.9 & 60.5 & 30.4 & 30.8 & 61.1 & 29.3 & 30.1 & -0.57 \\
Random-trained & Random-CA 0.7 & 59.4 & 24.7 & 25.4 & 59.4 & 24.3 & 25.0 & -0.04 \\
\midrule
ProTAS mixture, test-time & Random 0.9 & 48.8 & 33.4 & 27.6 & 48.9 & 32.8 & 28.2 & -0.16 \\
ProTAS mixture, test-time & Random-CA 0.7 & 49.0 & 34.2 & 28.9 & 49.1 & 34.1 & 29.0 & -0.06 \\
\bottomrule
\end{tabular}

\end{table*}

{\subsubsection{Greedy budget window}
\label{supp:add_greedy_window}}
 In the reference energy model at 20\,mW, Greedy with a 1\,s window stays at the Blind level (Table~\ref{tab:supp_greedy}). With 30\,s windows, it improves on Blind in Edit (+10.1, +19.9, and +27.5 points on Ego-Exo4D, CMU-MMAC, and CaptainCook4D) and F1 (+5.1, +14.4, and +23.7), and in Acc on Ego-Exo4D and CaptainCook4D (+8.4 and +11.9; $-0.3$ on CMU-MMAC), while staying below the best Fixed schedule in Acc.

\begin{table}[t]
\centering
\scriptsize
\setlength{\tabcolsep}{2.2pt}
\caption{\textbf{Greedy at 20\,mW with 1\,s and 30\,s budget windows, reference energy model} (ProTAS). With a 1\,s window, Greedy stays at the Blind level; with 30\,s, it spends the budget and improves Edit and F1, and Acc on Ego-Exo4D and CaptainCook4D. Power P (mW), Ego-Exo4D/CMU-MMAC/CaptainCook4D: Blind and 1\,s: 0.4/0.4/0.35; 30\,s: 16.1/11.0/10.9.}
\label{tab:supp_greedy}
\begin{tabular}{@{}l rrr rrr rrr@{}}
\toprule
 & \multicolumn{3}{c}{Blind} & \multicolumn{3}{c}{Greedy, 1\,s} & \multicolumn{3}{c}{Greedy, 30\,s} \\
\cmidrule(lr){2-4}\cmidrule(lr){5-7}\cmidrule(lr){8-10}
 & Acc & Edit & F1 & Acc & Edit & F1 & Acc & Edit & F1 \\
\midrule
Ego-Exo4D & 22.3 & 10.4 & 8.4 & 23.5 & 8.6 & 6.8 & 30.7 & 20.5 & 13.5 \\
CMU-MMAC & 66.3 & 0.0 & 0.0 & 60.7 & 0.0 & 0.0 & 66.0 & 19.9 & 14.4 \\
CaptainCook4D & 23.3 & 9.7 & 1.2 & 23.7 & 6.0 & 1.3 & 35.2 & 37.2 & 24.9 \\
\bottomrule
\end{tabular}
\end{table}

\subsection{{Backbones and training: additional results}}
\label{supp:add_backbones}

This section supports the comparison of TAS models and training in Table~\ref{tab:backbone_training}. \Cref{supp:add_testra} explains why the main paper uses a causal Transformer rather than TeSTra, a more optimized online model: TeSTra alone exceeds the 20\,mW budget. \Cref{supp:add_fixedref} checks that the Fixed schedules of Table~\ref{tab:backbone_training}, chosen with ProTAS in Table~\ref{tab:all_protas}, do not handicap the other models, so that the gain of the budget rule at low budgets is not an artifact of the reference. \Cref{supp:add_rule_selection} compares the budget rule with other ways of choosing the schedule. {\Cref{supp:add_capacity} checks that the advantage of MiniROAD is not due to its size.}

\subsubsection{TeSTra}
\label{supp:add_testra}
The main paper uses a vanilla causal Transformer rather than more optimized online TAS models such as TeSTra~\cite{zhao2022real}, because the cost of the model itself must fit the budgets we study, down to 20\,mW. A reader may ask whether this choice gives up accuracy. We therefore report what TeSTra achieves and what it costs. We trained TeSTra as a specialist on the Fixed schedule of each dataset and budget (Adam, learning rate $7\cdot10^{-5}$, weight decay $5\cdot10^{-5}$, warm-up and cosine schedule).
{On Ego-Exo4D, we use no long memory: TeSTra already runs at six steps per second with an 8\,s work window, and a long memory would add compute to a model that is already costly (see below). On CMU-MMAC and CaptainCook4D, the Fixed schedules read rarely, so the model takes one step per read and the 8\,s work window holds only one to four steps; we therefore keep the long memory, as it is the only component providing temporal context.}
First, the cost of the models without any read, per streamed step and with all five sensors as input: ProTAS needs {0.97}\,MMAC, MiniROAD {16.0}\,MMAC, the causal Transformer {1.64}\,MMAC, and TeSTra {1029}\,MMAC on its 8\,s work window. At 6 steps per second (Ego-Exo4D), this is {0.03, 0.44, 0.05}, and 28.4\,mW, respectively {(reference energy model, multiply-accumulates only)}. TeSTra alone is thus above the 20\,mW budget before any sensing, and too heavy for the low budgets, while the other three models use a small fraction of it. \cref{tab:supp_testra} reports what TeSTra reaches with the Fixed schedules of Table~\ref{tab:all_protas}, which are the same for all models. In accuracy TeSTra is close to MiniROAD (from $-2.9$ to $+0.7$ points), although at 2.8\,W its Edit and F1 are lower on Ego-Exo4D and CaptainCook4D. At 2.8\,W the model cost fits the budget, but the 20\,mW row of Ego-Exo4D (marked $^\dagger$) exceeds the budget once the model cost is counted, so it is not a like-for-like comparison. On CMU-MMAC and CaptainCook4D, instead, the TeSTra models take one step per read (0.05 steps per second at 20\,mW), so their model cost is 0.2\,mW and the three cells fit their budgets (17.0 and 16.6\,mW at 20\,mW, 159.1\,mW at 2.8\,W). The causal Transformer of the main paper costs {0.125\,mW (\cref{tab:supp_models})} and fits every budget: on Ego-Exo4D it is 0.6 points below TeSTra at 2.8\,W (53.4 vs.\ 54.0) and 1.1 points above it at 20\,mW (38.7 vs.\ 37.6), where TeSTra exceeds the budget (differences of rounded table values). TeSTra would thus give at most a small gain in accuracy at 2.8\,W and cannot run within 20\,mW at the 6 steps per second used on Ego-Exo4D, which is why the main paper uses the causal Transformer.
\begin{table*}[t]
\caption{\textbf{TeSTra vs.\ MiniROAD and Transformer specialists} (reference model). Schedules are the Fixed rows of the benchmark. F1 = F1@25. P: power of the pipeline. $^\dagger$: with the attention products counted, TeSTra at the 6 steps per second of Ego-Exo4D costs 28.4\,mW by itself, above the 20\,mW budget. {The CMU-MMAC and CaptainCook4D rows of MiniROAD and TeSTra come from runs with the original recipe of each model, so MiniROAD differs from Table~\ref{tab:backbone_training} (e.g., 87.4 vs.\ 88.3 and 51.9 vs.\ 52.7 at 2.8\,W); the Transformer rows are the specialists of Table~\ref{tab:backbone_training}; models are compared within this table.}}
\centering
\scriptsize
\setlength{\tabcolsep}{4pt}
\begin{tabular}{@{}lll rrr r@{}}
\toprule
Budget & Schedule & Model & Acc & Edit & F1 & P \\
\midrule
\multicolumn{7}{@{}l}{\textit{Ego-Exo4D (15 scenarios)}} \\
2.8\,W & 6\,FPS R+A+G & MiniROAD & 56.3 & 25.5 & 25.7 & 2.31\,W \\
 & & TeSTra & 54.0 & 19.3 & 20.3 & 2.32\,W \\
 & & Transformer & 53.4 & 12.1 & 11.6 & 2.31\,W \\
20\,mW & 0.1\,FPS G+M & MiniROAD & 40.5 & 33.4 & 25.4 & 17.2\,mW \\
 & & TeSTra$^\dagger$ & 37.6 & 31.2 & 22.2 & 33.2\,mW \\
 & & Transformer & 38.7 & 31.8 & 24.9 & {16.9}\,mW \\
\midrule
\multicolumn{7}{@{}l}{\textit{CMU-MMAC (5 scenarios)}} \\
2.8\,W & 6\,FPS R & MiniROAD & 87.4 & 34.7 & 39.8 & 1.86\,W \\
 & & TeSTra & 87.7 & 32.5 & 39.4 & 1.89\,W \\
 & & Transformer & 85.5 & 15.3 & 18.2 & 1.91\,W \\
20\,mW & 0.05\,FPS R+A+I & MiniROAD & 71.8 & 24.4 & 26.9 & 16.8\,mW \\
 & & TeSTra & 71.4 & 32.1 & 30.7 & 17.0\,mW \\
 & & Transformer & 72.1 & 26.6 & 30.2 & 17\,mW \\
\midrule
\multicolumn{7}{@{}l}{\textit{CaptainCook4D}} \\
2.8\,W & 0.5\,FPS R & MiniROAD & 51.9 & 23.8 & 22.8 & 155.2\,mW \\
 & & TeSTra & 49.4 & 19.0 & 16.1 & 159.1\,mW \\
 & & Transformer & 47.0 & 16.5 & 13.3 & {150\,mW} \\
20\,mW & 0.05\,FPS R+A & MiniROAD & 46.8 & 37.0 & 26.7 & 16.3\,mW \\
 & & TeSTra & 47.5 & 40.3 & 28.8 & 16.6\,mW \\
 & & Transformer & 45.6 & 38.1 & 23.5 & 16\,mW \\
\bottomrule
\end{tabular}

\label{tab:supp_testra}
\end{table*}

\subsubsection{Fixed schedule: ProTAS choice vs.\ each mixture's own}
\label{supp:add_fixedref}
Table~\ref{tab:backbone_training} runs every model on the Fixed schedule of Table~\ref{tab:all_protas}, which was chosen with ProTAS specialists. If that schedule suited ProTAS but not the other models, the Fixed rows of MiniROAD and the Transformer would be handicapped and the gain of the budget rule over Fixed would be inflated. We check this in two ways: whether the ranking of the schedules carries over from ProTAS specialists to the mixtures, and how each mixture does on the Fixed schedule chosen for it.

\paragraph{Rank of the schedules.} Is a schedule that is good for a ProTAS specialist also good for a mixture? \cref{tab:supp_rank} correlates the accuracy of 19 schedules of ProTAS specialists with that of each mixture on the same schedule (Ego-Exo4D, 15 scenarios, reference model, inference only). The ranking is nearly preserved over all schedules (Spearman 0.94--0.96, $p<0.001$), and, {among these 19 schedules, the best ProTAS schedule is also the best for every mixture}. Above 20\,mW the agreement is high (0.85--0.96, $p\le0.001$). At or below 20\,mW it is lower (0.64 for the three mixtures) and, with 8 schedules, not statistically significant ($p=0.09$). The ranking thus carries over at high power, but at low power a schedule chosen with ProTAS may not be the best one for another model, which is why we test the two budgets of Table~\ref{tab:backbone_training} directly below.
\begin{table}[t]
\caption{\textbf{Spearman correlation between ProTAS specialists and each mixture over the same schedules} (Ego-Exo4D, 15 scenarios; schedules grouped by reference power; in brackets, number of schedules; $p$: two-sided test of the correlation).}
\label{tab:supp_rank}
\centering
\scriptsize
\setlength{\tabcolsep}{3pt}
\begin{tabular}{@{}l rrr@{}}
\toprule
 & all (19) & $\le$20\,mW (8) & $>$20\,mW (11) \\
\midrule
MiniROAD mixture & 0.96 ($p<0.001$) & 0.64 ($p=0.09$) & 0.96 ($p<0.001$) \\
ProTAS mixture & 0.96 ($p<0.001$) & 0.64 ($p=0.09$) & 0.96 ($p<0.001$) \\
Transformer mixture & 0.94 ($p<0.001$) & 0.64 ($p=0.09$) & 0.85 ($p=0.001$) \\
\bottomrule
\end{tabular}

\end{table}

\paragraph{Each mixture on its own best Fixed schedule.} For each mixture we choose the most accurate Fixed schedule on its own validation videos, among a {common-rate grid} of Fixed schedules, each reading a subset of the five sensors at one common rate (between 0.01 and 6\,FPS), and within the budget, and evaluate it once on test (\cref{tab:supp_ownfixed}; Ego-Exo4D, reference model). {In Table~\ref{tab:backbone_training} the Fixed schedule is chosen once, with ProTAS, and shared by all mixtures. Here, instead, every mixture chooses its own, with the same kind of selection as the budget rule (one evaluation on validation, within the budget). As a result, the ProTAS numbers of \cref{tab:supp_ownfixed} are not identical to those of Table~\ref{tab:backbone_training}, but they are close ($-0.7$ Acc points at 2.8\,W and $0.0$ at 20\,mW; differences of rounded table values).} At 2.8\,W the choice differs from the ProTAS one for {all three mixtures: MiniROAD (A+I+G+M at 6\,FPS), ProTAS and the Transformer (all five sensors at 1\,FPS)}; at 20\,mW all three mixtures choose A+I+G+M at 0.1\,FPS instead of G+M. The accuracy changes by $-1.0$ to $+1.2$ points with respect to the Fixed schedule of Table~\ref{tab:backbone_training} (the Transformer at 2.8\,W loses 1.0: its validation choice does not carry over to test). The budget rule stays above the mixture's own best Fixed schedule at 20\,mW ($+5.8$ MiniROAD, {$+2.7$} ProTAS, $+4.4$ Transformer) and within {0.5} points of it at 2.8\,W (differences of rounded table values). The gain of the rule at low budgets in Table~\ref{tab:backbone_training} therefore does not come from a Fixed reference chosen for another model.
\begin{table}[t]
\caption{\textbf{Each mixture on its own best Fixed schedule} (Ego-Exo4D, 15 scenarios, reference model; schedule chosen on the validation videos within the budget; raw predictions; F1 = F1@25). Last two columns: Acc of the same mixture on the Fixed schedule chosen with ProTAS and with the budget rule (Table~\ref{tab:backbone_training}).}
\label{tab:supp_ownfixed}
\centering
\scriptsize
\setlength{\tabcolsep}{2.4pt}
\begin{tabular}{@{}l l rrr rr@{}}
\toprule
 & & \multicolumn{3}{c}{Own best Fixed} & \multicolumn{2}{c}{Acc, Table~\ref{tab:backbone_training}} \\
\cmidrule(lr){3-5}\cmidrule(l){6-7}
Mixture & Own best Fixed & Acc & F1 & P & Fixed & Rule \\
\midrule
\multicolumn{7}{@{}l}{\textit{2.8\,W}} \\
MiniROAD & A+I+G+M 6\,FPS & 55.2 & 21.2 & 1.06\,W & 55.0 & 55.4 \\
ProTAS & R+A+I+G+M 1\,FPS & 39.3 & 21.8 & 0.49\,W & {40.0} & {39.8} \\
Transformer & R+A+I+G+M 1\,FPS & 51.1 & 24.0 & 0.49\,W & 52.1 & 51.5 \\
\midrule
\multicolumn{7}{@{}l}{\textit{20\,mW}} \\
MiniROAD & A+I+G+M 0.1\,FPS & 43.0 & 26.7 & 18.6\,mW & 41.8 & 48.8 \\
ProTAS & A+I+G+M 0.1\,FPS & 34.7 & 17.3 & 18.2\,mW & {34.7} & {37.4} \\
Transformer & A+I+G+M 0.1\,FPS & 40.8 & 27.0 & 18.3\,mW & 40.7 & 45.2 \\
\bottomrule
\end{tabular}

\end{table}

\subsubsection{How to choose the schedule}
\label{supp:add_rule_selection}
The budget rule (\cref{supp:budget_rule}) builds its candidate schedules with hand-designed splits of the budget among sensors. We ask whether a different selection procedure would find more accurate schedules. \cref{tab:supp_rule_selection} compares the rule with three other ways of choosing the schedule, at nine budgets, with the MiniROAD-architecture model trained on the mixture (Ego-Exo4D, 15 scenarios). Every choice is made on the validation videos and evaluated once on test.

\paragraph{Competitors.}
\begin{itemize}
\item \emph{{Per-sensor rate grid}.} It does not derive the rates from the budget. It evaluates a fixed grid of 120 schedules, in which each sensor is either off or read at one of a few predefined rates (two to four values per sensor, chosen independently; the IMU is always read at 6\,FPS), and picks the most accurate one whose power fits the budget. The budget rule instead builds its candidates from the budget, so that each candidate spends it (even split among the sensors, each read at the rate its share pays for).
\item \emph{CMA-ES} learns how to split the budget among sensors with the covariance matrix adaptation evolution strategy~\cite{hansen2016cma}, a black-box optimizer that samples candidate solutions from a Gaussian distribution and adapts its mean and covariance towards the best ones. The parameters are five logits, one per scheduled sensor (RGB, audio, gaze, monochrome) and one for an unused share; their softmax gives the share of the budget left after the IMU (always read at 6\,FPS) that goes to each sensor, and each sensor is read at the rate its share pays for (at most 6 reads per second). The objective is the accuracy of the frozen mixture on the validation videos, averaged over the scenarios. CMA-ES runs from scratch for each budget with population 8, initial step size 1.5 and 48 evaluations; the best evaluated rates are kept and turned into a periodic schedule with the periods rounded up, as in the rule.
\item \emph{Random read times} keeps the rates of the budget rule but places the reads at random steps.
\end{itemize}

\paragraph{Results.} The budget rule is the most accurate or tied at eight of the nine budgets; CMA-ES is 0.8 points above it at 10\,mW and 1.1, 1.1 and 1.4 points below it at 50, 100 and 200\,mW (CMA-ES picks the rule's schedule at 20\,mW). Random read times lose 3.6, 4.4 and 2.9 points at 5, 10 and 20\,mW and tie at 2.8\,W, in line with \cref{supp:add_random}, and the {per-sensor} rate grid loses 8.2 points at 5\,mW. {Here every sensor is read at random times (at each step, independently, with probability equal to the inverse of its period; one draw per video), so the number of reads varies around the periodic one: this comparison asks whether choosing the rates matters more than choosing the read times. In \cref{tab:learn_where}(b), instead, only the routed sensor is randomized, with exactly the number of reads of the periodic schedule, to isolate timing; Random loses 1.8 points at 20\,mW there, so the two values differ.} The selection is also cheap: the 229 candidates of the rule are evaluated once on validation and serve all nine budgets, whereas CMA-ES needs 48 new evaluations per budget. Within this comparison, the budget rule is a reasonable choice: no competitor is more accurate at more than one budget, and the only one with fewer validation evaluations, the {per-sensor} rate grid (120 schedules, also shared by all budgets), loses 8.2 points at 5\,mW.

\begin{table*}[t]
\caption{{\textbf{How to choose the schedule.} Test accuracy of the schedule chosen on the validation videos (\emph{val}: validation accuracy the choice was made on) by four methods at nine budgets, MiniROAD-architecture model trained on the mixture, Ego-Exo4D, 15 scenarios. CMA-ES picks the budget rule's schedule at 20\,mW.}}
\label{tab:supp_rule_selection}
\centering
\scriptsize
\resizebox{\linewidth}{!}{\begin{tabular}{lccccccccc}\toprule
\multicolumn{10}{l}{\emph{(i) How to choose the schedule} (test Acc; \emph{val} = validation mean the choice was made on)}\\
Budget & 5\,mW & 10\,mW & 20\,mW & 50\,mW & 100\,mW & 200\,mW & 500\,mW & 1\,W & 2.8\,W\\\midrule
Budget rule (budget-filling, periodic) & \textbf{37.7} & 44.8 & \textbf{48.8} & \textbf{52.2} & \textbf{54.3} & \textbf{53.9} & \textbf{54.8} & \textbf{55.1} & \textbf{55.4}\\
\quad \emph{val} & 41.6 & 48.9 & 52.9 & 55.9 & 57.6 & 58.5 & 59.5 & 59.8 & 60.2\\
{Per-sensor rate grid} (fixed per-sensor rates) & 29.5 & 43.2 & 47.4 & 49.3 & 52.9 & 52.9 & 54.2 & 54.7 & 54.7\\
\quad \emph{val} & 33.7 & 46.2 & 50.4 & 53.2 & 58.0 & 58.0 & 59.0 & 59.8 & 59.8\\
CMA-ES (learned per-sensor rates) & \textbf{37.7} & \textbf{45.6} & \textbf{48.8} & 51.1 & 53.2 & 52.5 & \textbf{54.8} & \textbf{55.1} & \textbf{55.4}\\
Random read times (budget-rule rates) & 34.1 & 40.4 & 45.9 & 51.0 & 52.9 & 52.3 & 54.6 & 54.8 & \textbf{55.4}\\
\bottomrule\end{tabular}}

\end{table*}

\subsubsection{{Model capacity}}
\label{supp:add_capacity}
{MiniROAD has 16.0M parameters on Ego-Exo4D (hidden units 2048/1024), against 0.98M for ProTAS and 1.23M for the causal Transformer. While we follow the paper's original architecture~\cite{An_2023_ICCV}, a reader may ask whether the advantage of MiniROAD comes from its size. We therefore trained the mixture with 256/256 hidden units (1.22M parameters, the size of the causal Transformer), with the recipe of Table~\ref{tab:backbone_training} and the schedules selected for the larger model (the rule is not re-run on the smaller one). \Cref{tab:supp_capacity} compares it with the models of Table~\ref{tab:backbone_training} on the four columns of that table, and \cref{tab:supp_capacity_mismatch} on the schedule changes of Table~\ref{tab:mismatch}, with the all-sensors mixture of \cref{supp:mismatch_all}. The smaller MiniROAD is as accurate as the larger one: it is 0.5 points higher at 2.8\,W with Fixed schedule and 0.4 at 20\,mW with Fixed schedule, and 0.7 points lower at 2.8\,W with the budget rule, so the differences are mixed and small (differences of rounded table values). The ordering of the three models is the same in every column (MiniROAD above the causal Transformer above ProTAS), and the power changes by less than 0.6\,mW at 20\,mW. In \cref{tab:supp_capacity_mismatch}, the mixture keeps its advantage over the specialist when the schedule changes (for instance 29.9 vs.\ 13.8 with audio only and 53.8 vs.\ 49.5 with gaze only), so the robustness of the mixture does not depend on the size of the model. The main paper reports the larger model, which is the one used in all the other tables; the smaller one is a capacity check on Ego-Exo4D.}

\begin{table*}[t]
\caption{{\textbf{Model capacity on Ego-Exo4D.} Acc of the mixtures with Fixed schedules (R+A+G 6\,FPS at 2.8\,W, G+M 0.1\,FPS at 20\,mW) and with the budget rule, as in Table~\ref{tab:backbone_training}; gray: 2\,s causal filter. The 256/256 MiniROAD (1.22M parameters) is as accurate as the 2048/1024 one (16.0M) and the ordering of the models is unchanged.}}
\label{tab:supp_capacity}
\centering
\scriptsize
\setlength{\tabcolsep}{3pt}
\begin{tabular}{@{}l rrrr@{}}
\toprule
 & \multicolumn{2}{c}{2.8\,W} & \multicolumn{2}{c}{20\,mW} \\
\cmidrule(lr){2-3}\cmidrule(lr){4-5}
 & Fixed & mix.\,+\,rule & Fixed & mix.\,+\,rule \\
\midrule
MiniROAD, 256/256 hidden units & 55.5\makebox[1.3em][l]{$_{\scriptscriptstyle \textcolor{gray}{54.1}}$} & 54.7\makebox[1.3em][l]{$_{\scriptscriptstyle \textcolor{gray}{53.1}}$} & 42.2\makebox[1.3em][l]{$_{\scriptscriptstyle \textcolor{gray}{41.0}}$} & 48.9\makebox[1.3em][l]{$_{\scriptscriptstyle \textcolor{gray}{47.2}}$} \\
MiniROAD, 2048/1024 hidden units (main paper) & 55.0\makebox[1.3em][l]{$_{\scriptscriptstyle \textcolor{gray}{53.8}}$} & 55.4\makebox[1.3em][l]{$_{\scriptscriptstyle \textcolor{gray}{54.1}}$} & 41.8\makebox[1.3em][l]{$_{\scriptscriptstyle \textcolor{gray}{40.6}}$} & 48.8\makebox[1.3em][l]{$_{\scriptscriptstyle \textcolor{gray}{47.1}}$} \\
ProTAS & 40.0\makebox[1.3em][l]{$_{\scriptscriptstyle \textcolor{gray}{39.0}}$} & 39.8\makebox[1.3em][l]{$_{\scriptscriptstyle \textcolor{gray}{38.8}}$} & 34.7\makebox[1.3em][l]{$_{\scriptscriptstyle \textcolor{gray}{34.1}}$} & 37.4\makebox[1.3em][l]{$_{\scriptscriptstyle \textcolor{gray}{36.5}}$} \\
Causal Transformer & 52.1\makebox[1.3em][l]{$_{\scriptscriptstyle \textcolor{gray}{52.5}}$} & 51.5\makebox[1.3em][l]{$_{\scriptscriptstyle \textcolor{gray}{51.3}}$} & 40.7\makebox[1.3em][l]{$_{\scriptscriptstyle \textcolor{gray}{39.5}}$} & 45.2\makebox[1.3em][l]{$_{\scriptscriptstyle \textcolor{gray}{43.8}}$} \\
\textit{\quad P (MiniROAD, 256/256 / 2048/1024), mW} & \multicolumn{1}{r}{\textit{2307 / 2307}} & \multicolumn{1}{r}{\textit{1994 / 1995}} & \multicolumn{1}{r}{\textit{16.8 / 17.2}} & \multicolumn{1}{r}{\textit{19.2 / 19.7}} \\
\bottomrule
\end{tabular}

\end{table*}

\begin{table*}[t]
\caption{{\textbf{Model capacity under schedule changes (Acc).} {Same test schedules as Table~\ref{tab:mismatch}}, absolute values. The smaller mixture is within 1 point of the larger one except with audio only (+3.5).}}
\label{tab:supp_capacity_mismatch}
\centering
\scriptsize
\setlength{\tabcolsep}{2.4pt}
\begin{tabular}{@{}l rrrrrrrr@{}}
\toprule
 & train & 1\,FPS & 0.1\,FPS & no gaze & no RGB & audio only & gaze only & RGB only \\
\midrule
MiniROAD 256/256, mixture & 55.5 & 54.4 & 44.5 & 47.4 & 55.6 & 29.9 & 53.8 & 46.5 \\
MiniROAD 2048/1024, mixture & 55.0 & 54.1 & 43.6 & 47.7 & 54.9 & 26.4 & 53.6 & 46.7 \\
MiniROAD 2048/1024, specialist & 56.3 & 54.7 & 43.0 & 41.5 & 49.2 & 13.8 & 49.5 & 42.3 \\
\bottomrule
\end{tabular}

\end{table*}

\subsubsection{Statistical robustness}
\label{supp:stats}
The comparisons of the main paper use one training seed and report means over 15 Ego-Exo4D scenarios. We check whether the key differences exceed the variation across test videos (\cref{tab:supp_stats_ci}) and across training seeds (\cref{tab:supp_seeds}).

\paragraph{Test videos.} \Cref{tab:supp_stats_ci} reports 95\% paired bootstrap intervals for the Acc differences behind the claims of Tables~\ref{tab:backbone_training} and~\ref{tab:learn_where}. Each interval is obtained by repeating 10{,}000 times the following steps: draw scenarios at random with replacement, draw the test videos of each drawn scenario at random with replacement, compute the Acc of the two compared methods on the drawn videos, and take the difference. The interval contains the central 95\% of these differences. Since the two methods are scored on the same drawn videos, it measures the difference between them, and since both scenarios and videos are redrawn, it includes the variation across both. The rule's gain over the mixture at 20\,mW, the ProTAS mixture's gap at 2.8\,W, and the oracle's gain and the causal policy's loss at 20\,mW stay away from zero. Two differences include zero, in line with the main paper: the closed loop's deficit against periodic reads at 20\,mW (the learned policy gives no gain over periodic reads) and the small 2.8\,W gaps of MiniROAD on CMU-MMAC, which has only five scenarios (the mixture and the rule are close to the specialist at 2.8\,W).

\begin{table}[t]
\caption{\textbf{Bootstrap intervals over test videos.} Mean Acc difference over scenarios ($\Delta$, exact mean of per-scenario differences, so it can differ by 0.1 from the difference of rounded table values) and 95\% paired bootstrap interval, clustered by test video within scenarios.}
\label{tab:supp_stats_ci}
\centering
\scriptsize
\setlength{\tabcolsep}{2.5pt}
\begin{tabular}{@{}l r c@{}}
\toprule
Comparison (Acc) & $\Delta$ & 95\% CI \\
\midrule
\multicolumn{3}{@{}l}{\textit{Ego-Exo4D, Table~\ref{tab:backbone_training}}} \\
MiniROAD, 20\,mW, mix.\,+\,rule $-$ mix. & $+$7.0 & [$+$4.3, $+$9.6] \\
Transformer, 20\,mW, mix.\,+\,rule $-$ mix. & $+$4.5 & [$+$0.2, $+$8.3] \\
ProTAS, 20\,mW, mix.\,+\,rule $-$ mix. & $+$2.8 & [$+$1.1, $+$4.7] \\
ProTAS, 2.8\,W, mix. $-$ spec. & $-$8.1 & [$-$14.4, $-$3.6] \\
\midrule
\multicolumn{3}{@{}l}{\textit{CMU-MMAC, Table~\ref{tab:backbone_training}}} \\
MiniROAD, 2.8\,W, mix. $-$ spec. & $-$1.7 & [$-$3.8, $+$0.0] \\
MiniROAD, 2.8\,W, mix.\,+\,rule $-$ mix. & $-$0.8 & [$-$2.1, $+$0.2] \\
\midrule
\multicolumn{3}{@{}l}{\textit{Ego-Exo4D, Table~\ref{tab:learn_where}, 20\,mW, MiniROAD mixture}} \\
Closed loop $-$ periodic, same reads & $-$1.7 & [$-$3.3, $+$0.0] \\
Oracle (5\,s) $-$ periodic & $+$5.9 & [$+$3.4, $+$9.0] \\
Causal router $-$ periodic & $-$1.6 & [$-$2.6, $-$0.6] \\
\bottomrule
\end{tabular}

\end{table}

\paragraph{Training seeds.} \Cref{tab:supp_seeds} retrains the MiniROAD\footnote{We use the 256/256 model of \cref{supp:add_capacity} instead of the model of the main paper because it is cheaper to retrain and gives similar results.} mixture and the Fixed specialists with three training seeds\footnote{The seed sets the random number generators of Python and PyTorch, which determine the initialization, the order of the training data, and dropout. It does not change the sequence of randomized training schedules of the mixture, which is drawn from a fixed generator.} and reports the mean and standard deviation over seeds, and 95\% bootstrap confidence intervals for the difference between two methods. The schedule of the budget rule is the same for all seeds: it is selected once on validation and not selected again for each seed. Each interval is obtained by repeating 10{,}000 times the following steps: draw 15 scenarios at random with replacement, draw for each drawn scenario three of the training seeds at random with replacement (so a seed can appear several times or not at all), compute the Acc difference between the two methods trained with the same seed, and average it over the drawn scenarios and seeds. The interval goes from the 2.5th to the 97.5th percentile of these averages, so it includes the variation across scenarios and across training seeds. The standard deviation over seeds of each method (upper block of \cref{tab:supp_seeds}) is at most 1.2 points, and the rule's gain over the mixture at 20\,mW is $+6.11$ (95\% CI [$+3.00$, $+8.84$]), well away from zero. At 2.8\,W the differences are small and the intervals of the mixture and of the rule against the specialist include zero (mixture $-0.89$, 95\% CI [$-2.28$, $+0.37$]; rule $-1.40$, 95\% CI [$-2.98$, $+0.09$]), so we cannot distinguish them from the specialist. The rule is $0.51$ points below the mixture on the Fixed schedule, and its interval ([$-1.02$, $-0.03$]) only just excludes zero. The main conclusions at 20\,mW therefore hold across test videos and training seeds, while at 2.8\,W the rule, the mixture, and the specialist should be read as close.

\begin{table}[t]
\caption{\textbf{Training seeds.} MiniROAD 256/256 (\cref{supp:add_capacity}), Ego-Exo4D, 15 scenarios, three training seeds; specialists on the Fixed schedules of Table~\ref{tab:backbone_training}; the rule's schedule is the same for all seeds (it is not selected again for each seed). $\Delta$: mean over scenarios of seed-averaged differences; 95\% bootstrap intervals over scenarios and seeds.}
\label{tab:supp_seeds}
\centering
\scriptsize
\setlength{\tabcolsep}{2.5pt}
\begin{tabular}{@{}l ccc@{}}
\toprule
Acc, mean $\pm$ std & spec. & mix. & mix.\,+\,rule \\
\midrule
2.8\,W & 55.8\,$\pm$\,0.4 & 54.9\,$\pm$\,1.2 & 54.4\,$\pm$\,1.1 \\
20\,mW & 38.8\,$\pm$\,0.2 & 42.3\,$\pm$\,0.4 & 48.4\,$\pm$\,0.5 \\
\bottomrule
\end{tabular}
\par\vspace{4pt}
\begin{tabular}{@{}l l r c@{}}
\toprule
 & Comparison (Acc) & $\Delta$ & 95\% CI \\
\midrule
2.8\,W & mix. $-$ spec. & $-$0.89 & [$-$2.28, $+$0.37] \\
 & mix.\,+\,rule $-$ mix. & $-$0.51 & [$-$1.02, $-$0.03] \\
 & mix.\,+\,rule $-$ spec. & $-$1.40 & [$-$2.98, $+$0.09] \\
\midrule
20\,mW & mix. $-$ spec. & $+$3.45 & [$+$1.82, $+$5.36] \\
 & mix.\,+\,rule $-$ mix. & $+$6.11 & [$+$3.00, $+$8.84] \\
 & mix.\,+\,rule $-$ spec. & $+$9.56 & [$+$6.42, $+$12.64] \\
\bottomrule
\end{tabular}

\end{table}

\subsection{Robustness to schedule changes: full results}
\label{supp:add_mismatch}

Table~\ref{tab:mismatch} supports the claim that one mixture stays close to the specialist on the specialist's schedule and is better when sensors are missing. This section gives the absolute scores behind Table~\ref{tab:mismatch} (\cref{supp:mismatch_abs}), repeats the comparison with the all-sensors mixture of the other tables (\cref{supp:mismatch_all}), and checks the claim against a specialist trained at a low budget (\cref{supp:mismatch_20mw}).

\subsubsection{Absolute scores}
\label{supp:mismatch_abs}
\cref{tab:supp_mismatch_abs} reports the scores from which the differences of Table~\ref{tab:mismatch} are computed: for each model, the specialist (trained on RGB, audio and gaze at 6\,FPS) and the control mixture (trained on randomized schedules of the same three sensors), tested on the same schedules, with Edit in addition to Acc and F1. The absolute values show that the gaps of Table~\ref{tab:mismatch} are measured against strong references: on its own schedule the specialist reaches 48.0 (ProTAS), 53.4 (Transformer) and 56.3 (MiniROAD) Acc. They also show where the gains with missing sensors come from: the models lose accuracy when a sensor is removed, and the mixture loses less, so a positive difference in Table~\ref{tab:mismatch} means that the mixture degrades more gracefully, not that it is more accurate than on its training schedule (\eg with audio only the mixtures reach 28.1, 26.9 and 29.0 Acc, against 15.8, 16.9 and 13.8 for the specialists). Edit follows F1 on the specialist's own schedule: the mixture is below the specialist for MiniROAD ($-2.6$), the Transformer ($-1.0$), and ProTAS ($-3.3$).
\begin{table*}[t]
\caption{\textbf{Absolute scores behind Table~\ref{tab:mismatch}} (same rows and models; Ego-Exo4D, 15 scenarios). The differences of Table~\ref{tab:mismatch} are computed before rounding, so they can differ by 0.1 from the difference of the rounded values here.}
\label{tab:supp_mismatch_abs}
\centering
\scriptsize
\setlength{\tabcolsep}{2.4pt}
\begin{tabular}{@{}l rr rr rr @{\hspace{8pt}} rr rr rr @{\hspace{8pt}} rr rr rr@{}}
\toprule
 & \multicolumn{6}{c}{ProTAS} & \multicolumn{6}{c}{Transformer} & \multicolumn{6}{c}{MiniROAD} \\
\cmidrule(lr){2-7}\cmidrule(lr){8-13}\cmidrule(l){14-19}
 & \multicolumn{2}{c}{Acc} & \multicolumn{2}{c}{Edit} & \multicolumn{2}{c}{F1} & \multicolumn{2}{c}{Acc} & \multicolumn{2}{c}{Edit} & \multicolumn{2}{c}{F1} & \multicolumn{2}{c}{Acc} & \multicolumn{2}{c}{Edit} & \multicolumn{2}{c}{F1} \\
Test schedule (R, A, G) & Spec. & Mix. & Spec. & Mix. & Spec. & Mix. & Spec. & Mix. & Spec. & Mix. & Spec. & Mix. & Spec. & Mix. & Spec. & Mix. & Spec. & Mix. \\
\midrule
Regular, 6\,FPS (training schedule) & 48.0 & 42.3 & 29.2 & 25.9 & 25.9 & 21.9 & 53.4 & 53.3 & 12.1 & 11.1 & 11.6 & 10.7 & 56.3 & 56.8 & 25.5 & 22.9 & 25.7 & 23.4 \\
\midrule
Regular, 1\,FPS & 46.9 & 41.9 & 30.3 & 27.1 & 25.6 & 22.3 & 52.4 & 52.8 & 26.2 & 24.3 & 24.0 & 24.1 & 54.7 & 56.0 & 30.9 & 30.2 & 29.5 & 29.4 \\
Regular, 0.1\,FPS & 38.4 & 37.1 & 33.4 & 27.7 & 19.8 & 18.2 & 43.2 & 44.2 & 35.7 & 36.9 & 27.5 & 29.1 & 43.0 & 45.9 & 37.4 & 37.5 & 27.0 & 28.9 \\
\midrule
Gaze absent & 36.8 & 38.6 & 27.1 & 26.5 & 19.6 & 20.4 & 40.7 & 44.2 & 14.9 & 12.4 & 11.1 & 10.9 & 41.5 & 48.5 & 23.1 & 23.6 & 18.6 & 21.1 \\
RGB absent & 39.3 & 42.3 & 24.1 & 26.1 & 18.8 & 20.5 & 40.2 & 51.5 & 8.8 & 9.0 & 6.4 & 7.9 & 49.2 & 57.5 & 19.8 & 18.8 & 17.4 & 19.8 \\
Audio only & 15.8 & 28.1 & 16.7 & 21.8 & 7.0 & 14.1 & 16.9 & 26.9 & 10.3 & 9.8 & 4.4 & 5.2 & 13.8 & 29.0 & 7.5 & 16.5 & 3.3 & 11.1 \\
Gaze only & 38.6 & 41.0 & 21.0 & 24.4 & 16.0 & 19.6 & 40.9 & 50.1 & 6.6 & 8.0 & 4.3 & 6.8 & 49.5 & 57.0 & 16.3 & 18.2 & 16.1 & 19.3 \\
RGB only & 35.2 & 37.3 & 25.2 & 25.0 & 18.6 & 19.1 & 38.4 & 41.6 & 12.2 & 11.1 & 9.1 & 10.0 & 42.3 & 47.4 & 22.2 & 23.7 & 19.0 & 20.6 \\
\bottomrule
\end{tabular}
\par\vspace{2pt}
{\footnotesize R = RGB, A = Audio, G = Gaze. Spec.: specialist trained on 6\,FPS R+A+G; Mix.: control mixture, trained on randomized schedules of R, A, G only.}

\end{table*}

\subsubsection{Mixture trained on all five sensors}
\label{supp:mismatch_all}
The mixture of Table~\ref{tab:mismatch} is trained on randomized schedules of RGB, audio and gaze only, so that it is compared with a specialist that reads the same sensors. The mixtures of the other tables are instead trained on randomized schedules of all five sensors. \cref{tab:supp_mismatch_all_delta} repeats the comparison of Table~\ref{tab:mismatch} with that all-sensors mixture, and \cref{tab:supp_mismatch_abs_all} gives the absolute scores. The conclusion is the same: the mixture gains when sensors are missing (audio only $+10.3$ for ProTAS, $+8.2$ for the Transformer, $+12.6$ for MiniROAD Acc), but the gap on the specialist's own schedule is larger (ProTAS $-8.1$, Transformer $-1.4$, MiniROAD $-1.3$ Acc), possibly because part of the training capacity is spent on IMU and monochrome schedules that the specialist never sees.

\begin{table}[t]
\caption{\textbf{All-sensors mixture versus the specialist} (as in Table~\ref{tab:mismatch}, with the mixture trained on randomized schedules of all five sensors, as in the other tables; Ego-Exo4D, 15 scenarios).}
\label{tab:supp_mismatch_all_delta}
\centering
\scriptsize
\setlength{\tabcolsep}{3pt}
\begin{tabular}{@{}l rr rr rr@{}}
\toprule
 & \multicolumn{2}{c}{ProTAS} & \multicolumn{2}{c}{Transformer} & \multicolumn{2}{c}{MiniROAD} \\
\cmidrule(lr){2-3}\cmidrule(lr){4-5}\cmidrule(lr){6-7}
Test schedule (R, A, G) & Acc & F1 & Acc & F1 & Acc & F1 \\
\midrule
\textit{$\Delta$ (mixture $-$ specialist)} & \multicolumn{6}{r@{}}{\textit{Specialist trained on 6\,FPS R+A+G}} \\
Regular, 6\,FPS (training schedule) & \cellcolor{red!70!black!30}$-$8.1 & \cellcolor{red!70!black!30}$-$5.4 & $-$1.4 & $-$1.7 & $-$1.3 & \cellcolor{red!70!black!12}$-$4.6 \\
\midrule
Regular, 1\,FPS & \cellcolor{red!70!black!30}$-$7.6 & \cellcolor{red!70!black!12}$-$3.4 & $-$0.6 & $-$1.6 & $-$0.6 & \cellcolor{red!70!black!12}$-$2.8 \\
Regular, 0.1\,FPS & \cellcolor{red!70!black!12}$-$3.6 & \cellcolor{red!70!black!12}$-$2.1 & $+$0.2 & $+$1.6 & $+$0.6 & $+$0.5 \\
\midrule
Gaze absent & $-$0.7 & $-$0.4 & \cellcolor{green!50!black!12}$+$4.3 & $-$0.9 & \cellcolor{green!50!black!30}$+$6.2 & $+$1.3 \\
RGB absent & $+$0.2 & $+$1.9 & \cellcolor{green!50!black!30}$+$8.7 & $-$0.7 & \cellcolor{green!50!black!30}$+$5.7 & $+$0.3 \\
Audio only & \cellcolor{green!50!black!30}$+$10.3 & \cellcolor{green!50!black!30}$+$6.2 & \cellcolor{green!50!black!30}$+$8.2 & $+$0.5 & \cellcolor{green!50!black!30}$+$12.6 & \cellcolor{green!50!black!12}$+$2.1 \\
Gaze only & $+$0.5 & \cellcolor{green!50!black!12}$+$3.6 & \cellcolor{green!50!black!30}$+$7.7 & $+$0.4 & \cellcolor{green!50!black!12}$+$4.1 & $+$0.1 \\
RGB only & $+$0.6 & $+$0.3 & \cellcolor{green!50!black!12}$+$3.9 & $-$0.5 & \cellcolor{green!50!black!12}$+$4.4 & $-$0.6 \\
\bottomrule
\end{tabular}
\par\vspace{2pt}
{\footnotesize R = RGB, A = Audio, G = Gaze. $\Delta$: mixture minus specialist tested on the same schedule. Shades: light for $|\Delta|\geq 2$, strong for $|\Delta|\geq 5$ points.}

\end{table}

\begin{table*}[t]
\caption{\textbf{Absolute scores behind \cref{tab:supp_mismatch_all_delta}} (all-sensors mixture; same rows and models as Table~\ref{tab:mismatch}; Ego-Exo4D, 15 scenarios).}
\label{tab:supp_mismatch_abs_all}
\centering
\scriptsize
\setlength{\tabcolsep}{2.4pt}
\begin{tabular}{@{}l rr rr rr @{\hspace{8pt}} rr rr rr @{\hspace{8pt}} rr rr rr@{}}
\toprule
 & \multicolumn{6}{c}{ProTAS} & \multicolumn{6}{c}{Transformer} & \multicolumn{6}{c}{MiniROAD} \\
\cmidrule(lr){2-7}\cmidrule(lr){8-13}\cmidrule(l){14-19}
 & \multicolumn{2}{c}{Acc} & \multicolumn{2}{c}{Edit} & \multicolumn{2}{c}{F1} & \multicolumn{2}{c}{Acc} & \multicolumn{2}{c}{Edit} & \multicolumn{2}{c}{F1} & \multicolumn{2}{c}{Acc} & \multicolumn{2}{c}{Edit} & \multicolumn{2}{c}{F1} \\
Test schedule (R, A, G) & Spec. & Mix. & Spec. & Mix. & Spec. & Mix. & Spec. & Mix. & Spec. & Mix. & Spec. & Mix. & Spec. & Mix. & Spec. & Mix. & Spec. & Mix. \\
\midrule
Regular, 6\,FPS (training schedule) & 48.0 & 40.0 & 29.2 & 25.6 & 25.9 & 20.6 & 53.4 & 52.1 & 12.1 & 9.8 & 11.6 & 9.9 & 56.3 & 55.0 & 25.5 & 20.7 & 25.7 & 21.1 \\
\midrule
Regular, 1\,FPS & 46.9 & 39.4 & 30.3 & 27.2 & 25.6 & 22.2 & 52.4 & 51.8 & 26.2 & 22.7 & 24.0 & 22.4 & 54.7 & 54.1 & 30.9 & 28.1 & 29.5 & 26.7 \\
Regular, 0.1\,FPS & 38.4 & 34.9 & 33.4 & 26.7 & 19.8 & 17.7 & 43.2 & 43.4 & 35.7 & 36.8 & 27.5 & 29.0 & 43.0 & 43.6 & 37.4 & 37.1 & 27.0 & 27.4 \\
\midrule
Gaze absent & 36.8 & 36.2 & 27.1 & 26.0 & 19.6 & 19.2 & 40.7 & 45.0 & 14.9 & 11.4 & 11.1 & 10.3 & 41.5 & 47.7 & 23.1 & 20.8 & 18.6 & 19.9 \\
RGB absent & 39.3 & 39.6 & 24.1 & 25.9 & 18.8 & 20.6 & 40.2 & 48.9 & 8.8 & 6.8 & 6.4 & 5.7 & 49.2 & 54.9 & 19.8 & 18.5 & 17.4 & 17.7 \\
Audio only & 15.8 & 26.1 & 16.7 & 21.7 & 7.0 & 13.2 & 16.9 & 25.1 & 10.3 & 8.8 & 4.4 & 4.9 & 13.8 & 26.4 & 7.5 & 13.4 & 3.3 & 5.5 \\
Gaze only & 38.6 & 39.1 & 21.0 & 24.3 & 16.0 & 19.7 & 40.9 & 48.6 & 6.6 & 6.3 & 4.3 & 4.7 & 49.5 & 53.6 & 16.3 & 16.1 & 16.1 & 16.2 \\
RGB only & 35.2 & 35.8 & 25.2 & 25.0 & 18.6 & 18.8 & 38.4 & 42.3 & 12.2 & 11.0 & 9.1 & 8.6 & 42.3 & 46.7 & 22.2 & 20.5 & 19.0 & 18.5 \\
\bottomrule
\end{tabular}
\par\vspace{2pt}
{\footnotesize R = RGB, A = Audio, G = Gaze. Spec.: specialist trained on 6\,FPS R+A+G; Mix.: all-sensors mixture, trained on randomized schedules of all five sensors.}

\end{table*}

\subsubsection{Mixture versus the 20\,mW specialist}
\label{supp:mismatch_20mw}
Table~\ref{tab:mismatch} uses the specialist trained on the 2.8\,W schedule; does the result hold for a specialist trained on a sparse, low-power schedule? \cref{tab:supp_gaps} repeats the comparison of Table~\ref{tab:mismatch} with the all-sensors mixture against the specialist trained on the 20\,mW schedule (MiniROAD architecture, Ego-Exo4D). On its own schedule (G+M 0.1\,FPS) the mixture is $+1.3$ Acc above the specialist, and it is above it in Acc in every condition: the gap is largest when the reads become denser ($+8.4$ at 6\,FPS, $+7.8$ at 1\,FPS) and close to zero at 0.05\,FPS ($+0.2$). At low budget, then, the mixture loses nothing on the specialist's own schedule and is more robust to schedule changes, a stronger result than at 2.8\,W, where the all-sensors mixture is $-1.3$ Acc below the specialist on its own schedule (MiniROAD, \cref{tab:supp_mismatch_all_delta}).
\begin{table}[t]
\caption{\textbf{All-sensors mixture versus the 20\,mW specialist under test-time schedule changes} (MiniROAD architecture, Ego-Exo4D, 15 scenarios; Acc of specialist and mixture; $\Delta$ = mixture minus specialist; wins: scenarios where the mixture is higher).}
\label{tab:supp_gaps}
\centering
\scriptsize
\setlength{\tabcolsep}{2.6pt}
\begin{tabular}{@{}l rr rrr@{}}
\toprule
Test schedule & Spec. & Mix. & $\Delta$Acc (wins) & $\Delta$Edit & $\Delta$F1 \\
\midrule
Train (G+M 0.1\,FPS) & 40.5 & 41.8 & +1.3 (8) & +3.3 & +0.8 \\
6\,FPS (G+M) & 46.1 & 54.5 & +8.4 (13) & +5.8 & +8.7 \\
1\,FPS (G+M) & 45.4 & 53.2 & +7.8 (13) & +6.8 & +8.7 \\
0.05\,FPS (G+M) & 36.5 & 36.6 & +0.2 (8) & +1.3 & +0.0 \\
Random reads (G+M) & 35.9 & 37.1 & +1.2 (9) & +2.9 & +1.7 \\
Gaze only & 37.7 & 41.1 & +3.5 (11) & +2.4 & {0.0} \\
Mono only & 28.2 & 30.3 & +2.2 (9) & +3.5 & +1.0 \\
+ IMU (G+M+I) & 40.5 & 43.2 & +2.8 (10) & +1.6 & -1.7 \\
\bottomrule
\end{tabular}

\end{table}

\subsection{Learned policies: training variants}
\label{supp:learned_results}

The main paper finds that learned policies do not beat periodic reads with the same reads (Table~\ref{tab:learn_where}). Those policies are trained with the cross-entropy against the labels and receive the time since the last read of each sensor (\cref{supp:learned_pol}). This section checks that the conclusion does not depend on these two choices (\cref{supp:learned_variants}) and reports AdaMML and HCMS trained end-to-end with ProTAS (\cref{supp:learned_e2e}).

\subsubsection{Training objective and time-since input}
\label{supp:learned_variants}
\paragraph{Training objective.} The cross-entropy needs frame labels. We also tried a label-free objective: the policy is trained to keep the prediction of the frozen TAS model close to its prediction with all sensors (KL divergence between the two), which would allow training the policy on unlabeled video. This is the objective of the oracle search (\cref{supp:oracle_search}), where it finds reads that raise accuracy, so it is a natural candidate for the learned policies. \cref{tab:supp_policy_obj} compares the two objectives on five scenarios. The objective changes little: KL minus cross-entropy is $-1.1$ Acc for the IMU-only policy, $+0.1$ for the HCMS-style cascade and $+0.8$ for the closed loop at 20\,mW, and $-0.2$, $+1.2$ and $0.0$ at 2.8\,W (differences of rounded table values). With either objective, no policy beats periodic reads with the same reads by more than 0.1 Acc at either budget, so the conclusion of Table~\ref{tab:learn_where} does not depend on the objective; the main paper reports the cross-entropy, the standard choice. At 2.8\,W, the calibration of the HCMS-style cascade {overshot the budget in three of the 15 scenarios (10, 14, and 15), so its power exceeds the budget there}; scenario 14, shown here, gives the HCMS-style cross-entropy row a power of 2.54\,W.

\paragraph{Time since last read.} The AdaMML-style and closed-loop policies receive the log time since the last read of each sensor (\cref{supp:learned_pol}), which tells them how stale each sensor is, the information a policy needs to time its reads. The rows ``since: no'' of \cref{tab:supp_policy_obj} remove this input. Without it the IMU-only policy loses 9.3 Acc with the cross-entropy and 7.8 with KL at 20\,mW, while the closed loop barely changes ($-0.3$ and $-0.4$), since it sees its own previous reads (differences of rounded table values). The input is therefore kept, and without it {the IMU-only policy falls further below periodic reads while the closed loop changes little}, so the conclusion of Table~\ref{tab:learn_where} holds in both cases.

\begin{table*}[t]
\caption{\textbf{Training objective and time-since-last-read input} (Ego-Exo4D, scenarios 1, 4, 11, 13, 14; frozen MiniROAD mixture; reference energy model; whole test videos). CE: cross-entropy with the labels (the objective of the main paper); KL: label-free alternative. since: time-since-last-read input. P in mW (20\,mW) and W (2.8\,W), including the policy network. $\Delta$: Acc minus periodic reads with the same reads (same runs). The HCMS-style cascade has no since input. $^{*}$Above the budget on average.}
\label{tab:supp_policy_obj}
\centering
\scriptsize
\setlength{\tabcolsep}{3.6pt}
\begin{tabular}{@{}lll rrrr rrrr@{}}
\toprule
 & & & \multicolumn{4}{c}{20\,mW} & \multicolumn{4}{c}{2.8\,W} \\
\cmidrule(lr){4-7}\cmidrule(lr){8-11}
Policy & Obj. & since & Acc & F1 & P & $\Delta$ & Acc & F1 & P & $\Delta$ \\
\midrule
IMU-only & CE & yes & 46.9 & 23.9 & 18.9 & -6.5 & 59.6 & 22.8 & 2.78 & -0.2 \\
IMU-only & KL & yes & 45.8 & 25.3 & 16.4 & -7.9 & 59.4 & 21.9 & 2.79 & -0.1 \\
IMU-only & CE & no & 37.6 & 19.3 & 14.6 & -14.2 & 59.4 & 23.3 & 2.80 & -0.5 \\
IMU-only & KL & no & 38.0 & 18.0 & 16.1 & -13.9 & 58.6 & 22.0 & 2.79 & -0.9 \\
\cmidrule(lr){1-11}
HCMS-style & CE & yes & 43.4 & 26.4 & 15.8 & -9.2 & 58.9 & 22.1 & 2.54 & -1.1 \\
HCMS-style & KL & yes & 43.5 & 22.6 & 15.3 & -8.7 & 60.1 & 22.3 & 2.80 & -0.0 \\
\cmidrule(lr){1-11}
Closed loop & CE & yes & 51.6 & 27.5 & 20.2$^{*}$ & -2.1 & 59.6 & 23.3 & 2.79 & -0.4 \\
Closed loop & KL & yes & 52.4 & 29.5 & 18.8 & -1.0 & 59.6 & 21.9 & 2.76 & +0.1 \\
Closed loop & CE & no & 51.3 & 29.0 & 18.9 & -1.1 & 60.1 & 23.3 & 2.81 & +0.1 \\
Closed loop & KL & no & 52.0 & 31.2 & 19.0 & -1.7 & 59.5 & 22.2 & 2.75 & -0.1 \\
\bottomrule
\end{tabular}

\end{table*}

\subsubsection{End-to-end AdaMML and HCMS}
\label{supp:learned_e2e}
\paragraph{Protocol closer to the original papers (no budget term).}
The policies of the main paper run on a frozen mixture, whereas the original AdaMML and HCMS train the policy together with the task model. Here we test whether an implementation closer to the originals changes the conclusion. We train AdaMML- and HCMS-style policies end-to-end with ProTAS and without a Lagrangian term: the policy and the TAS model are optimized jointly, and a budget is approached by adjusting a threshold over the gate predictions at inference (in brackets), so the realized power is not guaranteed. The policies follow the setup of AdaMML~\cite{panda2021adamml}: the TAS model is first warmed up for 5 epochs, after which AdaMML alternates policy and task-model training for 20 epochs, followed by 10 epochs of TAS fine-tuning. As in the original AdaMML, the policy routes three modalities, with a cheaper substitute for RGB: gaze on Ego-Exo4D and a downscaled RGB image on CMU-MMAC and CaptainCook4D. For HCMS~\cite{weng2024hcms}, since no specific training pipeline is provided, we replace the alternating scheme with joint training while keeping the remaining AdaMML training specifications unchanged. For AdaMML~\cite{panda2021adamml} and HCMS~\cite{weng2024hcms}, each trained policy produces a fixed trade-off between performance and energy consumption. Since the target energy budget is not enforced during training within their pipelines, we adapt these policies to different budget constraints at inference time by varying the threshold applied to the Gumbel-Softmax decisions~\cite{jang2016categorical}. This allows us to shift the modality-selection behavior of the policy and obtain operating points with different energy-accuracy trade-offs.

\cref{tab:supp_e2e_maria} reports the operating points we obtained at 2.8\,W and 20\,mW on the three datasets. Several operating points exceed the budget ($^\dagger$; AdaMML at 20\,mW, HCMS at 2.8\,W on Ego-Exo4D). {For reference, the table also reports the Blind baseline (no sensors) and the best Fixed schedule of Table~\ref{tab:all_protas}, both with ProTAS, for each dataset and budget. At 20\,mW the HCMS operating points are below the Blind baseline on all three datasets (17.8 vs.\ 22.3 on Ego-Exo4D, 42.8 vs.\ 66.3 on CMU-MMAC, 18.8 vs.\ 23.3 on CaptainCook4D).} These are different runs from the frozen-model policies of \cref{tab:learn_where} (backbone, training, and recipe differ), so the two sets of numbers are not comparable. {Every operating point either exceeds the budget or is below the best Fixed schedule of Table~\ref{tab:all_protas} for the same dataset and budget. Since the budget is not enforced during training and several points fall below the Blind baseline, we draw no conclusion on the policies from these runs and repeat the experiment with a budget term in the next paragraph.} 
\begin{table}[t]
\caption{\textbf{End-to-end AdaMML and HCMS with ProTAS} (threshold in brackets; Fixed and Blind: ProTAS specialists of Table~\ref{tab:all_protas}). $^\dagger$ exceeds the budget.}
\label{tab:supp_e2e_maria}
\centering
\scriptsize
\setlength{\tabcolsep}{4pt}
\begin{tabular}{@{}l rrrr@{}}
\toprule
 & Acc & Edit & F1@25 & P \\
\midrule
\multicolumn{5}{@{}l}{\textit{Ego-Exo4D, 2.8\,W}} \\
AdaMML (0.7) & 22.1 & 12.7 & 6.3 & 2.25\,W \\
HCMS (0.99) & 40.9 & 22.8 & 17.2 & 3.35\,W$^\dagger$ \\
Fixed (6\,FPS R+A+G) & 48.0 & 29.2 & 25.9 & 2.31\,W \\
Rule (ProTAS mix.) & 39.8 & 26.9 & 21.5 & 1.99\,W \\
\multicolumn{5}{@{}l}{\textit{20\,mW}} \\
AdaMML (1.0) & 18.1 & 5.4 & 2.9 & 2211.9\,mW$^\dagger$ \\
HCMS (1.0) & 17.8 & 15.6 & 6.3 & 5.0\,mW \\
Fixed (0.1\,FPS G+M) & 34.6 & 22.5 & 18.1 & 16.7\,mW \\
Rule (ProTAS mix.) & 37.4 & 26.9 & 20.4 & 19\,mW \\
Blind (no sensors) & 22.3 & 10.4 & 8.4 & 0.4\,mW \\
\midrule
\multicolumn{5}{@{}l}{\textit{CMU-MMAC, 2.8\,W}} \\
AdaMML & 66.6 & 20.0 & 10.8 & 2.21\,W \\
HCMS (0.9) & 83.9 & 30.1 & 32.5 & 2.61\,W \\
Fixed (6\,FPS R) & 84.3 & 39.1 & 40.7 & {1.86}\,W \\
Rule (ProTAS mix.) & 80.9 & 35.1 & 36.8 & 1.94\,W \\
\multicolumn{5}{@{}l}{\textit{20\,mW}} \\
AdaMML & 66.6 & 20.0 & 10.8 & 2213\,mW$^\dagger$ \\
HCMS (1.0) & 42.8 & 13.2 & 6.7 & 5.0\,mW \\
Fixed (0.05\,FPS all) & 70.8 & 24.2 & 26.8 & 16.8\,mW \\
Rule (ProTAS mix.) & 72.1 & 32.1 & 28.1 & 20\,mW \\
Blind (no sensors) & 66.3 & 0.0 & 0.0 & 0.4\,mW \\
\midrule
\multicolumn{5}{@{}l}{\textit{CaptainCook4D, 2.8\,W}} \\
AdaMML (0.9) & 23.9 & 6.0 & 1.3 & 2.17\,W \\
HCMS (0.9) & 42.1 & 17.7 & 11.4 & 2.52\,W \\
Fixed (0.5\,FPS R) & 43.5 & 38.2 & 23.1 & 0.15\,W \\
Rule (ProTAS mix.) & 47.3 & 33.0 & 20.6 & 1.91\,W \\
\multicolumn{5}{@{}l}{\textit{20\,mW}} \\
AdaMML (0.9) & 23.9 & 6.0 & 1.3 & 2168.5\,mW$^\dagger$ \\
HCMS (1.0) & 18.8 & 23.7 & 2.9 & 4.9\,mW \\
Fixed (0.05\,FPS R+A) & 40.3 & 31.9 & 17.7 & 16.0\,mW \\
Rule (ProTAS mix.) & 44.7 & 34.6 & 20.9 & 19\,mW \\
Blind (no sensors) & 23.3 & 9.7 & 1.2 & 0.35\,mW \\
\bottomrule
\end{tabular}

\end{table}

\paragraph{Our protocol (with a Lagrangian budget term).} Since the runs above leave the budget unenforced (AdaMML) or fall below Blind (HCMS), we repeat the experiment under the protocol of the main paper. We train the three policies of \cref{tab:learn_where} (AdaMML-style IMU-only policy, HCMS-style cascade, closed-loop policy) with the Lagrangian budget term of \cref{supp:learned_pol}, calibrate the decision threshold on the validation videos to 20\,mW, and evaluate them on the scenarios of the main paper with each of the three backbones. The policy and the backbone are trained jointly: AdaMML-style with alternating phases of 5, 20, and 10 epochs, HCMS-style and closed loop with joint training. \cref{tab:supp_e2e_matrix} reports the three datasets in the format of \cref{tab:all_protas}. The reference that matters is periodic reads with the same reads per sensor on the same backbone: on Ego-Exo4D, every policy is below it with all three backbones (9 of 9 cells; from $-0.6$ for the closed loop with MiniROAD to $-12.7$ for HCMS with the Transformer), while on CMU-MMAC and CaptainCook4D the policies are above it in Acc in 8 of 18 cells, by $+0.1$ to $+1.1$ points. In three of these cells the policy is also above in Edit and F1 (closed loop with MiniROAD on CMU-MMAC, and with the Transformer and ProTAS on CaptainCook4D, which has one scenario). In the other five, the gain in Acc comes with a loss in Edit or F1: the closed loop with ProTAS on CMU-MMAC ($+0.6$ Acc, $+2.5$ F1, $-1.5$ Edit), the closed loop with the Transformer on CMU-MMAC ($+0.2$ Acc, $+0.2$ Edit, $-1.0$ F1), the closed loop with MiniROAD on CaptainCook4D ($+0.7$ Acc, $-0.4$ Edit, $-0.3$ F1), and HCMS on CMU-MMAC with MiniROAD ($+1.1$ Acc, $-2.9$ Edit, $-6.4$ F1) and ProTAS ($+0.8$ Acc, $-1.6$ Edit, $-1.9$ F1). All eight cells are on the two datasets with few scenarios (CMU-MMAC: 5, CaptainCook4D: 1). With MiniROAD on Ego-Exo4D, the closed loop (48.6) also matches the budget rule (48.8), which needs no learning, and it is above the best Fixed schedule (40.5), a weaker baseline than the rule. The policies are above the budget rule in three cells: the closed loop with ProTAS on Ego-Exo4D (42.0 vs.\ 37.4), which is not like-for-like because it spends 22.8\,mW, above the budget, and is itself below periodic reads with the same reads (43.0); HCMS with ProTAS on Ego-Exo4D (37.8 vs.\ 37.4); and the closed loop with MiniROAD on CaptainCook4D (49.0 vs.\ 48.2, one scenario). Several cells exceed the budget ($^\dagger$). Joint training of the backbone therefore does not make the learned reads better than periodic reads.
\begin{table*}[t]
\caption{{\textbf{End-to-end read policies with a budget term} (20\,mW, reference model; policy and backbone trained jointly; P in mW). \emph{periodic}: periodic reads with the same reads per sensor as the policy above, on the same jointly trained backbone. \emph{Fixed}: best Fixed schedule of the backbone (Table~\ref{tab:all_protas} for ProTAS, \cref{tab:supp_testra} for MiniROAD and the Transformer). \emph{Budget rule}: mixture-trained backbone with the budget rule (Table~\ref{tab:backbone_training}). $^\dagger$ mean power above 20\,mW.}}
\label{tab:supp_e2e_matrix}
\centering
\scriptsize
\setlength{\tabcolsep}{2.2pt}
\begin{tabular}[t]{@{}l rrr r@{}}
\toprule
\multicolumn{5}{c}{\textbf{Ego-Exo4D}} \\
\midrule
Policy & Acc & Edit & F1 & P \\
\midrule
\multicolumn{5}{@{}l}{\textit{MiniROAD}} \\
Fixed (0.1\,FPS G+M) & 40.5 & 33.4 & 25.4 & 17.2 \\
Budget rule & 48.8 & 32.7 & 27.4 & 20 \\
\cmidrule(lr){1-5}
AdaMML & 41.7 & 25.8 & 18.8 & 18.3 \\
\quad periodic & 48.7 & 29.1 & 23.2 & 18.3 \\
HCMS & 41.1 & 25.8 & 19.0 & 18.1 \\
\quad periodic & 48.4 & 27.8 & 23.8 & 18.1 \\
Closed loop & 48.6 & 30.0 & 25.6 & 18.1 \\
\quad periodic & 49.1 & 30.9 & 26.2 & 18.1 \\
\midrule
\multicolumn{5}{@{}l}{\textit{ProTAS}} \\
Fixed (0.1\,FPS G+M) & 34.6 & 22.5 & 18.1 & 16.7 \\
Budget rule & 37.4 & 26.9 & 20.4 & 19 \\
\cmidrule(lr){1-5}
AdaMML & 36.5 & 26.8 & 19.1 & 17.6 \\
\quad periodic & 39.8 & 26.9 & 20.9 & 17.6 \\
HCMS & 37.8 & 28.7 & 19.3 & 14.9 \\
\quad periodic & 40.7 & 29.2 & 22.8 & 14.9 \\
Closed loop & 42.0 & 29.8 & 23.2 & 22.8\smash{$^\dagger$} \\
\quad periodic & 43.0 & 30.3 & 22.7 & 22.8 \\
\midrule
\multicolumn{5}{@{}l}{\textit{Transformer}} \\
Fixed (0.1\,FPS G+M) & 38.7 & 31.8 & 24.9 & 16.9 \\
Budget rule & 45.2 & 24.0 & 18.8 & 19 \\
\cmidrule(lr){1-5}
AdaMML & 31.3 & 13.4 & 7.4 & 18.0 \\
\quad periodic & 41.4 & 16.7 & 12.6 & 18.0 \\
HCMS & 26.7 & 16.6 & 10.3 & 16.5 \\
\quad periodic & 39.4 & 18.4 & 14.0 & 16.5 \\
Closed loop & 35.7 & 16.6 & 11.8 & 24.5\smash{$^\dagger$} \\
\quad periodic & 38.9 & 15.8 & 10.8 & 24.5 \\
\bottomrule
\end{tabular}
\hfill
\begin{tabular}[t]{@{}l rrr r@{}}
\toprule
\multicolumn{5}{c}{\textbf{CMU-MMAC}} \\
\midrule
Policy & Acc & Edit & F1 & P \\
\midrule
\multicolumn{5}{@{}l}{\textit{MiniROAD}} \\
Fixed (0.05\,FPS R+A+I) & 71.8 & 24.4 & 26.9 & 16.8 \\
Budget rule & 73.2 & 33.6 & 27.9 & 20 \\
\cmidrule(lr){1-5}
AdaMML & 69.9 & 19.4 & 12.6 & 31.4\smash{$^\dagger$} \\
\quad periodic & 73.8 & 18.0 & 15.9 & 31.4 \\
HCMS & 71.1 & 20.2 & 12.5 & 20.9\smash{$^\dagger$} \\
\quad periodic & 70.0 & 23.1 & 18.9 & 20.9 \\
Closed loop & 71.6 & 30.5 & 25.2 & 18.6 \\
\quad periodic & 71.0 & 29.8 & 21.4 & 18.6 \\
\midrule
\multicolumn{5}{@{}l}{\textit{ProTAS}} \\
Fixed (0.05\,FPS all) & 70.8 & 24.2 & 26.8 & 16.8 \\
Budget rule & 72.1 & 32.1 & 28.1 & 20 \\
\cmidrule(lr){1-5}
AdaMML & 68.4 & 24.6 & 19.2 & 23.4\smash{$^\dagger$} \\
\quad periodic & 71.4 & 27.8 & 25.0 & 23.4 \\
HCMS & 68.7 & 25.4 & 15.2 & 15.1 \\
\quad periodic & 68.0 & 27.0 & 17.1 & 15.1 \\
Closed loop & 71.2 & 27.2 & 23.4 & 18.8 \\
\quad periodic & 70.6 & 28.7 & 20.9 & 18.8 \\
\midrule
\multicolumn{5}{@{}l}{\textit{Transformer}} \\
Fixed (0.05\,FPS R+A+I) & 72.1 & 26.6 & 30.2 & 17.0 \\
Budget rule & 74.5 & 22.9 & 21.0 & 20 \\
\cmidrule(lr){1-5}
AdaMML & 68.2 & 15.2 & 8.3 & 14.5 \\
\quad periodic & 70.6 & 14.4 & 10.0 & 14.5 \\
HCMS & 67.5 & 12.2 & 6.0 & 15.2 \\
\quad periodic & 73.1 & 16.1 & 14.7 & 15.2 \\
Closed loop & 68.1 & 14.2 & 7.7 & 16.3 \\
\quad periodic & 67.9 & 14.0 & 8.8 & 16.3 \\
\bottomrule
\end{tabular}
\hfill
\begin{tabular}[t]{@{}l rrr r@{}}
\toprule
\multicolumn{5}{c}{\textbf{CaptainCook4D}} \\
\midrule
Policy & Acc & Edit & F1 & P \\
\midrule
\multicolumn{5}{@{}l}{\textit{MiniROAD}} \\
Fixed (0.05\,FPS R+A) & 46.8 & 37.0 & 26.7 & 16.3 \\
Budget rule & 48.2 & 31.2 & 20.9 & 19 \\
\cmidrule(lr){1-5}
AdaMML & 39.9 & 18.3 & 8.9 & 13.7 \\
\quad periodic & 47.5 & 18.5 & 14.1 & 13.7 \\
HCMS & 34.5 & 15.8 & 5.8 & 15.2 \\
\quad periodic & 44.2 & 14.9 & 12.4 & 15.2 \\
Closed loop & 49.0 & 19.5 & 14.3 & 18.2 \\
\quad periodic & 48.4 & 19.9 & 14.6 & 18.2 \\
\midrule
\multicolumn{5}{@{}l}{\textit{ProTAS}} \\
Fixed (0.05\,FPS R+A) & 40.3 & 31.9 & 17.7 & 16.0 \\
Budget rule & 44.7 & 34.6 & 20.9 & 19 \\
\cmidrule(lr){1-5}
AdaMML & 24.1 & 22.7 & 11.1 & 14.7 \\
\quad periodic & 39.9 & 26.2 & 15.4 & 14.7 \\
HCMS & 35.3 & 27.8 & 13.2 & 9.7 \\
\quad periodic & 39.5 & 26.5 & 15.7 & 9.7 \\
Closed loop & 42.0 & 24.7 & 17.1 & 15.7 \\
\quad periodic & 42.0 & 24.6 & 16.7 & 15.7 \\
\midrule
\multicolumn{5}{@{}l}{\textit{Transformer}} \\
Fixed (0.05\,FPS R+A) & 45.6 & 38.1 & 23.5 & 16.0 \\
Budget rule & 47.7 & 13.5 & 10.3 & 19 \\
\cmidrule(lr){1-5}
AdaMML & 37.4 & 5.8 & 4.3 & 12.7 \\
\quad periodic & 43.8 & 7.7 & 5.8 & 12.7 \\
HCMS & 33.5 & 6.0 & 3.2 & 1.8 \\
\quad periodic & 36.1 & 10.2 & 6.0 & 1.8 \\
Closed loop & 44.2 & 9.0 & 6.2 & 19.8 \\
\quad periodic & 43.3 & 7.1 & 4.8 & 19.8 \\
\bottomrule
\end{tabular}
\end{table*}

\subsubsection{Frozen versus end-to-end policies on other datasets}
\label{supp:learned_frozen}
The policies of the main paper are trained on a frozen mixture backbone. \cref{tab:supp_frozen_e2e} reports, with MiniROAD on the three datasets, the three policies trained on the frozen mixture backbone with the Lagrangian budget term (\cref{supp:learned_pol}) and the same policies trained jointly with the backbone (\cref{tab:supp_e2e_matrix}). With a frozen backbone, every policy is below periodic reads with the same reads (9 of 9 cells, from {$-1.0$} to $-9.8$ Acc; differences of rounded table values), as in Table~\ref{tab:learn_where}. Training the backbone jointly changes the accuracy of the policy by $-3.3$ to $+7.3$ points with no consistent sign, and the gap to periodic reads does not consistently shrink: {on Ego-Exo4D it is similar for AdaMML ($-7.0$ vs.\ $-7.4$) and HCMS ($-7.3$ vs.\ $-6.9$) and smaller for the closed loop ($-0.5$ vs.\ $-1.7$)}, while on CMU-MMAC and CaptainCook4D the closed loop improves (from $-1.1$ and $-2.4$ to $+0.6$ and $+0.6$), HCMS improves on CMU-MMAC ($-1.0$ to $+1.1$), and AdaMML worsens ($-3.4$ to $-3.9$ and $-6.1$ to $-7.6$).
\begin{table*}[t]
\caption{\textbf{Policies with MiniROAD: frozen versus jointly trained backbone} (20\,mW, reference model; Lagrangian budget term; P in mW). \emph{Frozen backbone}: policies trained on the frozen mixture backbone; \emph{End-to-end}: policy and backbone trained jointly (as \cref{tab:supp_e2e_matrix}). \emph{periodic}: periodic reads with the same reads per sensor as the policy above, on the same backbone. $^\dagger$ mean power above 20\,mW.}
\label{tab:supp_frozen_e2e}
\centering
\scriptsize
\setlength{\tabcolsep}{2.2pt}
\begin{tabular}[t]{@{}l rrr r@{}}
\toprule
\multicolumn{5}{c}{\textbf{Ego-Exo4D}} \\
\midrule
Policy & Acc & Edit & F1 & P \\
\midrule
Fixed (0.1\,FPS G+M) & 40.5 & 33.4 & 25.4 & 17.2 \\
Budget rule & 48.8 & 32.7 & 27.4 & 20 \\
\midrule
\multicolumn{5}{@{}l}{\textit{Frozen backbone}} \\
AdaMML & 40.0 & 31.3 & 21.1 & 18.5 \\
\quad periodic & 47.4 & 33.1 & 26.5 & 18.5 \\
HCMS & 40.2 & 29.4 & 21.9 & 19.3 \\
\quad periodic & 47.1 & 32.0 & 27.1 & 19.3 \\
Closed loop & 46.0 & 31.4 & 25.8 & 21.7\smash{$^\dagger$} \\
\quad periodic & 47.7 & 32.1 & 26.4 & 21.7 \\
\midrule
\multicolumn{5}{@{}l}{\textit{End-to-end}} \\
AdaMML & 41.7 & 25.8 & 18.8 & 18.3 \\
\quad periodic & 48.7 & 29.1 & 23.2 & 18.3 \\
HCMS & 41.1 & 25.8 & 19.0 & 18.1 \\
\quad periodic & 48.4 & 27.8 & 23.8 & 18.1 \\
Closed loop & 48.6 & 30.0 & 25.6 & 18.1 \\
\quad periodic & 49.1 & 30.9 & 26.2 & 18.1 \\
\bottomrule
\end{tabular}
\hfill
\begin{tabular}[t]{@{}l rrr r@{}}
\toprule
\multicolumn{5}{c}{\textbf{CMU-MMAC}} \\
\midrule
Policy & Acc & Edit & F1 & P \\
\midrule
Fixed (0.05\,FPS R+A+I) & 71.8 & 24.4 & 26.9 & 16.8 \\
Budget rule & 73.2 & 33.6 & 27.9 & 20 \\
\midrule
\multicolumn{5}{@{}l}{\textit{Frozen backbone}} \\
AdaMML & 66.1 & 20.1 & 16.1 & 16.2 \\
\quad periodic & 69.5 & 32.2 & 19.5 & 16.2 \\
HCMS & 64.7 & 13.8 & 8.7 & 8.4 \\
\quad periodic & 65.7 & 28.6 & 10.8 & 8.4 \\
Closed loop & 72.8 & 31.3 & 27.7 & 21.0\smash{$^\dagger$} \\
\quad periodic & 73.9 & 33.5 & 25.0 & 21.0 \\
\midrule
\multicolumn{5}{@{}l}{\textit{End-to-end}} \\
AdaMML & 69.9 & 19.4 & 12.6 & 31.4\smash{$^\dagger$} \\
\quad periodic & 73.8 & 18.0 & 15.9 & 31.4 \\
HCMS & 71.1 & 20.2 & 12.5 & 20.9\smash{$^\dagger$} \\
\quad periodic & 70.0 & 23.1 & 18.9 & 20.9 \\
Closed loop & 71.6 & 30.5 & 25.2 & 18.6 \\
\quad periodic & 71.0 & 29.8 & 21.4 & 18.6 \\
\bottomrule
\end{tabular}
\hfill
\begin{tabular}[t]{@{}l rrr r@{}}
\toprule
\multicolumn{5}{c}{\textbf{CaptainCook4D}} \\
\midrule
Policy & Acc & Edit & F1 & P \\
\midrule
Fixed (0.05\,FPS R+A) & 46.8 & 37.0 & 26.7 & 16.3 \\
Budget rule & 48.2 & 31.2 & 20.9 & 19 \\
\midrule
\multicolumn{5}{@{}l}{\textit{Frozen backbone}} \\
AdaMML & 43.2 & 28.6 & 16.8 & 16.7 \\
\quad periodic & 49.3 & 28.4 & 19.3 & 16.7 \\
HCMS & 35.5 & 26.3 & 9.5 & 11.1 \\
\quad periodic & 45.3 & 28.6 & 17.9 & 11.1 \\
Closed loop & 41.7 & 27.8 & 12.1 & 15.2 \\
\quad periodic & 44.1 & 26.6 & 13.8 & 15.2 \\
\midrule
\multicolumn{5}{@{}l}{\textit{End-to-end}} \\
AdaMML & 39.9 & 18.3 & 8.9 & 13.7 \\
\quad periodic & 47.5 & 18.5 & 14.1 & 13.7 \\
HCMS & 34.5 & 15.8 & 5.8 & 15.2 \\
\quad periodic & 44.2 & 14.9 & 12.4 & 15.2 \\
Closed loop & 49.0 & 19.5 & 14.3 & 18.2 \\
\quad periodic & 48.4 & 19.9 & 14.6 & 18.2 \\
\bottomrule
\end{tabular}
\end{table*}

\subsection{Read timing: window length, other models, and learnability}
\label{supp:add_policies}

Part (b) of Table~\ref{tab:learn_where} asks how much accuracy the timing of the reads can add when the number of reads is fixed. The oracle search and the causal policy used there are described in \cref{supp:read_timing}. This section checks two properties of the oracle's gain: how it depends on the window length and on the model, and whether the oracle's read decisions can be predicted causally.

\subsubsection{Oracle windows and transfer of the oracle's reads}
\label{supp:oracle_windows}
\cref{tab:supp_transfer_b2} gives the accuracy of the reads found by the oracle search with the MiniROAD mixture, for windows of 2, 5 and 20\,s, and the same reads scored with the ProTAS and Transformer mixtures (15 scenarios, whole test videos, reference model). At 20\,mW the oracle gains $+5.9$ Acc over periodic reads with the MiniROAD mixture (5\,s window); shorter windows gain less ($+4.3$ at 2\,s) and the longer one no more ($+6.1$ at 20\,s). The reads transfer to the Transformer ($+3.6$) but much less to ProTAS ($+1.2$): part of the headroom is specific to the model that searched them. At 2.8\,W no window gains more than {$0.1$} points with any model. The headroom of timing is thus limited to low budgets and depends on the model, as stated in the main paper.
\begin{table}[t]
\caption{\textbf{Oracle windows and transfer of the oracle's reads} (Acc; Ego-Exo4D, 15 scenarios, whole test videos, reference model). Reads searched with the MiniROAD mixture and applied unchanged to the other mixtures. At 2.8\,W, the reads are those of the budget rule (RGB read at 3\,FPS, all other sensors at 6\,FPS; 2.00\,W).}
\label{tab:supp_transfer_b2}
\centering
\scriptsize
\setlength{\tabcolsep}{2.4pt}
\begin{tabular}{@{}ll r rrr r@{}}
\toprule
 & Model & Periodic & 2\,s & 5\,s & 20\,s & 5\,s $-$ per. \\
\midrule
20\,mW & MiniROAD (searched with it) & 48.8 & 53.1 & 54.7 & 54.9 & +5.9 \\
 & ProTAS & 38.4 & 39.3 & 39.6 & 39.9 & +1.2 \\
 & Transformer & 45.5 & 48.0 & 49.0 & 48.9 & +3.6 \\
\midrule
2.8\,W & MiniROAD (searched with it) & {55.4} & {55.5} & {55.5} & {55.5} & {+0.0} \\
 & ProTAS & {40.4} & {40.5} & {40.4} & {40.4} & {+0.0} \\
 & Transformer & {51.5} & {51.6} & {51.6} & {51.6} & {+0.0} \\
\bottomrule
\end{tabular}

\end{table}

\subsubsection{Learnability of the oracle's reads}
\label{supp:oracle_learnability}
A causal predictor of the oracle's read decisions reaches an AUROC of 0.828 at 20\,mW and 0.716 at 2.8\,W from the time since the last read and the position in the window alone (MiniROAD mixture, whole videos, 15 scenarios, two predictor seeds). Adding the IMU features or the features of all sensors changes it by at most 0.006 (ranges over scenarios 0.81--0.84 at 20\,mW and 0.70--0.73 at 2.8\,W): the sensors do not tell when to read. This is consistent with the causal policy of Table~\ref{tab:learn_where} not recovering the oracle's gain.

\subsection{{Results under an edge-SoC energy model}}
\label{supp:add_edge}
The main paper summarises the edge-SoC results on Ego-Exo4D in Table~\ref{tab:edge_small}. This section reports the full tables behind that summary, computed with the edge-SoC energy model (0.4\,pJ/MAC, 60\,pJ/byte, capture unchanged; \cref{supp:energy_grounding}) and its three budgets (5\,mW, 20\,mW, 2.8\,W). Each table has the layout of a main-paper table, and the paragraphs below follow the main order: sensing strategies (Table~\ref{tab:supp_edge_t1}, as Table~\ref{tab:all_protas}), TAS models and training (Table~\ref{tab:supp_edge_t2}, as Table~\ref{tab:backbone_training}), and learned policies and read timing (Table~\ref{tab:supp_edge_t4}, as Table~\ref{tab:learn_where}). Table~\ref{tab:mismatch} does not depend on power, so it is not repeated.

\subsubsection{Sensing strategies}
\label{supp:edge_strategies}
Table~\ref{tab:supp_edge_t1} confirms the findings of Table~\ref{tab:all_protas}: on the three datasets, the most accurate Fixed configuration is the most accurate strategy in all nine dataset-budget blocks, and Greedy, Random, and Random-CA do not improve on it. Random-CA comes closest in some blocks (e.g., Ego-Exo4D 5\,mW: 34.0 vs.\ 34.6 Acc). The difference is that the cheaper hardware changes which configuration fits each budget (at 5 and 20\,mW, {one to three sensors at 0.05--1\,FPS, or the IMU alone at 30\,FPS}), and Greedy needs a 30\,s budget window to afford even one RGB+Audio+IMU frame (81\,mJ) at 5 and 20\,mW.

\subsubsection{TAS models and training}
\label{supp:edge_models}
Table~\ref{tab:supp_edge_t2} confirms the findings of Table~\ref{tab:backbone_training} on Ego-Exo4D: MiniROAD is the most accurate TAS model in every block (56.2, 51.7, and 48.7 Acc at 2.8\,W, 20\,mW, and 5\,mW), and the budget rule helps the mixture most at low budgets (+6.9 at 5\,mW, 48.7 vs.\ 41.8; +0.5 and +0.3 at 20\,mW and 2.8\,W). The difference is that, at 2.8\,W, the budget does not bind and the mixture with the rule stays 0.7 points below its specialist (55.5 vs.\ 56.2). On CMU-MMAC and CaptainCook4D, MiniROAD is again the most accurate backbone in every block, and the budget rule helps the mixture at 5\,mW in seven of nine model-dataset pairs (exceptions: ProTAS and MiniROAD on CaptainCook4D). \Cref{supp:energy_accuracy} plots accuracy against edge-SoC power for the three datasets (\cref{fig:tradeoff_edge,fig:tradeoff_edge_cmu,fig:tradeoff_edge_capt}).

\subsubsection{Learned policies and read timing}
\label{supp:edge_policies}
Table~\ref{tab:supp_edge_t4} confirms the findings of Table~\ref{tab:learn_where} on Ego-Exo4D with the MiniROAD mixture. Every learned policy is below the budget rule at 5 and 20\,mW and below periodic reads with its own read counts, while spending more power (e.g., closed loop: 46.4 and 50.3 Acc vs.\ 46.6 and 51.7 for periodic reads). Random reads and reads at true boundaries do not improve on periodic reads, and the causal policy imitating the oracle loses 1.2 and 1.5 points; only the non-causal oracle improves on them ($+$6.0 and $+$2.9 points at 5 and 20\,mW). The difference is that at 2.8\,W the budget does not bind, so every row reaches 55.5 Acc at 734--735\,mW, except HCMS-style and its periodic counterpart (54.2 at 621\,mW).

\begin{table*}[t]
\caption{{\textbf{Sensing strategies with ProTAS, edge-SoC energy model} (0.4\,pJ/MAC, 60\,pJ/byte, capture unchanged). Each Fixed configuration is listed in the tightest budget it fits (2.8\,W: $20\,\text{mW}<P\le 2.8$\,W; 20\,mW: $5<P\le 20$\,mW; 5\,mW: $P\le 5$\,mW), the three most accurate per block. Bold: best within the budget. Greedy uses a 30\,s budget window at 5 and 20\,mW and a 1\,s window at 2.8\,W: one RGB+Audio+IMU frame costs 81\,mJ under the edge-SoC model, which a 1\,s window cannot afford at the lower budgets.}
}
\centering
\scriptsize
\setlength{\tabcolsep}{3.4pt}
\begin{tabular}[t]{@{}l rrr r@{}}
\toprule
\multicolumn{5}{c}{\textbf{Ego-Exo4D}} \\
\midrule
Policy & Acc & Edit & F1 & P \\
\midrule
\multicolumn{5}{@{}l}{\textit{Budget: 2.8\,W} (P in W)} \\
Fixed (30\,FPS A+I+G+M) & \textbf{49.7} & 23.0 & 20.6 & 1.34 \\
Fixed (30\,FPS G+M) & 48.5 & 21.2 & 19.3 & 1.24 \\
Fixed (6\,FPS R+A+G) & 48.0 & \textbf{29.2} & \textbf{25.9} & 0.58 \\
\cmidrule(lr){1-5}
Greedy & 49.1 & 26.3 & 21.2 & 2.70 \\
Random (0.9/0.7) & 48.2 & 26.2 & 21.3 & 1.11 \\
Random-CA (0.7/0.7) & 48.1 & 27.4 & 22.9 & 0.64 \\
Random (0.7/0.24) & 47.0 & 24.5 & 21.2 & 2.79 \\
Random-CA (0.7/0.0) & 48.2 & 24.9 & 21.3 & 2.14 \\
\midrule
\multicolumn{5}{@{}l}{\textit{Budget: 20\,mW} (P in mW)} \\
Fixed (0.45\,FPS G+M) & \textbf{42.8} & 29.5 & \textbf{24.9} & 18.7 \\
Fixed (0.3\,FPS G+M) & 39.3 & 27.7 & 22.6 & 12.5 \\
Fixed (0.2\,FPS R+A+G) & 39.2 & 26.9 & 21.2 & 19.6 \\
\cmidrule(lr){1-5}
Greedy & 30.1 & 21.8 & 13.3 & 16.5 \\
Random (0.99/0.995) & 40.2 & \textbf{30.0} & 20.1 & 18.3 \\
Random-CA (0.99/0.99) & 38.5 & 27.7 & 19.5 & 11.1 \\
Random-CA (0.99/0.991) & 41.0 & 28.2 & 20.9 & 19.4 \\
\midrule
\multicolumn{5}{@{}l}{\textit{Budget: 5\,mW} (P in mW)} \\
Fixed (0.1\,FPS G+M) & \textbf{34.6} & 22.5 & \textbf{18.1} & 4.2 \\
Fixed (0.05\,FPS G+M) & 30.7 & 16.8 & 12.8 & 2.2 \\
Fixed (30\,FPS I) & 28.4 & 21.2 & 15.1 & 0.9 \\
\cmidrule(lr){1-5}
Greedy & 30.4 & 19.6 & 12.6 & 4.2 \\
Random (0.99/0.999) & 33.4 & \textbf{26.0} & 13.9 & 4.8 \\
Random-CA (0.99/0.998) & 34.0 & 25.8 & 14.6 & 4.7 \\
\midrule
Blind (no sensors) & 22.3 & 10.4 & 8.4 & 0.2 \\
\bottomrule
\end{tabular}
\hfill
\begin{tabular}[t]{@{}l rrr r@{}}
\toprule
\multicolumn{5}{c}{\textbf{CMU-MMAC}} \\
\midrule
Policy & Acc & Edit & F1 & P \\
\midrule
\multicolumn{5}{@{}l}{\textit{Budget: 2.8\,W} (P in W)} \\
Fixed (30\,FPS R+A) & \textbf{85.6} & 30.4 & 32.8 & 2.43 \\
Fixed (30\,FPS R) & 85.0 & 31.7 & 36.4 & 2.33 \\
Fixed (15\,FPS R) & 84.9 & \textbf{32.9} & \textbf{36.5} & 1.22 \\
\cmidrule(lr){1-5}
Greedy & 81.4 & 27.4 & 26.5 & 2.43 \\
Random (0.7/0.1) & 81.8 & 28.0 & 27.7 & 2.19 \\
Random-CA (0.7/0.7) & 80.8 & 26.8 & 26.3 & 0.32 \\
Random (0.7/0) & 81.8 & 28.0 & 28.1 & 2.43 \\
Random-CA (0.9/0.1) & 81.1 & 26.8 & 28.5 & 0.95 \\
\midrule
\multicolumn{5}{@{}l}{\textit{Budget: 20\,mW} (P in mW)} \\
Fixed (0.24\,FPS R+A) & \textbf{79.8} & \textbf{44.4} & \textbf{43.6} & 19.5 \\
Fixed (0.25\,FPS R) & 79.5 & 40.5 & 41.5 & 19.5 \\
Fixed (0.1\,FPS R+A) & 74.1 & 34.6 & 34.6 & 8.2 \\
Fixed (0.1\,FPS R+A+I) & 73.7 & 35.1 & 32.8 & 8.2 \\
Fixed (0.1\,FPS R+I) & 73.7 & 36.3 & 35.6 & 7.9 \\
\cmidrule(lr){1-5}
Greedy & 67.3 & 18.4 & 15.2 & 19.0 \\
Random (0.97/0.992) & 76.8 & 33.8 & 33.0 & 19.8 \\
Random-CA (0.9/0.982) & 76.3 & 28.2 & 27.7 & 19.8 \\
\midrule
\multicolumn{5}{@{}l}{\textit{Budget: 5\,mW} (P in mW)} \\
Fixed (0.05\,FPS R+A+I) & \textbf{70.8} & 24.2 & \textbf{26.8} & 4.2 \\
Fixed (0.05\,FPS R+A) & 70.3 & 24.6 & 24.9 & 4.2 \\
Fixed (0.05\,FPS R) & 69.4 & 24.1 & 23.8 & 4.0 \\
\cmidrule(lr){1-5}
Greedy & 66.9 & 14.9 & 13.3 & 2.9 \\
Random (0.97/0.998) & 68.0 & \textbf{30.7} & 20.7 & 4.5 \\
Random-CA (0.99/0.997) & 70.6 & 21.0 & 19.5 & 3.7 \\
\midrule
Blind (no sensors) & 66.3 & 0.0 & 0.0 & 0.2 \\
\bottomrule
\end{tabular}
\hfill
\begin{tabular}[t]{@{}l rrr r@{}}
\toprule
\multicolumn{5}{c}{\textbf{CaptainCook4D}} \\
\midrule
Policy & Acc & Edit & F1 & P \\
\midrule
\multicolumn{5}{@{}l}{\textit{Budget: 2.8\,W} (P in W)} \\
Fixed (15\,FPS R+A) & \textbf{44.7} & 17.5 & 12.9 & 1.19 \\
Fixed (5\,FPS R+A+I) & \textbf{44.7} & 25.8 & 20.5 & 0.40 \\
Fixed (0.5\,FPS R) & 43.5 & \textbf{38.2} & \textbf{23.1} & 0.04 \\
\cmidrule(lr){1-5}
Greedy & 39.5 & 13.4 & 12.0 & 2.38 \\
Random (0.7/0.9) & 31.3 & 18.8 & 15.3 & 0.24 \\
Random-CA (0.97/0.9) & 44.0 & 26.3 & 20.1 & 0.10 \\
\midrule
\multicolumn{5}{@{}l}{\textit{Budget: 20\,mW} (P in mW)} \\
Fixed (0.24\,FPS R+A) & \textbf{42.1} & 29.4 & 19.4 & 19.5 \\
Fixed (0.25\,FPS R) & 38.9 & 30.4 & \textbf{25.7} & 19.5 \\
Fixed (0.1\,FPS R+I) & 39.2 & 35.5 & 23.7 & 7.7 \\
Fixed (0.1\,FPS R) & 39.2 & 29.9 & 19.6 & 7.7 \\
Fixed (0.1\,FPS R+A+I) & 38.8 & \textbf{37.2} & 21.0 & 8.0 \\
\cmidrule(lr){1-5}
Greedy & 37.0 & 25.9 & 22.5 & 18.7 \\
Random (0.9/0.993) & 29.4 & 29.2 & 21.4 & 17.9 \\
Random-CA (0.99/0.99) & 29.0 & 31.0 & 20.9 & 10.7 \\
\midrule
\multicolumn{5}{@{}l}{\textit{Budget: 5\,mW} (P in mW)} \\
Fixed (0.05\,FPS R+A) & \textbf{40.3} & 31.9 & 17.7 & 4.0 \\
Fixed (1\,FPS A+I) & 38.9 & 20.0 & 13.4 & 3.2 \\
Fixed (0.05\,FPS R) & 38.3 & 26.1 & 17.8 & 3.8 \\
\cmidrule(lr){1-5}
Greedy & 33.6 & 36.2 & 20.6 & 2.8 \\
Random (0.99/0.9985) & 28.6 & \textbf{38.3} & 20.5 & 3.8 \\
Random-CA (0.99/0.996) & 28.0 & 32.8 & \textbf{20.9} & 4.3 \\
\midrule
Blind (no sensors) & 23.3 & 9.7 & 1.2 & 0.2 \\
\bottomrule
\end{tabular}

\vspace{2pt}
{\footnotesize
\textbf{Sensors:} R = RGB, A = Audio, I = IMU, G = Gaze, M = Monochrome, all = all sensors. \textbf{P:} average power under the edge-SoC model. Random and Random-CA: (training/inference dropout).}
\label{tab:supp_edge_t1}
\end{table*}

\begin{table*}[t]
\caption{\textbf{TAS models and training under the edge-SoC energy model} (same layout as Table~\ref{tab:backbone_training}; 0.4\,pJ/MAC, 60\,pJ/byte). As in the main paper, MiniROAD is the most accurate model in every block, and at 5\,mW the budget rule raises the accuracy of the mixture in seven of nine model-dataset pairs (exceptions: ProTAS and MiniROAD on CaptainCook4D). Large: raw predictions; \textcolor{gray}{small gray}: with a causal 2\,s mode filter. Bold: best per budget. P of the Fixed schedule is that of the models run at the 6\,Hz clock; the ProTAS specialists are the rows of Table~\ref{tab:supp_edge_t1}, at the frame rates and powers listed there, while the other specialists use our 6\,Hz recipe. At 20\,mW the Fixed schedule is the same for all models (G+M, one read every about 2.2\,s), but its rate depends on the model clock: 0.43\,FPS (period of 14 steps at 6\,Hz, 17.9\,mW) for MiniROAD and Transformer, 0.45\,FPS (67 steps at 30\,FPS, 18.7\,mW) for ProTAS. On CMU-MMAC and CaptainCook4D at 2.8\,W, all models use the same Fixed schedule, R+A at 6\,FPS: the most accurate Fixed schedules of Table~\ref{tab:supp_edge_t1} at this budget (30 and 15\,FPS) exceed the 6\,FPS maximum rate of the mixtures and of the budget rule. {Likewise, on Ego-Exo4D at 2.8\,W the ProTAS specialist is run at 6\,FPS (A+I+G+M, 0.27\,W) rather than at the 30\,FPS of Table~\ref{tab:supp_edge_t1} (1.34\,W), since the mixtures and the budget rule read at 6\,FPS at most.} Rule: reads per second of each sensor (R = RGB, A = Audio, I = IMU, G = Gaze, M = Monochrome), selected on validation. The rule is re-selected under the edge model on the validation videos; at 2.8\,W on Ego-Exo4D the budget never binds (largest schedule 735\,mW). CMU-MMAC: mean over five scenarios; CaptainCook4D: one scenario.}
\centering
\scriptsize
\setlength{\tabcolsep}{4pt}
\begin{tabular}[t]{@{}l rrr@{\hspace{0.3em}}}
\toprule
\multicolumn{4}{c}{\textbf{Ego-Exo4D}} \\
\midrule
Model & Acc & Edit & F1 \\
\midrule
\textit{2.8\,W} & \multicolumn{3}{r@{}}{\makebox[0pt][r]{\textit{Fixed: A+I+G+M 6\,FPS (P = 268\,mW)}}} \\
ProTAS spec. & 47.2\makebox[1.3em][l]{$_{\scriptscriptstyle \textcolor{gray}{45.8}}$} & \textbf{28.0}\makebox[1.3em][l]{$_{\scriptscriptstyle \textcolor{gray}{33.1}}$} & 23.9\makebox[1.3em][l]{$_{\scriptscriptstyle \textcolor{gray}{27.9}}$} \\
ProTAS mix. & 40.2\makebox[1.3em][l]{$_{\scriptscriptstyle \textcolor{gray}{39.3}}$} & 27.8\makebox[1.3em][l]{$_{\scriptscriptstyle \textcolor{gray}{29.4}}$} & 23.7\makebox[1.3em][l]{$_{\scriptscriptstyle \textcolor{gray}{24.6}}$} \\
ProTAS mix. + rule & 40.5\makebox[1.3em][l]{$_{\scriptscriptstyle \textcolor{gray}{39.4}}$} & 27.8\makebox[1.3em][l]{$_{\scriptscriptstyle \textcolor{gray}{28.9}}$} & 23.6\makebox[1.3em][l]{$_{\scriptscriptstyle \textcolor{gray}{24.0}}$} \\
\textit{\quad P = 735\,mW} & \multicolumn{3}{r@{}}{\makebox[0pt][r]{\textit{Rule: R$_{6}$A$_{6}$I$_{6}$G$_{6}$M$_{6}$}}} \\
\cmidrule(lr){1-4}
MiniROAD spec. & \textbf{56.2}\makebox[1.3em][l]{$_{\scriptscriptstyle \textcolor{gray}{54.4}}$} & 23.4\makebox[1.3em][l]{$_{\scriptscriptstyle \textcolor{gray}{36.4}}$} & \textbf{24.4}\makebox[1.3em][l]{$_{\scriptscriptstyle \textcolor{gray}{34.7}}$} \\
MiniROAD mix. & 55.2\makebox[1.3em][l]{$_{\scriptscriptstyle \textcolor{gray}{53.8}}$} & 20.8\makebox[1.3em][l]{$_{\scriptscriptstyle \textcolor{gray}{36.0}}$} & 21.2\makebox[1.3em][l]{$_{\scriptscriptstyle \textcolor{gray}{33.9}}$} \\
MiniROAD mix. + rule & 55.5\makebox[1.3em][l]{$_{\scriptscriptstyle \textcolor{gray}{54.2}}$} & 22.6\makebox[1.3em][l]{$_{\scriptscriptstyle \textcolor{gray}{36.2}}$} & 22.4\makebox[1.3em][l]{$_{\scriptscriptstyle \textcolor{gray}{34.9}}$} \\
\textit{\quad P = 735\,mW} & \multicolumn{3}{r@{}}{\makebox[0pt][r]{\textit{Rule: R$_{6}$A$_{6}$I$_{6}$G$_{6}$M$_{6}$}}} \\
\cmidrule(lr){1-4}
Transformer spec. & 52.1\makebox[1.3em][l]{$_{\scriptscriptstyle \textcolor{gray}{51.8}}$} & 10.8\makebox[1.3em][l]{$_{\scriptscriptstyle \textcolor{gray}{28.2}}$} & 9.7\makebox[1.3em][l]{$_{\scriptscriptstyle \textcolor{gray}{25.8}}$} \\
Transformer mix. & 48.8\makebox[1.3em][l]{$_{\scriptscriptstyle \textcolor{gray}{48.8}}$} & 11.1\makebox[1.3em][l]{$_{\scriptscriptstyle \textcolor{gray}{26.2}}$} & 9.7\makebox[1.3em][l]{$_{\scriptscriptstyle \textcolor{gray}{24.3}}$} \\
Transformer mix. + rule & 51.5\makebox[1.3em][l]{$_{\scriptscriptstyle \textcolor{gray}{50.8}}$} & 15.0\makebox[1.3em][l]{$_{\scriptscriptstyle \textcolor{gray}{28.2}}$} & 14.7\makebox[1.3em][l]{$_{\scriptscriptstyle \textcolor{gray}{27.3}}$} \\
\textit{\quad P = 177\,mW} & \multicolumn{3}{r@{}}{\makebox[0pt][r]{\textit{Rule: R$_{0.8}$A$_{6}$I$_{6}$G$_{3}$M$_{2}$}}} \\
\midrule
\textit{20\,mW} & \multicolumn{3}{r@{}}{\makebox[0pt][r]{\textit{Fixed: G+M 0.43\,FPS (P = 17.9\,mW)}}} \\
ProTAS spec. & 42.8\makebox[1.3em][l]{$_{\scriptscriptstyle \textcolor{gray}{41.8}}$} & 29.5\makebox[1.3em][l]{$_{\scriptscriptstyle \textcolor{gray}{29.7}}$} & 24.9\makebox[1.3em][l]{$_{\scriptscriptstyle \textcolor{gray}{23.6}}$} \\
ProTAS mix. & 38.3\makebox[1.3em][l]{$_{\scriptscriptstyle \textcolor{gray}{37.4}}$} & 27.2\makebox[1.3em][l]{$_{\scriptscriptstyle \textcolor{gray}{27.9}}$} & 20.3\makebox[1.3em][l]{$_{\scriptscriptstyle \textcolor{gray}{20.2}}$} \\
ProTAS mix. + rule & 39.0\makebox[1.3em][l]{$_{\scriptscriptstyle \textcolor{gray}{38.0}}$} & 28.2\makebox[1.3em][l]{$_{\scriptscriptstyle \textcolor{gray}{29.2}}$} & 22.6\makebox[1.3em][l]{$_{\scriptscriptstyle \textcolor{gray}{22.3}}$} \\
\textit{\quad P = 18.4\,mW} & \multicolumn{3}{r@{}}{\makebox[0pt][r]{\textit{Rule: I$_{6}$G$_{0.9}$M$_{0.2}$}}} \\
\cmidrule(lr){1-4}
MiniROAD spec. & 49.6\makebox[1.3em][l]{$_{\scriptscriptstyle \textcolor{gray}{48.1}}$} & 31.3\makebox[1.3em][l]{$_{\scriptscriptstyle \textcolor{gray}{33.6}}$} & 27.6\makebox[1.3em][l]{$_{\scriptscriptstyle \textcolor{gray}{27.6}}$} \\
MiniROAD mix. & 51.2\makebox[1.3em][l]{$_{\scriptscriptstyle \textcolor{gray}{49.4}}$} & \textbf{34.0}\makebox[1.3em][l]{$_{\scriptscriptstyle \textcolor{gray}{36.7}}$} & \textbf{30.8}\makebox[1.3em][l]{$_{\scriptscriptstyle \textcolor{gray}{30.5}}$} \\
MiniROAD mix. + rule & \textbf{51.7}\makebox[1.3em][l]{$_{\scriptscriptstyle \textcolor{gray}{50.0}}$} & 28.7\makebox[1.3em][l]{$_{\scriptscriptstyle \textcolor{gray}{35.8}}$} & 26.1\makebox[1.3em][l]{$_{\scriptscriptstyle \textcolor{gray}{30.0}}$} \\
\textit{\quad P = 18.4\,mW} & \multicolumn{3}{r@{}}{\makebox[0pt][r]{\textit{Rule: I$_{6}$G$_{0.9}$M$_{0.2}$}}} \\
\cmidrule(lr){1-4}
Transformer spec. & 48.5\makebox[1.3em][l]{$_{\scriptscriptstyle \textcolor{gray}{47.0}}$} & 28.7\makebox[1.3em][l]{$_{\scriptscriptstyle \textcolor{gray}{29.9}}$} & 26.1\makebox[1.3em][l]{$_{\scriptscriptstyle \textcolor{gray}{24.1}}$} \\
Transformer mix. & 46.9\makebox[1.3em][l]{$_{\scriptscriptstyle \textcolor{gray}{45.4}}$} & 29.4\makebox[1.3em][l]{$_{\scriptscriptstyle \textcolor{gray}{30.2}}$} & 26.4\makebox[1.3em][l]{$_{\scriptscriptstyle \textcolor{gray}{24.4}}$} \\
Transformer mix. + rule & 47.2\makebox[1.3em][l]{$_{\scriptscriptstyle \textcolor{gray}{45.8}}$} & 21.5\makebox[1.3em][l]{$_{\scriptscriptstyle \textcolor{gray}{30.1}}$} & 19.1\makebox[1.3em][l]{$_{\scriptscriptstyle \textcolor{gray}{24.0}}$} \\
\textit{\quad P = 18.8\,mW} & \multicolumn{3}{r@{}}{\makebox[0pt][r]{\textit{Rule: I$_{6}$G$_{0.6}$M$_{0.4}$}}} \\
\midrule
\textit{5\,mW} & \multicolumn{3}{r@{}}{\makebox[0pt][r]{\textit{Fixed: G+M 0.1\,FPS (P = 4.3\,mW)}}} \\
ProTAS spec. & 34.6\makebox[1.3em][l]{$_{\scriptscriptstyle \textcolor{gray}{34.0}}$} & 22.5\makebox[1.3em][l]{$_{\scriptscriptstyle \textcolor{gray}{22.5}}$} & 18.1\makebox[1.3em][l]{$_{\scriptscriptstyle \textcolor{gray}{17.6}}$} \\
ProTAS mix. & 34.9\makebox[1.3em][l]{$_{\scriptscriptstyle \textcolor{gray}{34.3}}$} & 27.8\makebox[1.3em][l]{$_{\scriptscriptstyle \textcolor{gray}{27.5}}$} & 18.4\makebox[1.3em][l]{$_{\scriptscriptstyle \textcolor{gray}{18.4}}$} \\
ProTAS mix. + rule & 37.7\makebox[1.3em][l]{$_{\scriptscriptstyle \textcolor{gray}{36.8}}$} & 28.2\makebox[1.3em][l]{$_{\scriptscriptstyle \textcolor{gray}{28.9}}$} & 21.9\makebox[1.3em][l]{$_{\scriptscriptstyle \textcolor{gray}{21.1}}$} \\
\textit{\quad P = 4.7\,mW} & \multicolumn{3}{r@{}}{\makebox[0pt][r]{\textit{Rule: I$_{6}$G$_{0.3}$}}} \\
\cmidrule(lr){1-4}
MiniROAD spec. & 40.5\makebox[1.3em][l]{$_{\scriptscriptstyle \textcolor{gray}{39.8}}$} & 33.4\makebox[1.3em][l]{$_{\scriptscriptstyle \textcolor{gray}{33.3}}$} & 25.4\makebox[1.3em][l]{$_{\scriptscriptstyle \textcolor{gray}{25.8}}$} \\
MiniROAD mix. & 41.8\makebox[1.3em][l]{$_{\scriptscriptstyle \textcolor{gray}{40.6}}$} & \textbf{36.7}\makebox[1.3em][l]{$_{\scriptscriptstyle \textcolor{gray}{36.5}}$} & 26.2\makebox[1.3em][l]{$_{\scriptscriptstyle \textcolor{gray}{26.0}}$} \\
MiniROAD mix. + rule & \textbf{48.7}\makebox[1.3em][l]{$_{\scriptscriptstyle \textcolor{gray}{46.8}}$} & 32.8\makebox[1.3em][l]{$_{\scriptscriptstyle \textcolor{gray}{36.8}}$} & \textbf{27.3}\makebox[1.3em][l]{$_{\scriptscriptstyle \textcolor{gray}{28.8}}$} \\
\textit{\quad P = 4.8\,mW} & \multicolumn{3}{r@{}}{\makebox[0pt][r]{\textit{Rule: I$_{6}$G$_{0.3}$}}} \\
\cmidrule(lr){1-4}
Transformer spec. & 38.7\makebox[1.3em][l]{$_{\scriptscriptstyle \textcolor{gray}{37.9}}$} & 31.8\makebox[1.3em][l]{$_{\scriptscriptstyle \textcolor{gray}{31.5}}$} & 24.9\makebox[1.3em][l]{$_{\scriptscriptstyle \textcolor{gray}{24.2}}$} \\
Transformer mix. & 40.7\makebox[1.3em][l]{$_{\scriptscriptstyle \textcolor{gray}{39.5}}$} & 33.2\makebox[1.3em][l]{$_{\scriptscriptstyle \textcolor{gray}{33.3}}$} & 26.9\makebox[1.3em][l]{$_{\scriptscriptstyle \textcolor{gray}{26.0}}$} \\
Transformer mix. + rule & 45.6\makebox[1.3em][l]{$_{\scriptscriptstyle \textcolor{gray}{44.3}}$} & 21.4\makebox[1.3em][l]{$_{\scriptscriptstyle \textcolor{gray}{30.1}}$} & 17.9\makebox[1.3em][l]{$_{\scriptscriptstyle \textcolor{gray}{24.3}}$} \\
\textit{\quad P = 4.8\,mW} & \multicolumn{3}{r@{}}{\makebox[0pt][r]{\textit{Rule: I$_{6}$G$_{0.3}$}}} \\
\bottomrule
\end{tabular}
\hfill
\begin{tabular}[t]{@{}l rrr@{\hspace{0.3em}}}
\toprule
\multicolumn{4}{c}{\textbf{CMU-MMAC}} \\
\midrule
Model & Acc & Edit & F1 \\
\midrule
\textit{2.8\,W} & \multicolumn{3}{r@{}}{\makebox[0pt][r]{\textit{Fixed: R+A 6\,FPS (P = 486\,mW)}}} \\
ProTAS spec. & 83.6\makebox[1.3em][l]{$_{\scriptscriptstyle \textcolor{gray}{81.9}}$} & 32.3\makebox[1.3em][l]{$_{\scriptscriptstyle \textcolor{gray}{46.6}}$} & 35.7\makebox[1.3em][l]{$_{\scriptscriptstyle \textcolor{gray}{43.7}}$} \\
ProTAS mix. & 82.4\makebox[1.3em][l]{$_{\scriptscriptstyle \textcolor{gray}{81.4}}$} & 38.3\makebox[1.3em][l]{$_{\scriptscriptstyle \textcolor{gray}{42.0}}$} & 40.0\makebox[1.3em][l]{$_{\scriptscriptstyle \textcolor{gray}{43.0}}$} \\
ProTAS mix. + rule & 80.9\makebox[1.3em][l]{$_{\scriptscriptstyle \textcolor{gray}{79.9}}$} & 35.1\makebox[1.3em][l]{$_{\scriptscriptstyle \textcolor{gray}{37.5}}$} & 36.8\makebox[1.3em][l]{$_{\scriptscriptstyle \textcolor{gray}{38.4}}$} \\
\textit{\quad P = 486\,mW} & \multicolumn{3}{r@{}}{\makebox[0pt][r]{\textit{Rule: R$_{6}$A$_{6}$I$_{6}$}}} \\
\cmidrule(lr){1-4}
MiniROAD spec. & \textbf{88.2}\makebox[1.3em][l]{$_{\scriptscriptstyle \textcolor{gray}{86.3}}$} & \textbf{41.5}\makebox[1.3em][l]{$_{\scriptscriptstyle \textcolor{gray}{58.3}}$} & \textbf{48.6}\makebox[1.3em][l]{$_{\scriptscriptstyle \textcolor{gray}{60.5}}$} \\
MiniROAD mix. & 86.8\makebox[1.3em][l]{$_{\scriptscriptstyle \textcolor{gray}{86.0}}$} & 28.0\makebox[1.3em][l]{$_{\scriptscriptstyle \textcolor{gray}{52.5}}$} & 35.9\makebox[1.3em][l]{$_{\scriptscriptstyle \textcolor{gray}{58.6}}$} \\
MiniROAD mix. + rule & 85.8\makebox[1.3em][l]{$_{\scriptscriptstyle \textcolor{gray}{84.8}}$} & 30.9\makebox[1.3em][l]{$_{\scriptscriptstyle \textcolor{gray}{50.5}}$} & 35.4\makebox[1.3em][l]{$_{\scriptscriptstyle \textcolor{gray}{55.5}}$} \\
\textit{\quad P = 486\,mW} & \multicolumn{3}{r@{}}{\makebox[0pt][r]{\textit{Rule: R$_{6}$A$_{6}$I$_{6}$}}} \\
\cmidrule(lr){1-4}
Transformer spec. & 86.2\makebox[1.3em][l]{$_{\scriptscriptstyle \textcolor{gray}{85.1}}$} & 18.1\makebox[1.3em][l]{$_{\scriptscriptstyle \textcolor{gray}{46.7}}$} & 22.5\makebox[1.3em][l]{$_{\scriptscriptstyle \textcolor{gray}{51.3}}$} \\
Transformer mix. & 84.0\makebox[1.3em][l]{$_{\scriptscriptstyle \textcolor{gray}{84.0}}$} & 13.7\makebox[1.3em][l]{$_{\scriptscriptstyle \textcolor{gray}{38.1}}$} & 15.0\makebox[1.3em][l]{$_{\scriptscriptstyle \textcolor{gray}{41.6}}$} \\
Transformer mix. + rule & 83.8\makebox[1.3em][l]{$_{\scriptscriptstyle \textcolor{gray}{83.5}}$} & 15.0\makebox[1.3em][l]{$_{\scriptscriptstyle \textcolor{gray}{39.1}}$} & 16.3\makebox[1.3em][l]{$_{\scriptscriptstyle \textcolor{gray}{42.2}}$} \\
\textit{\quad P = 486\,mW} & \multicolumn{3}{r@{}}{\makebox[0pt][r]{\textit{Rule: R$_{6}$A$_{6}$I$_{6}$}}} \\
\midrule
\textit{20\,mW} & \multicolumn{3}{r@{}}{\makebox[0pt][r]{\textit{Fixed: R+A 0.24\,FPS (P = 19.5\,mW)}}} \\
ProTAS spec. & 79.8\makebox[1.3em][l]{$_{\scriptscriptstyle \textcolor{gray}{78.3}}$} & 44.4\makebox[1.3em][l]{$_{\scriptscriptstyle \textcolor{gray}{44.4}}$} & 43.6\makebox[1.3em][l]{$_{\scriptscriptstyle \textcolor{gray}{43.0}}$} \\
ProTAS mix. & 79.2\makebox[1.3em][l]{$_{\scriptscriptstyle \textcolor{gray}{78.1}}$} & 36.5\makebox[1.3em][l]{$_{\scriptscriptstyle \textcolor{gray}{39.4}}$} & 36.6\makebox[1.3em][l]{$_{\scriptscriptstyle \textcolor{gray}{37.5}}$} \\
ProTAS mix. + rule & 78.0\makebox[1.3em][l]{$_{\scriptscriptstyle \textcolor{gray}{77.1}}$} & 38.2\makebox[1.3em][l]{$_{\scriptscriptstyle \textcolor{gray}{41.6}}$} & 35.0\makebox[1.3em][l]{$_{\scriptscriptstyle \textcolor{gray}{38.2}}$} \\
\textit{\quad P = 19.2\,mW} & \multicolumn{3}{r@{}}{\makebox[0pt][r]{\textit{Rule: R$_{0.2}$A$_{2}$I$_{6}$}}} \\
\cmidrule(lr){1-4}
MiniROAD spec. & 81.7\makebox[1.3em][l]{$_{\scriptscriptstyle \textcolor{gray}{80.4}}$} & \textbf{52.8}\makebox[1.3em][l]{$_{\scriptscriptstyle \textcolor{gray}{52.1}}$} & 51.8\makebox[1.3em][l]{$_{\scriptscriptstyle \textcolor{gray}{51.5}}$} \\
MiniROAD mix. & \textbf{83.2}\makebox[1.3em][l]{$_{\scriptscriptstyle \textcolor{gray}{81.5}}$} & 51.5\makebox[1.3em][l]{$_{\scriptscriptstyle \textcolor{gray}{52.2}}$} & \textbf{56.2}\makebox[1.3em][l]{$_{\scriptscriptstyle \textcolor{gray}{52.7}}$} \\
MiniROAD mix. + rule & 81.7\makebox[1.3em][l]{$_{\scriptscriptstyle \textcolor{gray}{80.0}}$} & 41.6\makebox[1.3em][l]{$_{\scriptscriptstyle \textcolor{gray}{51.4}}$} & 42.9\makebox[1.3em][l]{$_{\scriptscriptstyle \textcolor{gray}{46.5}}$} \\
\textit{\quad P = 19.2\,mW} & \multicolumn{3}{r@{}}{\makebox[0pt][r]{\textit{Rule: R$_{0.2}$A$_{2}$I$_{6}$}}} \\
\cmidrule(lr){1-4}
Transformer spec. & 81.3\makebox[1.3em][l]{$_{\scriptscriptstyle \textcolor{gray}{79.9}}$} & 46.5\makebox[1.3em][l]{$_{\scriptscriptstyle \textcolor{gray}{47.0}}$} & 48.3\makebox[1.3em][l]{$_{\scriptscriptstyle \textcolor{gray}{47.0}}$} \\
Transformer mix. & 81.8\makebox[1.3em][l]{$_{\scriptscriptstyle \textcolor{gray}{80.4}}$} & 49.1\makebox[1.3em][l]{$_{\scriptscriptstyle \textcolor{gray}{49.5}}$} & 52.9\makebox[1.3em][l]{$_{\scriptscriptstyle \textcolor{gray}{50.3}}$} \\
Transformer mix. + rule & 81.0\makebox[1.3em][l]{$_{\scriptscriptstyle \textcolor{gray}{79.5}}$} & 29.3\makebox[1.3em][l]{$_{\scriptscriptstyle \textcolor{gray}{46.0}}$} & 28.7\makebox[1.3em][l]{$_{\scriptscriptstyle \textcolor{gray}{39.7}}$} \\
\textit{\quad P = 19.3\,mW} & \multicolumn{3}{r@{}}{\makebox[0pt][r]{\textit{Rule: R$_{0.2}$A$_{2}$I$_{6}$}}} \\
\midrule
\textit{5\,mW} & \multicolumn{3}{r@{}}{\makebox[0pt][r]{\textit{Fixed: R+A+I 0.05\,FPS (P = 4.2\,mW)}}} \\
ProTAS spec. & 70.8\makebox[1.3em][l]{$_{\scriptscriptstyle \textcolor{gray}{70.2}}$} & 24.2\makebox[1.3em][l]{$_{\scriptscriptstyle \textcolor{gray}{24.2}}$} & 26.8\makebox[1.3em][l]{$_{\scriptscriptstyle \textcolor{gray}{26.8}}$} \\
ProTAS mix. & 70.0\makebox[1.3em][l]{$_{\scriptscriptstyle \textcolor{gray}{69.2}}$} & 37.0\makebox[1.3em][l]{$_{\scriptscriptstyle \textcolor{gray}{38.1}}$} & 26.7\makebox[1.3em][l]{$_{\scriptscriptstyle \textcolor{gray}{26.2}}$} \\
ProTAS mix. + rule & 72.2\makebox[1.3em][l]{$_{\scriptscriptstyle \textcolor{gray}{71.3}}$} & 37.3\makebox[1.3em][l]{$_{\scriptscriptstyle \textcolor{gray}{39.1}}$} & 27.8\makebox[1.3em][l]{$_{\scriptscriptstyle \textcolor{gray}{28.2}}$} \\
\textit{\quad P = 4.7\,mW} & \multicolumn{3}{r@{}}{\makebox[0pt][r]{\textit{Rule: R$_{0.06}$I$_{6}$}}} \\
\cmidrule(lr){1-4}
MiniROAD spec. & 70.2\makebox[1.3em][l]{$_{\scriptscriptstyle \textcolor{gray}{69.8}}$} & 27.6\makebox[1.3em][l]{$_{\scriptscriptstyle \textcolor{gray}{27.6}}$} & 27.3\makebox[1.3em][l]{$_{\scriptscriptstyle \textcolor{gray}{25.9}}$} \\
MiniROAD mix. & 71.3\makebox[1.3em][l]{$_{\scriptscriptstyle \textcolor{gray}{70.3}}$} & \textbf{40.3}\makebox[1.3em][l]{$_{\scriptscriptstyle \textcolor{gray}{40.9}}$} & 31.3\makebox[1.3em][l]{$_{\scriptscriptstyle \textcolor{gray}{31.6}}$} \\
MiniROAD mix. + rule & \textbf{74.9}\makebox[1.3em][l]{$_{\scriptscriptstyle \textcolor{gray}{73.9}}$} & 36.2\makebox[1.3em][l]{$_{\scriptscriptstyle \textcolor{gray}{41.1}}$} & \textbf{32.9}\makebox[1.3em][l]{$_{\scriptscriptstyle \textcolor{gray}{34.3}}$} \\
\textit{\quad P = 4.7\,mW} & \multicolumn{3}{r@{}}{\makebox[0pt][r]{\textit{Rule: R$_{0.06}$I$_{6}$}}} \\
\cmidrule(lr){1-4}
Transformer spec. & 72.1\makebox[1.3em][l]{$_{\scriptscriptstyle \textcolor{gray}{71.7}}$} & 26.6\makebox[1.3em][l]{$_{\scriptscriptstyle \textcolor{gray}{26.6}}$} & 30.2\makebox[1.3em][l]{$_{\scriptscriptstyle \textcolor{gray}{30.2}}$} \\
Transformer mix. & 70.1\makebox[1.3em][l]{$_{\scriptscriptstyle \textcolor{gray}{69.1}}$} & 39.8\makebox[1.3em][l]{$_{\scriptscriptstyle \textcolor{gray}{40.3}}$} & 30.0\makebox[1.3em][l]{$_{\scriptscriptstyle \textcolor{gray}{29.1}}$} \\
Transformer mix. + rule & 73.8\makebox[1.3em][l]{$_{\scriptscriptstyle \textcolor{gray}{72.7}}$} & 22.3\makebox[1.3em][l]{$_{\scriptscriptstyle \textcolor{gray}{34.1}}$} & 17.9\makebox[1.3em][l]{$_{\scriptscriptstyle \textcolor{gray}{27.4}}$} \\
\textit{\quad P = 4.7\,mW} & \multicolumn{3}{r@{}}{\makebox[0pt][r]{\textit{Rule: R$_{0.04}$A$_{0.3}$I$_{6}$}}} \\
\bottomrule
\end{tabular}
\hfill
\begin{tabular}[t]{@{}l rrr@{\hspace{0.3em}}}
\toprule
\multicolumn{4}{c}{\textbf{CaptainCook4D}} \\
\midrule
Model & Acc & Edit & F1 \\
\midrule
\textit{2.8\,W} & \multicolumn{3}{r@{}}{\makebox[0pt][r]{\textit{Fixed: R+A 6\,FPS (P = 476\,mW)}}} \\
ProTAS spec. & 44.2\makebox[1.3em][l]{$_{\scriptscriptstyle \textcolor{gray}{44.0}}$} & 20.9\makebox[1.3em][l]{$_{\scriptscriptstyle \textcolor{gray}{32.8}}$} & 13.9\makebox[1.3em][l]{$_{\scriptscriptstyle \textcolor{gray}{21.7}}$} \\
ProTAS mix. & 46.0\makebox[1.3em][l]{$_{\scriptscriptstyle \textcolor{gray}{46.8}}$} & 31.9\makebox[1.3em][l]{$_{\scriptscriptstyle \textcolor{gray}{36.1}}$} & 19.3\makebox[1.3em][l]{$_{\scriptscriptstyle \textcolor{gray}{22.4}}$} \\
ProTAS mix. + rule & 47.3\makebox[1.3em][l]{$_{\scriptscriptstyle \textcolor{gray}{47.4}}$} & \textbf{33.0}\makebox[1.3em][l]{$_{\scriptscriptstyle \textcolor{gray}{35.1}}$} & \textbf{20.6}\makebox[1.3em][l]{$_{\scriptscriptstyle \textcolor{gray}{22.8}}$} \\
\textit{\quad P = 476\,mW} & \multicolumn{3}{r@{}}{\makebox[0pt][r]{\textit{Rule: R$_{6}$A$_{6}$I$_{6}$}}} \\
\cmidrule(lr){1-4}
MiniROAD spec. & \textbf{55.4}\makebox[1.3em][l]{$_{\scriptscriptstyle \textcolor{gray}{55.7}}$} & 15.4\makebox[1.3em][l]{$_{\scriptscriptstyle \textcolor{gray}{26.4}}$} & 15.6\makebox[1.3em][l]{$_{\scriptscriptstyle \textcolor{gray}{26.5}}$} \\
MiniROAD mix. & 54.7\makebox[1.3em][l]{$_{\scriptscriptstyle \textcolor{gray}{54.8}}$} & 11.0\makebox[1.3em][l]{$_{\scriptscriptstyle \textcolor{gray}{24.4}}$} & 10.3\makebox[1.3em][l]{$_{\scriptscriptstyle \textcolor{gray}{22.9}}$} \\
MiniROAD mix. + rule & 53.9\makebox[1.3em][l]{$_{\scriptscriptstyle \textcolor{gray}{53.9}}$} & 11.3\makebox[1.3em][l]{$_{\scriptscriptstyle \textcolor{gray}{23.0}}$} & 10.8\makebox[1.3em][l]{$_{\scriptscriptstyle \textcolor{gray}{20.3}}$} \\
\textit{\quad P = 458\,mW} & \multicolumn{3}{r@{}}{\makebox[0pt][r]{\textit{Rule: R$_{6}$I$_{6}$}}} \\
\cmidrule(lr){1-4}
Transformer spec. & 52.4\makebox[1.3em][l]{$_{\scriptscriptstyle \textcolor{gray}{53.1}}$} & 3.3\makebox[1.3em][l]{$_{\scriptscriptstyle \textcolor{gray}{11.8}}$} & 2.6\makebox[1.3em][l]{$_{\scriptscriptstyle \textcolor{gray}{10.5}}$} \\
Transformer mix. & 51.7\makebox[1.3em][l]{$_{\scriptscriptstyle \textcolor{gray}{52.0}}$} & 3.0\makebox[1.3em][l]{$_{\scriptscriptstyle \textcolor{gray}{11.4}}$} & 2.4\makebox[1.3em][l]{$_{\scriptscriptstyle \textcolor{gray}{10.3}}$} \\
Transformer mix. + rule & 53.1\makebox[1.3em][l]{$_{\scriptscriptstyle \textcolor{gray}{53.9}}$} & 3.1\makebox[1.3em][l]{$_{\scriptscriptstyle \textcolor{gray}{11.3}}$} & 2.2\makebox[1.3em][l]{$_{\scriptscriptstyle \textcolor{gray}{9.7}}$} \\
\textit{\quad P = 476\,mW} & \multicolumn{3}{r@{}}{\makebox[0pt][r]{\textit{Rule: R$_{6}$A$_{6}$I$_{6}$}}} \\
\midrule
\textit{20\,mW} & \multicolumn{3}{r@{}}{\makebox[0pt][r]{\textit{Fixed: R+A 0.24\,FPS (P = 19.5\,mW)}}} \\
ProTAS spec. & 42.1\makebox[1.3em][l]{$_{\scriptscriptstyle \textcolor{gray}{42.0}}$} & 29.4\makebox[1.3em][l]{$_{\scriptscriptstyle \textcolor{gray}{29.4}}$} & 19.4\makebox[1.3em][l]{$_{\scriptscriptstyle \textcolor{gray}{19.4}}$} \\
ProTAS mix. & 45.4\makebox[1.3em][l]{$_{\scriptscriptstyle \textcolor{gray}{46.2}}$} & 33.2\makebox[1.3em][l]{$_{\scriptscriptstyle \textcolor{gray}{34.7}}$} & 22.5\makebox[1.3em][l]{$_{\scriptscriptstyle \textcolor{gray}{23.4}}$} \\
ProTAS mix. + rule & 46.9\makebox[1.3em][l]{$_{\scriptscriptstyle \textcolor{gray}{47.0}}$} & \textbf{34.3}\makebox[1.3em][l]{$_{\scriptscriptstyle \textcolor{gray}{34.7}}$} & 22.6\makebox[1.3em][l]{$_{\scriptscriptstyle \textcolor{gray}{24.0}}$} \\
\textit{\quad P = 18.9\,mW} & \multicolumn{3}{r@{}}{\makebox[0pt][r]{\textit{Rule: R$_{0.1}$A$_{3}$I$_{6}$}}} \\
\cmidrule(lr){1-4}
MiniROAD spec. & 51.7\makebox[1.3em][l]{$_{\scriptscriptstyle \textcolor{gray}{51.7}}$} & 26.2\makebox[1.3em][l]{$_{\scriptscriptstyle \textcolor{gray}{28.1}}$} & 23.5\makebox[1.3em][l]{$_{\scriptscriptstyle \textcolor{gray}{25.5}}$} \\
MiniROAD mix. & \textbf{52.8}\makebox[1.3em][l]{$_{\scriptscriptstyle \textcolor{gray}{52.8}}$} & 28.2\makebox[1.3em][l]{$_{\scriptscriptstyle \textcolor{gray}{30.1}}$} & \textbf{24.2}\makebox[1.3em][l]{$_{\scriptscriptstyle \textcolor{gray}{26.4}}$} \\
MiniROAD mix. + rule & 52.3\makebox[1.3em][l]{$_{\scriptscriptstyle \textcolor{gray}{52.2}}$} & 21.7\makebox[1.3em][l]{$_{\scriptscriptstyle \textcolor{gray}{28.7}}$} & 17.7\makebox[1.3em][l]{$_{\scriptscriptstyle \textcolor{gray}{23.0}}$} \\
\textit{\quad P = 18.8\,mW} & \multicolumn{3}{r@{}}{\makebox[0pt][r]{\textit{Rule: R$_{0.2}$A$_{2}$I$_{6}$}}} \\
\cmidrule(lr){1-4}
Transformer spec. & 50.8\makebox[1.3em][l]{$_{\scriptscriptstyle \textcolor{gray}{50.7}}$} & 22.7\makebox[1.3em][l]{$_{\scriptscriptstyle \textcolor{gray}{23.8}}$} & 18.8\makebox[1.3em][l]{$_{\scriptscriptstyle \textcolor{gray}{19.6}}$} \\
Transformer mix. & 50.1\makebox[1.3em][l]{$_{\scriptscriptstyle \textcolor{gray}{50.1}}$} & 21.6\makebox[1.3em][l]{$_{\scriptscriptstyle \textcolor{gray}{22.7}}$} & 18.2\makebox[1.3em][l]{$_{\scriptscriptstyle \textcolor{gray}{19.5}}$} \\
Transformer mix. + rule & 51.1\makebox[1.3em][l]{$_{\scriptscriptstyle \textcolor{gray}{51.1}}$} & 8.7\makebox[1.3em][l]{$_{\scriptscriptstyle \textcolor{gray}{17.3}}$} & 7.4\makebox[1.3em][l]{$_{\scriptscriptstyle \textcolor{gray}{15.0}}$} \\
\textit{\quad P = 18.8\,mW} & \multicolumn{3}{r@{}}{\makebox[0pt][r]{\textit{Rule: R$_{0.2}$A$_{2}$I$_{6}$}}} \\
\midrule
\textit{5\,mW} & \multicolumn{3}{r@{}}{\makebox[0pt][r]{\textit{Fixed: R+A 0.05\,FPS (P = 4.1\,mW)}}} \\
ProTAS spec. & 40.3\makebox[1.3em][l]{$_{\scriptscriptstyle \textcolor{gray}{40.8}}$} & 31.9\makebox[1.3em][l]{$_{\scriptscriptstyle \textcolor{gray}{32.8}}$} & 17.7\makebox[1.3em][l]{$_{\scriptscriptstyle \textcolor{gray}{18.1}}$} \\
ProTAS mix. & 45.0\makebox[1.3em][l]{$_{\scriptscriptstyle \textcolor{gray}{45.1}}$} & 36.4\makebox[1.3em][l]{$_{\scriptscriptstyle \textcolor{gray}{37.7}}$} & 23.7\makebox[1.3em][l]{$_{\scriptscriptstyle \textcolor{gray}{24.4}}$} \\
ProTAS mix. + rule & 44.4\makebox[1.3em][l]{$_{\scriptscriptstyle \textcolor{gray}{44.4}}$} & 35.5\makebox[1.3em][l]{$_{\scriptscriptstyle \textcolor{gray}{36.3}}$} & 23.7\makebox[1.3em][l]{$_{\scriptscriptstyle \textcolor{gray}{23.3}}$} \\
\textit{\quad P = 4.5\,mW} & \multicolumn{3}{r@{}}{\makebox[0pt][r]{\textit{Rule: R$_{0.04}$A$_{0.3}$I$_{6}$}}} \\
\cmidrule(lr){1-4}
MiniROAD spec. & 43.6\makebox[1.3em][l]{$_{\scriptscriptstyle \textcolor{gray}{43.7}}$} & 35.2\makebox[1.3em][l]{$_{\scriptscriptstyle \textcolor{gray}{35.5}}$} & 22.0\makebox[1.3em][l]{$_{\scriptscriptstyle \textcolor{gray}{22.6}}$} \\
MiniROAD mix. & \textbf{47.5}\makebox[1.3em][l]{$_{\scriptscriptstyle \textcolor{gray}{47.5}}$} & \textbf{40.6}\makebox[1.3em][l]{$_{\scriptscriptstyle \textcolor{gray}{42.5}}$} & \textbf{29.0}\makebox[1.3em][l]{$_{\scriptscriptstyle \textcolor{gray}{29.7}}$} \\
MiniROAD mix. + rule & 47.3\makebox[1.3em][l]{$_{\scriptscriptstyle \textcolor{gray}{47.8}}$} & 29.9\makebox[1.3em][l]{$_{\scriptscriptstyle \textcolor{gray}{35.5}}$} & 21.7\makebox[1.3em][l]{$_{\scriptscriptstyle \textcolor{gray}{25.9}}$} \\
\textit{\quad P = 4.5\,mW} & \multicolumn{3}{r@{}}{\makebox[0pt][r]{\textit{Rule: R$_{0.06}$I$_{6}$}}} \\
\cmidrule(lr){1-4}
Transformer spec. & 45.6\makebox[1.3em][l]{$_{\scriptscriptstyle \textcolor{gray}{46.1}}$} & 38.1\makebox[1.3em][l]{$_{\scriptscriptstyle \textcolor{gray}{38.2}}$} & 23.5\makebox[1.3em][l]{$_{\scriptscriptstyle \textcolor{gray}{23.8}}$} \\
Transformer mix. & 45.2\makebox[1.3em][l]{$_{\scriptscriptstyle \textcolor{gray}{45.1}}$} & 37.1\makebox[1.3em][l]{$_{\scriptscriptstyle \textcolor{gray}{37.9}}$} & 25.0\makebox[1.3em][l]{$_{\scriptscriptstyle \textcolor{gray}{25.1}}$} \\
Transformer mix. + rule & 46.7\makebox[1.3em][l]{$_{\scriptscriptstyle \textcolor{gray}{47.1}}$} & 14.8\makebox[1.3em][l]{$_{\scriptscriptstyle \textcolor{gray}{24.5}}$} & 10.4\makebox[1.3em][l]{$_{\scriptscriptstyle \textcolor{gray}{16.8}}$} \\
\textit{\quad P = 4.5\,mW} & \multicolumn{3}{r@{}}{\makebox[0pt][r]{\textit{Rule: R$_{0.04}$A$_{0.3}$I$_{6}$}}} \\
\bottomrule
\end{tabular}

\label{tab:supp_edge_t2}
\end{table*}

\begin{table*}[t]
\caption{\textbf{What is worth learning under the edge-SoC energy model} (same layout as Table~\ref{tab:learn_where}; Ego-Exo4D, MiniROAD mixture, whole test videos; P in mW). Bold: best per budget in each block. The findings of the main paper hold: (a) {at 5 and 20\,mW} no learned policy beats periodic reads with the same reads, and all spend more power than the budget rule; (b) with the same reads, only the non-causal oracle improves on periodic reads, and a causal policy imitating it does not. Budget rule and policy calibration are re-selected under the edge model on validation (cross-entropy checkpoints with the time-since input, HCMS-style without; true boundaries use the labels); P includes the policy network and can exceed the budget by up to 7\%. At 2.8\,W the budget never binds (largest schedule 735\,mW), so the rule and every timing row coincide and the learned policies saturate (logit shift $+32$ in 15/15 scenarios): AdaMML-style reduces to the IMU-only policy, and HCMS-style spends 621\,mW. 15 scenarios.}
\label{tab:supp_edge_t4}
\centering
\scriptsize
\setlength{\tabcolsep}{6pt}
\begin{tabular}{@{}l rrr rrr rrr@{}}
\toprule
 & \multicolumn{3}{c}{5\,mW} & \multicolumn{3}{c}{20\,mW} & \multicolumn{3}{c}{2.8\,W} \\
\cmidrule(lr){2-4}\cmidrule(lr){5-7}\cmidrule(lr){8-10}
 & Acc & F1 & P & Acc & F1 & P & Acc & F1 & P \\
\midrule
\multicolumn{10}{@{}l}{\textit{(a) Learned policies, same budget}} \\
Budget rule (periodic) & \textbf{48.7} & \textbf{27.3} & 4.75 & 51.7 & 26.1 & 18.4 & \textbf{55.5} & \textbf{22.4} & 735 \\
AdaMML-style & 40.0 & 21.2 & 5.01 & 46.4 & 22.9 & 20.4 & \textbf{55.5} & \textbf{22.4} & 735 \\
\quad periodic, same reads & 48.1 & 26.4 & 5.01 & 51.0 & 23.8 & 20.4 & \textbf{55.5} & \textbf{22.4} & 735 \\
HCMS-style & 40.9 & 22.4 & 5.20 & 49.1 & 24.4 & 21.4 & 54.2 & 21.3 & 621 \\
\quad periodic, same reads & 47.5 & 26.9 & 5.20 & \textbf{52.8} & 25.9 & 21.4 & 54.2 & 21.4 & 621 \\
Closed loop & 46.4 & 26.7 & 5.20 & 50.3 & 26.8 & 20.6 & \textbf{55.5} & 22.3 & 734 \\
\quad periodic, same reads & 46.6 & 25.8 & 5.20 & 51.7 & \textbf{27.9} & 20.6 & \textbf{55.5} & \textbf{22.4} & 734 \\
\midrule
\multicolumn{10}{@{}l}{\textit{(b) Read timing, same reads}} \\
Periodic & 48.7 & 27.3 & 4.75 & 51.7 & 26.1 & 18.4 & 55.5 & 22.4 & 735 \\
Random & 46.2 & 25.9 & 4.75 & 50.7 & 26.3 & 18.4 & 55.5 & 22.4 & 735 \\
At true action boundaries & 48.5 & 27.1 & 4.75 & 51.2 & 27.5 & 18.4 & 55.5 & 22.4 & 735 \\
Oracle (non-causal, optim.) & \textbf{54.7} & \textbf{33.1} & 4.75 & \textbf{54.6} & \textbf{30.0} & 18.4 & 55.5 & 22.4 & 735 \\
Learned, imitating the oracle (causal) & 47.5 & 26.9 & 4.75 & 50.2 & 27.1 & 18.4 & 55.5 & 22.4 & 735 \\
\bottomrule
\end{tabular}

\end{table*}

\FloatBarrier
\subsection{Energy-accuracy trade-off curves and qualitative examples}
\label{supp:energy_accuracy}
\begin{figure}[t]
    \centering
    \includegraphics[width=\linewidth]{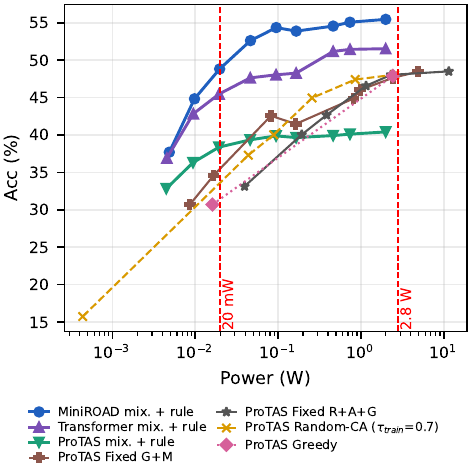}
    \caption{\textbf{Accuracy versus power on Ego-Exo4D, reference energy model.} MiniROAD, Transformer, and ProTAS mixtures + budget rule; ProTAS Fixed (G+M, R+A+G), Random-CA, and Greedy. The MiniROAD mixture + budget rule is above all other curves, which confirms the main findings.}
    \label{fig:tradeoff}
\end{figure}

\begin{figure}[t]
    \centering
    \includegraphics[width=\linewidth]{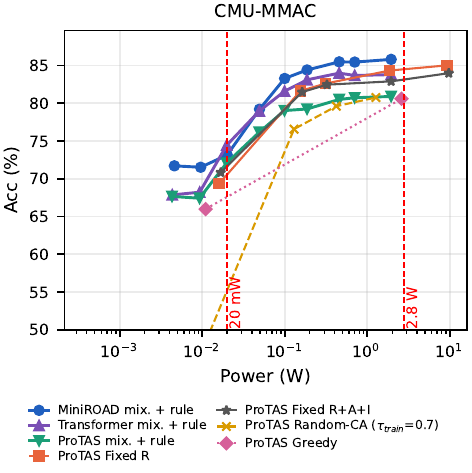}
    \caption{\textbf{Accuracy versus power on CMU-MMAC, reference energy model.} MiniROAD, Transformer, and ProTAS mixtures + budget rule; ProTAS Fixed (R, R+A+I), Random-CA, and Greedy. The MiniROAD mixture + budget rule is above all other curves except the Transformer mixture at 20\,mW, which confirms the main findings.}
    \label{fig:tradeoff_cmu}
\end{figure}

\begin{figure}[t]
    \centering
    \includegraphics[width=\linewidth]{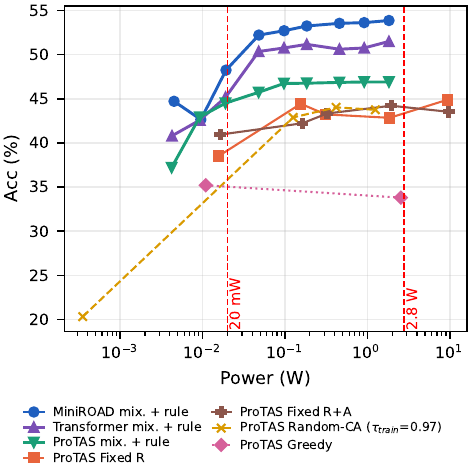}
    \caption{\textbf{Accuracy versus power on CaptainCook4D, reference energy model.} MiniROAD, Transformer, and ProTAS mixtures + budget rule; ProTAS Fixed (R, R+A), Random-CA, and Greedy. The MiniROAD mixture + budget rule is above all other curves except the ProTAS mixture at 10\,mW, which confirms the main findings.}
    \label{fig:tradeoff_capt}
\end{figure}

\begin{figure}[t]
    \centering
    \includegraphics[width=\linewidth]{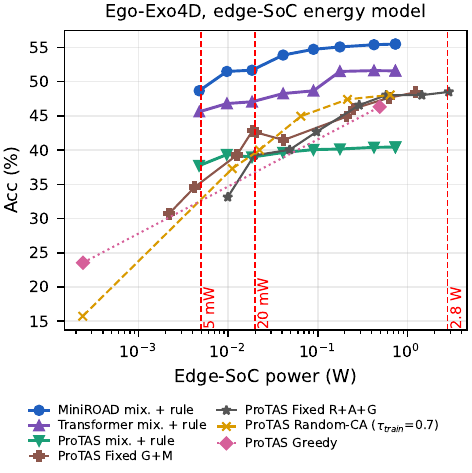}
    \caption{\textbf{Accuracy versus power on Ego-Exo4D, edge-SoC energy model.} MiniROAD, Transformer, and ProTAS mixtures + budget rule; ProTAS Fixed (G+M, R+A+G), Random-CA, and Greedy. The MiniROAD mixture + budget rule is above all other curves, which confirms the main findings.}
    \label{fig:tradeoff_edge}
\end{figure}

\begin{figure}[t]
    \centering
    \includegraphics[width=\linewidth]{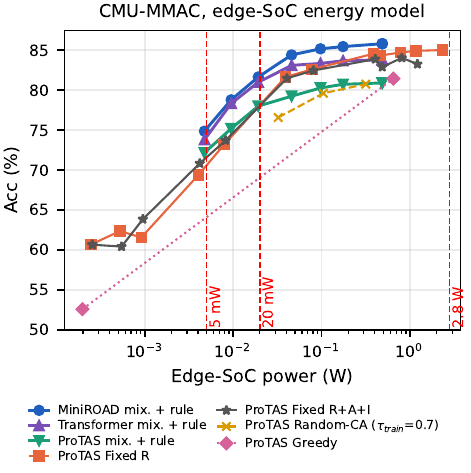}
    \caption{\textbf{Accuracy versus power on CMU-MMAC, edge-SoC energy model.} MiniROAD, Transformer, and ProTAS mixtures + budget rule; ProTAS Fixed (R, R+A+I), Random-CA, and Greedy. The MiniROAD mixture + budget rule is above all other curves, which confirms the main findings.}
    \label{fig:tradeoff_edge_cmu}
\end{figure}

\begin{figure}[t]
    \centering
    \includegraphics[width=\linewidth]{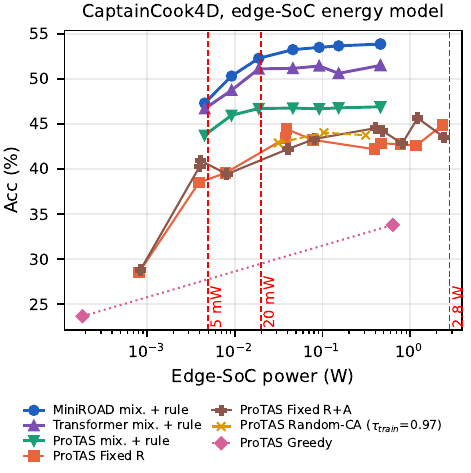}
    \caption{\textbf{Accuracy versus power on CaptainCook4D, edge-SoC energy model.} MiniROAD, Transformer, and ProTAS mixtures + budget rule; ProTAS Fixed (R, R+A), Random-CA, and Greedy. The MiniROAD mixture + budget rule is above all other curves, which confirms the main findings.}
    \label{fig:tradeoff_edge_capt}
\end{figure}

\begin{figure}[t]
    \centering
    \includegraphics[width=\linewidth]{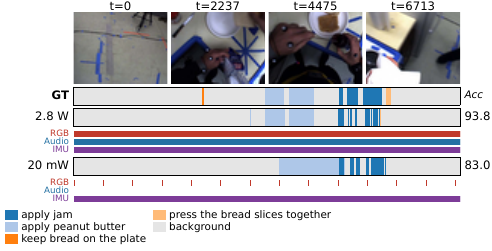}
    \caption{\textbf{Qualitative example on CMU-MMAC.} A test video (S12 Sandwich) processed by one MiniROAD model trained on the mixture, under the budget-rule schedules at 2.8\,W (RGB, audio, IMU at 6\,FPS) and 20\,mW (IMU at 6\,FPS, RGB every 17\,s). Accuracy is 93.8 at 2.8\,W and 83.0 at 20\,mW.}
    \label{fig:supp_qualitative_cmu}
\end{figure}

\raggedbottom
The trade-off curves plot accuracy against power over a range of budgets, from 5\,mW to 2.8\,W, and not only at the budgets of the main tables. They show whether the comparison between the mixtures with the budget rule and the heuristic sensing strategies depends on the budget, on the dataset, or on the energy model. \Cref{fig:tradeoff,fig:tradeoff_cmu,fig:tradeoff_capt} show Ego-Exo4D, CMU-MMAC, and CaptainCook4D under the reference energy model, and \cref{fig:tradeoff_edge,fig:tradeoff_edge_cmu,fig:tradeoff_edge_capt} show the same datasets under the edge-SoC energy model. \Cref{fig:supp_qualitative_cmu} then adds a qualitative example on CMU-MMAC to the Ego-Exo4D example of the main paper.

\begin{itemize}
    \item \emph{Mixture + budget rule} (MiniROAD, Transformer, and ProTAS mixtures): the budget fed to the budget rule (\cref{supp:budget_rule}), at 5, 10, 20, 50, 100, 200, 500, and 1000\,mW and 2.8\,W. For each budget, the rule selects one schedule on the validation videos of the MiniROAD mixture, and the three mixtures are tested with that schedule (inference only). Under the edge-SoC model, the rule is re-run with the edge-SoC costs. A point lies at the power that the schedule spends, which can be well below the budget: under the edge-SoC model the budget stops binding above 0.46--0.74\,W, so the last points coincide. Where a mixture's own choice of the rule differs from the MiniROAD one ({\eg ProTAS at 20\,mW on Ego-Exo4D,} Transformer, edge-SoC, 20\,mW and 2.8\,W), the point differs from the table.
    \item \emph{ProTAS Fixed}: the frame rate. Each point is a ProTAS specialist trained and tested on a Fixed schedule; each dataset has two sensor sets (G+M and R+A+G on Ego-Exo4D, R and R+A+I on CMU-MMAC, R and R+A on CaptainCook4D).
    \item \emph{ProTAS Random-CA}: the drop rate at inference, $\tau_{\text{inf}}$, for one model trained with the $\tau_{\text{train}}$ of Table~\ref{tab:all_protas} at 2.8\,W (0.7 on Ego-Exo4D and CMU-MMAC, 0.97 on CaptainCook4D; \cref{supp:random_policy}). Points below 0.2\,mW, where almost no sensor is read, are omitted.
    \item \emph{ProTAS Greedy}: the two runs of Table~\ref{tab:all_protas}, at the 20\,mW and 2.8\,W budgets.
    
\end{itemize}
The dashed red lines mark the budgets of the main tables: 20\,mW and 2.8\,W under the reference model, and 5\,mW, 20\,mW, and 2.8\,W under the edge-SoC model.

\paragraph{Findings.} On the six panels, the MiniROAD mixture with the budget rule is above the ProTAS Fixed, Random-CA, and Greedy curves at every power where it has a point; only operating points cheaper than its lowest budget (5\,mW) lie to its left. It is also the most accurate mixture at every budget, with two exceptions under the reference model: on CMU-MMAC at 20\,mW, the Transformer mixture is 1.3 points higher (74.5 vs.\ 73.2), and on CaptainCook4D at 10\,mW, the ProTAS mixture is 0.3 points higher (42.9 vs.\ 42.6). The advantage of the mixture with the budget rule is therefore not specific to the budgets of the main tables, to Ego-Exo4D, or to the reference energy model.
\clearpage

\FloatBarrier
\clearpage

\section{Complete Annotations and Dataset Tables}
\label{sec:supp_complete_tables}
This section reports the complete tables behind the benchmark construction of {\cref{supp:benchmark_construction}}: the label mappings of the re-annotation and the per-scenario and per-class statistics of the three datasets. The descriptions come first, and all tables follow at the end of the section.

\subsection{Re-annotation of Ego-Exo4D}
\label{supp:complete_reannot_ego}
The original Ego-Exo4D keysteps are too fine-grained and partly duplicated for the training videos of each scenario (\cref{supp:reannotation_egoexo}). \Crefrange{tab:new_label_mapping_clean_chain_617}{tab:new_label_mapping_making_chai_tea_617} give, for 14 scenarios, the Ego-METAS class to which each original keystep is mapped: the re-annotation groups 14 to 192 original keysteps per scenario into 6 to 18 classes (background included). Every Ego-METAS label can thus be traced back to the original annotation. First Aid -- CPR has no mapping table: its original annotations were already adequate, so it keeps its original labels.

\subsection{Re-annotation of CaptainCook4D}
\label{supp:complete_reannot_captain}
CaptainCook4D has the same problems, with background as the dominant class (\cref{supp:reannotation_captain}). \Cref{tab:class_mapping_captain} gives its mapping, which is shared by all recipes: the re-annotation groups 297 original step labels (background included) into 12 classes.

\subsection{Ego-Exo4D per-scenario statistics}
\label{supp:complete_stats_ego}
\Crefrange{tab:cls_stats_Covid-19_Rapid_Antigen_Test}{tab:cls_stats_First_Aid_-_CPR} report, for each of the 15 Ego-Exo4D scenarios, the number of instances and the duration (minimum, maximum, mean, standard deviation) of each class after re-annotation. Scenarios have 6 to 18 classes (12.4 on average). Background is the most frequent class in every scenario (40--54\% of the instances), but its instances are short (mean 1.2--6.9\,s per scenario).

\subsection{CMU-MMAC per-scenario statistics}
\label{supp:complete_stats_cmu}
\Cref{tab:cmu_stats} reports, for the five CMU-MMAC recipes, the videos and minutes of each split, the number of classes (5 to 10, 7.8 on average), and the video lengths (mean 178--499\,s per recipe); \crefrange{tab:classes_1}{tab:classes_5} give the per-class statistics. Compared with Ego-Exo4D, recipes have fewer classes and shorter videos. Background is again the most frequent class in every recipe (51--54\% of the instances), with longer instances (mean 18.6--29.4\,s per recipe).

\subsection{CaptainCook4D per-scenario statistics}
\label{supp:complete_stats_captain}
After removing the videos without audio or IMU, 78 videos of 21 recipes remain. \Cref{tab:split_comparison} counts them in the three official splits (recordings, person, environment): every official split leaves many recipes without validation or test videos (e.g., the recordings split has no validation video for 13 recipes and no test video for 5). We therefore define a new split and train on all recipes jointly (\cref{supp:dataset_stats}). \Cref{tab:captain_stats} reports this split per recipe, with the videos and minutes of each split, the number of classes (4 to 9), and the video lengths.

\clearpage
\onecolumn
{\small
\setlength{\LTleft}{\fill}
\setlength{\LTright}{\fill}
\begin{longtable}{|c|l|l|}
\caption{Label mapping for ``Clean and Lubricate the Chain''} \label{tab:new_label_mapping_clean_chain_617} \\
\hline
\textbf{New ID} & \textbf{Step Name} & \textbf{Old Step Name} \\ \hline
\endfirsthead

\multicolumn{3}{c}%
{{ \tablename\ \thetable{} -- continued from previous page}} \\
\hline
\textbf{New ID} & \textbf{Step Name} & \textbf{Old Step Name} \\ \hline
\endhead

\hline
\multicolumn{3}{|r|}{Continued on next page...} \\ \hline
\endfoot

\hline
\endlastfoot

0 & Background & Background \\ \hline

\multirow{7}{*}{1} & \multirow{7}{*}{Prepare Tools \& Setup} & Check the brake \\
 &  & Get the brush \\
 &  & Get the degreaser \\
 &  & Get the cleaning tool \\
 &  & Fill the cleaning tool with degreaser \\
 &  & Get a cloth \\
 &  & Clean the brush \\ \hline

\multirow{2}{*}{2} & \multirow{2}{*}{Apply Degreaser} & Add degreaser to the brush \\
 &  & Add degreaser to the chain \\ \hline

3 & Move Chain (Backpedal) & Slowly move the chain by backpedaling with hand \\ \hline

\multirow{2}{*}{4} & \multirow{2}{*}{Scrub Chain} & Hold the brush to the chain as you backpedal with hand \\
 &  & brush the chain while it remains stationary \\ \hline

5 & Dry Chain & Use a cloth to dry off any liquid on the chain \\ \hline

\multirow{2}{*}{6} & \multirow{2}{*}{Lubricate Chain} & Get the chain lube \\
 &  & Place a cloth underneath the chain to catch lube droppings \\ \hline
\end{longtable}}
{\small
\setlength{\LTleft}{\fill}
\setlength{\LTright}{\fill}
\begin{longtable}{|c|l|l|}
\caption{Label mapping for ``Install a Wheel''} \label{tab:new_label_mapping_install_wheel_617} \\
\hline
\textbf{New ID} & \textbf{Step Name} & \textbf{Old Step Name} \\ \hline
\endfirsthead

\multicolumn{3}{c}%
{{ \tablename\ \thetable{} -- continued from previous page}} \\
\hline
\textbf{New ID} & \textbf{Step Name} & \textbf{Old Step Name} \\ \hline
\endhead

\hline
\multicolumn{3}{|r|}{Continued on next page...} \\ \hline
\endfoot

\hline
\endlastfoot

\multirow{6}{*}{0} & \multirow{6}{*}{Background} & Put the bike on a bike repair stand \\
 &  & Put the bike upside down \\
 &  & Return wrenches \\
 &  & Loosen the axle nuts \\
 &  & Background \\
 &  & Deflate the tire \\ \hline

\multirow{4}{*}{1} & \multirow{4}{*}{Get Tools} & Get appropriate wrenches \\
 &  & Get bike tire lever \\
 &  & Get bike air pump \\
 &  & Get thru-axle \\ \hline

\multirow{5}{*}{2} & \multirow{5}{*}{Mount Wheel to Frame} & Get bike wheel \\
 &  & Put the wheel fully seated into the fork \\
 &  & Get the wheel \\
 &  & Clip the wheel to the hub and axle \\
 &  & Lift the wheel into the fork \\ \hline

\multirow{1}{*}{3} & Tighten Axle Nuts & Tighten both axle nuts using a wrench \\ \hline

\multirow{2}{*}{4} & \multirow{2}{*}{Secure Quick Release} & Push the release lever inwards \\
 &  & Push level inward and turn axle CW \\ \hline

\multirow{4}{*}{5} & \multirow{4}{*}{Secure Brakes} & Tighten brake cable to rear axle \\
 &  & Tighten the break pads \\
 &  & Connect the brake noodle \\
 &  & Clip the brake to the wheel \\ \hline

\multirow{1}{*}{6} & Adjust Derailleur & Pivot the derailleur back \\ \hline

\multirow{8}{*}{7} & \multirow{8}{*}{Check Function} & Roll the wheel \\
 &  & Rotate the wheel \\
 &  & Check the both hand brakes \\
 &  & Rotate the pedal \\
 &  & Squeeze the brake \\
 &  & Check the brake \\
 &  & Hold hand brake \\
 &  & Check the wheel \\ \hline
\end{longtable}}
{\small
\setlength{\LTleft}{\fill}
\setlength{\LTright}{\fill}
\begin{longtable}{|c|l|l|}
\caption{Label mapping for ``Remove a Wheel''} \label{tab:new_label_mapping_remove_wheel_617} \\
\hline
\textbf{New ID} & \textbf{Step Name} & \textbf{Old Step Name} \\ \hline
\endfirsthead

\multicolumn{3}{c}%
{{ \tablename\ \thetable{} -- continued from previous page}} \\
\hline
\textbf{New ID} & \textbf{Step Name} & \textbf{Old Step Name} \\ \hline
\endhead

\hline
\multicolumn{3}{|r|}{Continued on next page...} \\ \hline
\endfoot

\hline
\endlastfoot

\multirow{5}{*}{0} & \multirow{5}{*}{Background} & Shift the gear to the smallest cog \\
 &  & Put the bike on a bike repair stand \\
 &  & Put the bike upside down \\
 &  & Get bike air pump \\
 &  & Background \\ \hline

\multirow{1}{*}{1} & Get Tools & Get appropriate wrenches \\ \hline

\multirow{1}{*}{2} & Adjust Derailleur & Pivot the derailleur back \\ \hline

\multirow{5}{*}{3} & \multirow{5}{*}{Open Release} & Release caliper rim brake pads \\
 &  & Pull the release lever outwards \\
 &  & Loosen both axle nuts using a wrench \\
 &  & Loosen the axle nuts \\
 &  & Pull lever outward and turn axle CCW \\ \hline

\multirow{1}{*}{4} & Remove Wheel & Pull out the wheel from the frame \\ \hline

\multirow{1}{*}{5} & Deflate Tube & Deflate the tube \\ \hline
\end{longtable}}
{\small
\setlength{\LTleft}{\fill}
\setlength{\LTright}{\fill}
\begin{longtable}{|c|l|l|}
\caption{Label mapping for ``Fix a Flat Tire''} \label{tab:new_label_mapping_fix_tire_617} \\
\hline
\textbf{New ID} & \textbf{Step Name} & \textbf{Old Step Name} \\ \hline
\endfirsthead

\multicolumn{3}{c}%
{{ \tablename\ \thetable{} -- continued from previous page}} \\
\hline
\textbf{New ID} & \textbf{Step Name} & \textbf{Old Step Name} \\ \hline
\endhead

\hline
\multicolumn{3}{|r|}{Continued on next page...} \\ \hline
\endfoot

\hline
\endlastfoot

\multirow{2}{*}{0} & \multirow{2}{*}{Background} & Get bike tire lever \\
 &  & Background \\ \hline

\multirow{6}{*}{1} & \multirow{6}{*}{Remove Valve Hardware} & Remove dust cap from the wheel \\
 &  & Remove presta valve from the wheel \\
 &  & Remove stem nut from the wheel \\
 &  & Loosen lock nut at the valve stem \\
 &  & Remove the valve cap \\
 &  & Remove the dust cap \\
\hline
\end{longtable}}
{\small
\setlength{\LTleft}{\fill}
\setlength{\LTright}{\fill}
\begin{longtable}{|c|l|l|}
\caption{Label mapping for ``Covid-19 Rapid Antigen Test''} \label{tab:new_label_mapping_covid_617} \\
\hline
\textbf{New ID} & \textbf{Step Name} & \textbf{Old Step Name} \\ \hline
\endfirsthead

\multicolumn{3}{c}%
{{ \tablename\ \thetable{} -- continued from previous page}} \\
\hline
\textbf{New ID} & \textbf{Step Name} & \textbf{Old Step Name} \\ \hline
\endhead

\hline
\multicolumn{3}{|r|}{Continued on next page...} \\ \hline
\endfoot

\hline
\endlastfoot

\multirow{13}{*}{0} & \multirow{13}{*}{Background} & Set a timer \\
 &  & Check the expiration date \\
 &  & Blow nose \\
 &  & Get a timer \\
 &  & Wait for testing incubation time \\
 &  & Fold package box \\
 &  & Stop timer \\
 &  & Wait for 15 minutes \\
\hline
\end{longtable}}
{\small
\setlength{\LTleft}{\fill}
\setlength{\LTright}{\fill}
\begin{longtable}{|c|l|l|}
\caption{Label mapping for ``Cooking Noodles''} \label{tab:new_label_mapping_cooking_noodles_617} \\
\hline
\textbf{New ID} & \textbf{Step Name} & \textbf{Old Step Name} \\ \hline
\endfirsthead

\multicolumn{3}{c}%
{{ \tablename\ \thetable{} -- continued from previous page}} \\
\hline
\textbf{New ID} & \textbf{Step Name} & \textbf{Old Step Name} \\ \hline
\endhead

\hline
\multicolumn{3}{|r|}{Continued on next page...} \\ \hline
\endfoot

\hline
\endlastfoot

\multirow{4}{*}{0} & \multirow{4}{*}{Background} & Visually inspect the recipe \\
 &  & Check paper recipe \\
 &  & Set a timer \\
 &  & Background \\ \hline

\multirow{20}{*}{1} & \multirow{20}{*}{Get Kitchenware} & Get chopping board \\
 &  & Get chopsticks \\
 &  & Get cutting board \\
 &  & Get fork \\
\hline
\end{longtable}}
{\small
\setlength{\LTleft}{\fill}
\setlength{\LTright}{\fill}
\begin{longtable}{|c|l|l|}
\caption{Label mapping for ``Making Coffee latte''} \label{tab:new_label_mapping_making_coffee_617} \\
\hline
\textbf{New ID} & \textbf{Step Name} & \textbf{Old Step Name} \\ \hline
\endfirsthead

\multicolumn{3}{c}%
{{ \tablename\ \thetable{} -- continued from previous page}} \\
\hline
\textbf{New ID} & \textbf{Step Name} & \textbf{Old Step Name} \\ \hline
\endhead

\hline
\multicolumn{3}{|r|}{Continued on next page...} \\ \hline
\endfoot

\hline
\endlastfoot

\multirow{2}{*}{0} & \multirow{2}{*}{Background} & Cut ginger \\
 &  & Background \\ \hline

\multirow{14}{*}{1} & \multirow{14}{*}{Get Kitchenware} & Get chopping board \\
 &  & Get coffee filter \\
 &  & Get fork \\
 &  & Get kitchen towel \\
 &  & Get knife \\
 &  & Get measuring tool \\
\hline
\end{longtable}}
{\small
\setlength{\LTleft}{\fill}
\setlength{\LTright}{\fill}
\begin{longtable}{|c|l|l|}
\caption{Label mapping for ``Making Cucumber \& Tomato Salad''} \label{tab:new_label_mapping_making_cucumber_617} \\
\hline
\textbf{New ID} & \textbf{Step Name} & \textbf{Old Step Name} \\ \hline
\endfirsthead

\multicolumn{3}{c}%
{{ \tablename\ \thetable{} -- continued from previous page}} \\
\hline
\textbf{New ID} & \textbf{Step Name} & \textbf{Old Step Name} \\ \hline
\endhead

\hline
\multicolumn{3}{|r|}{Continued on next page...} \\ \hline
\endfoot

\hline
\endlastfoot

\multirow{2}{*}{0} & \multirow{2}{*}{Background} & Turn on the stove \\
 &  & Background \\ \hline

\multirow{11}{*}{1} & \multirow{11}{*}{Get Kitchenware} & Place the skillet/pan/pot on the stove \\
 &  & Get chopping board \\
 &  & Get cutting board \\
 &  & Get knife \\
 &  & Get peeler \\
 &  & Get plate \\
\hline
\end{longtable}}
{\small
\setlength{\LTleft}{\fill}
\setlength{\LTright}{\fill}
\begin{longtable}{|c|l|l|}
\caption{Label mapping for ``Cooking Scrambled Eggs''} \label{tab:new_label_mapping_cooking_eggs_617} \\
\hline
\textbf{New ID} & \textbf{Step Name} & \textbf{Old Step Name} \\ \hline
\endfirsthead

\multicolumn{3}{c}%
{{ \tablename\ \thetable{} -- continued from previous page}} \\
\hline
\textbf{New ID} & \textbf{Step Name} & \textbf{Old Step Name} \\ \hline
\endhead

\hline
\multicolumn{3}{|r|}{Continued on next page...} \\ \hline
\endfoot

\hline
\endlastfoot

\multirow{3}{*}{0} & \multirow{3}{*}{Background} & Remove the skillet or pan or pot from the stove \\
 &  & Check phone \\
 &  & Background \\ \hline

\multirow{23}{*}{1} & \multirow{23}{*}{Get Items} & Get black pepper \\
 &  & Get butter \\
 &  & Get cheese \\
 &  & Get cherry tomatoes \\
 &  & Get chili powder \\
\hline
\end{longtable}}
{\small
\setlength{\LTleft}{\fill}
\setlength{\LTright}{\fill}
\begin{longtable}{|c|l|l|}
\caption{Label mapping for ``Cooking an Omelet''} \label{tab:new_label_mapping_cooking_omelet_617} \\
\hline
\textbf{New ID} & \textbf{Step Name} & \textbf{Old Step Name} \\ \hline
\endfirsthead

\multicolumn{3}{c}%
{{ \tablename\ \thetable{} -- continued from previous page}} \\
\hline
\textbf{New ID} & \textbf{Step Name} & \textbf{Old Step Name} \\ \hline
\endhead

\hline
\multicolumn{3}{|r|}{Continued on next page...} \\ \hline
\endfoot

\hline
\endlastfoot

\multirow{6}{*}{0} & \multirow{6}{*}{Background} & Remove skillet/pan/pot from stove \\
 &  & Check paper recipe \\
 &  & Pour cooked egg mixture into a plate \\
 &  & Cook until the omelet is cooked \\
 &  & Background \\
 &  & Soak up the excess oil from the skillet. \\ \hline

\multirow{16}{*}{1} & \multirow{16}{*}{Get Kitchenware} & Get chopping board \\
 &  & Get chopsticks \\
\hline
\end{longtable}}
{\small
\setlength{\LTleft}{\fill}
\setlength{\LTright}{\fill}
\begin{longtable}{|c|l|l|}
\caption{Label mapping for ``Making Milk Tea''} \label{tab:new_label_mapping_making_milk_tea_617} \\
\hline
\textbf{New ID} & \textbf{Step Name} & \textbf{Old Step Name} \\ \hline
\endfirsthead

\multicolumn{3}{c}%
{{ \tablename\ \thetable{} -- continued from previous page}} \\
\hline
\textbf{New ID} & \textbf{Step Name} & \textbf{Old Step Name} \\ \hline
\endhead

\hline
\multicolumn{3}{|r|}{Continued on next page...} \\ \hline
\endfoot

\hline
\endlastfoot

\multirow{5}{*}{0} & \multirow{5}{*}{Background} & Cut tea bag \\
 & & Taste the recipe \\
 & & Measure the temperature \\
 & & Visually inspect the recipe \\
 & & Background \\ \hline

\multirow{23}{*}{1} & \multirow{23}{*}{Get Items} & Get black pepper \\
 & & Get coffee grounds \\
 & & Get dried herbs \\
\hline
\end{longtable}}
{\small
\setlength{\LTleft}{\fill}
\setlength{\LTright}{\fill}
\begin{longtable}{|c|l|l|}
\caption{Label mapping for ``Making Sesame-Ginger Asian Salad''} \label{tab:new_label_mapping_making_sesame_617} \\
\hline
\textbf{New ID} & \textbf{Step Name} & \textbf{Old Step Name} \\ \hline
\endfirsthead

\multicolumn{3}{c}%
{{ \tablename\ \thetable{} -- continued from previous page}} \\
\hline
\textbf{New ID} & \textbf{Step Name} & \textbf{Old Step Name} \\ \hline
\endhead

\hline
\multicolumn{3}{|r|}{Continued on next page...} \\ \hline
\endfoot

\hline
\endlastfoot

\multirow{7}{*}{0} & \multirow{7}{*}{Background} & Measure the temperature \\
 & & Check phone \\
 & & Check paper recipe \\
 & & Place the skillet/pan/pot on the stove \\
 & & Remove the skillet/pan/pot from the stove \\
 & & Mistake \\
 & & Background \\ \hline

\multirow{28}{*}{1} & \multirow{28}{*}{Get Items} & Get almonds \\
\hline
\end{longtable}}
{\small
\setlength{\LTleft}{\fill}
\setlength{\LTright}{\fill}
\begin{longtable}{|c|l|l|}
\caption{Label mapping for ``Cooking Tomato \& Eggs''} \label{tab:new_label_mapping_cooking_tomato_617} \\
\hline
\textbf{New ID} & \textbf{Step Name} & \textbf{Old Step Name} \\ \hline
\endfirsthead

\multicolumn{3}{c}%
{{ \tablename\ \thetable{} -- continued from previous page}} \\
\hline
\textbf{New ID} & \textbf{Step Name} & \textbf{Old Step Name} \\ \hline
\endhead

\hline
\multicolumn{3}{|r|}{Continued on next page...} \\ \hline
\endfoot

\hline
\endlastfoot

\multirow{3}{*}{0} & \multirow{3}{*}{Background} & Check paper recipe \\
 & & Visually inspect the recipe \\
 & & Mistake \\
\multirow{1}{*}{0} & \multirow{1}{*}{Background} & Background \\ \hline

\multirow{16}{*}{1} & \multirow{16}{*}{Get Kitchenware} & Get chopping board \\
 & & Get chopsticks \\
 & & Get cutting board \\
 & & Get fork \\
\hline
\end{longtable}}
{\small
\setlength{\LTleft}{\fill}
\setlength{\LTright}{\fill}
\begin{longtable}{|c|l|l|}
\caption{Label mapping for ``Making Chai Tea''} \label{tab:new_label_mapping_making_chai_tea_617} \\
\hline
\textbf{New ID} & \textbf{Step Name} & \textbf{Old Step Name} \\ \hline
\endfirsthead

\multicolumn{3}{c}%
{{ \tablename\ \thetable{} -- continued from previous page}} \\
\hline
\textbf{New ID} & \textbf{Step Name} & \textbf{Old Step Name} \\ \hline
\endhead

\hline
\multicolumn{3}{|r|}{Continued on next page...} \\ \hline
\endfoot

\hline
\endlastfoot

\multirow{5}{*}{0} & \multirow{5}{*}{Background} & Check paper recipe \\
 & & Set a timer \\
 & & Get a timer \\
 & & Check the heat \\
 & & Background \\ \hline

\multirow{16}{*}{1} & \multirow{16}{*}{Get Items} & Get cardamom \\
 & & Get dried herbs \\
 & & Get fresh thyme \\
\hline
\end{longtable}}

\clearpage
{\small
\setlength{\LTleft}{\fill}
\setlength{\LTright}{\fill}
\begin{longtable}{|c|l|l|}
\caption{Label mapping for CaptainCook4D \cite{peddi2024captaincook4d}} \label{tab:class_mapping_captain} \\
\hline
\textbf{New ID} & \textbf{Step Name} & \textbf{Old Step Name} \\ \hline
\endfirsthead

\multicolumn{3}{c}%
{{ \tablename\ \thetable{} -- continued from previous page}} \\
\hline
\textbf{New ID} & \textbf{Step Name} & \textbf{Old Step Name} \\ \hline
\endhead

\hline
\multicolumn{3}{|r|}{Continued on next page...} \\ \hline
\endfoot

\hline
\endlastfoot

\multirow{1}{*}{0} & Background & Background \\ \hline

\multirow{48}{*}{1} & \multirow{48}{*}{Cut Ingredients} & Continue slicing with floss \\
 & & Slice one tomato \\
 & & Peel 4 large garlic cloves \\
 & & Chop tomato roughly \\
 & & Mince the garlic \\
 & & Crack one egg in the bowl \\
 & & Chop 1/4 tomato \\
\hline
\end{longtable}}
\twocolumn

\newpage
\begin{table}[htbp]
\caption{Per-class stats for ``Covid-19 Rapid Antigen Test''}
\label{tab:cls_stats_Covid-19_Rapid_Antigen_Test}
\centering
\resizebox{\columnwidth}{!}{
\begin{tabular}{llccccc}
\toprule
Class ID & Class Name & Instances & Min (s) & Max (s) & Avg (s) & Std (s) \\
\midrule
0 & Background & 2680 & 0.03 & 163.20 & 2.22 & 7.68 \\
1 & Unbox package & 133 & 1.47 & 53.73 & 15.63 & 9.58 \\
2 & Unbox \& Arrange Materials & 303 & 0.70 & 95.93 & 11.93 & 14.57 \\
3 & Prepare Tube & 310 & 0.70 & 60.87 & 8.23 & 6.76 \\
4 & Fill Tube with Solution & 36 & 10.30 & 47.10 & 26.69 & 8.97 \\
5 & Prepare Swab & 192 & 1.60 & 44.30 & 12.91 & 7.55 \\
6 & Swab Nose (Insert/Extract) & 411 & 0.07 & 155.93 & 5.42 & 12.52 \\
7 & Swab Nose (Rotate) & 305 & 0.47 & 127.40 & 15.57 & 16.61 \\
8 & Mix Swab in Tube & 431 & 0.40 & 60.97 & 6.35 & 6.55 \\
9 & Prepare Test Cassette & 184 & 1.33 & 63.10 & 12.47 & 7.17 \\
10 & Apply Sample to Cassette & 186 & 1.00 & 67.17 & 14.84 & 8.62 \\
11 & Check Result & 109 & 1.03 & 65.57 & 7.27 & 9.43 \\
12 & Dispose Waste & 448 & 0.83 & 74.17 & 10.39 & 10.19 \\
13 & Read Instructions & 557 & 0.03 & 145.53 & 12.67 & 14.38 \\
\bottomrule
\end{tabular}
}

\end{table}

\begin{table}[htbp]
\caption{Per-class stats for ``Clean and Lubricate the Chain''}
\label{tab:cls_stats_Clean_and_Lubricate_the_Chain}
\centering
\resizebox{\columnwidth}{!}{
\begin{tabular}{llccccc}
\toprule
Class ID & Class Name & Instances & Min (s) & Max (s) & Avg (s) & Std (s) \\
\midrule
0 & Background & 211 & 0.03 & 43.70 & 3.04 & 6.50 \\
1 & Prepare Tools \& Setup & 56 & 0.93 & 41.77 & 6.95 & 8.94 \\
2 & Apply Degreaser & 45 & 0.10 & 45.00 & 7.55 & 10.54 \\
3 & Move Chain (Backpedal) & 56 & 0.53 & 109.33 & 9.87 & 17.96 \\
4 & Scrub Chain & 71 & 1.60 & 110.90 & 14.26 & 17.06 \\
5 & Dry Chain & 22 & 4.60 & 139.97 & 51.17 & 42.37 \\
6 & Lubricate Chain & 35 & 1.13 & 110.07 & 22.89 & 22.82 \\
\bottomrule
\end{tabular}
}

\end{table}

\begin{table}[htbp]
\caption{Per-class stats for ``Cooking Scrambled Eggs''}
\label{tab:cls_stats_Cooking_Scrambled_Eggs}
\centering
\resizebox{\columnwidth}{!}{
\begin{tabular}{llccccc}
\toprule
Class ID & Class Name & Instances & Min (s) & Max (s) & Avg (s) & Std (s) \\
\midrule
0 & Background & 694 & 0.03 & 20.97 & 1.55 & 3.05 \\
1 & Get Items & 188 & 0.70 & 24.20 & 5.10 & 4.12 \\
2 & Heat Skillet & 160 & 0.47 & 88.57 & 10.08 & 12.22 \\
3 & Prepare Ingredients (Cut/Peel) & 35 & 2.83 & 181.47 & 36.48 & 39.41 \\
4 & Whisk Eggs & 79 & 1.50 & 88.83 & 14.06 & 11.94 \\
5 & Fry Vegetables & 50 & 2.57 & 61.27 & 17.01 & 13.43 \\
6 & Scramble Eggs & 91 & 1.83 & 301.73 & 35.97 & 45.65 \\
7 & Season Food & 41 & 2.17 & 48.23 & 7.74 & 7.61 \\
8 & Serve Food & 24 & 4.67 & 42.67 & 19.40 & 9.25 \\
9 & Cleanup & 132 & 0.50 & 39.73 & 5.83 & 6.29 \\
\bottomrule
\end{tabular}
}

\end{table}

\begin{table}[htbp]
\caption{Per-class stats for ``Cooking Noodles''}
\label{tab:cls_stats_Cooking_Noodles}
\centering
\resizebox{\columnwidth}{!}{
\begin{tabular}{llccccc}
\toprule
Class ID & Class Name & Instances & Min (s) & Max (s) & Avg (s) & Std (s) \\
\midrule
0 & Background & 1554 & 0.03 & 239.13 & 3.47 & 10.16 \\
1 & Get Kitchenware & 192 & 0.37 & 68.70 & 5.90 & 7.22 \\
2 & Get Ingredients & 258 & 0.30 & 108.83 & 6.19 & 8.72 \\
3 & Prepare Ingredients & 147 & 1.63 & 250.93 & 39.39 & 38.48 \\
4 & Heat Water & 25 & 2.07 & 97.47 & 19.38 & 22.92 \\
5 & Boil Noodles & 208 & 1.00 & 138.83 & 15.43 & 17.22 \\
6 & Drain Noodles & 38 & 1.43 & 53.07 & 19.56 & 11.75 \\
7 & Prepare Skillet & 351 & 0.07 & 67.17 & 6.13 & 6.76 \\
8 & Stir-Fry Aromatics & 71 & 1.60 & 31.50 & 10.69 & 6.00 \\
9 & Stir-Fry Noodles & 265 & 0.03 & 155.13 & 17.71 & 19.51 \\
10 & Mix Sauce (Bowl) & 35 & 1.03 & 64.33 & 16.85 & 14.08 \\
11 & Season Food & 203 & 0.67 & 72.70 & 13.98 & 10.55 \\
12 & Serve Food & 38 & 4.23 & 76.30 & 21.37 & 14.42 \\
13 & Wash Items & 90 & 0.77 & 39.50 & 8.49 & 7.38 \\
14 & Put Away Items & 178 & 0.20 & 32.13 & 4.48 & 4.22 \\
\bottomrule
\end{tabular}
}

\end{table}

\begin{table}[htbp]
\caption{Per-class stats for ``Cooking an Omelet''}
\label{tab:cls_stats_Cooking_an_Omelet}
\centering
\resizebox{\columnwidth}{!}{
\begin{tabular}{llccccc}
\toprule
Class ID & Class Name & Instances & Min (s) & Max (s) & Avg (s) & Std (s) \\
\midrule
0 & Background & 1376 & 0.03 & 49.53 & 1.38 & 4.34 \\
1 & Get Kitchenware & 294 & 0.37 & 22.07 & 4.54 & 3.17 \\
2 & Get Ingredients & 312 & 0.20 & 34.50 & 4.05 & 4.48 \\
3 & Wash Ingredients & 15 & 1.80 & 39.40 & 9.86 & 9.66 \\
4 & Peel Ingredients & 33 & 1.43 & 44.93 & 13.33 & 10.94 \\
5 & Cut Ingredients & 116 & 1.73 & 115.30 & 17.21 & 17.56 \\
6 & Heat Skillet & 341 & 0.43 & 61.20 & 5.89 & 6.31 \\
7 & Whisk Eggs & 173 & 0.77 & 55.83 & 11.54 & 9.42 \\
8 & Saute Aromatics & 129 & 0.83 & 99.37 & 12.96 & 15.88 \\
9 & Pour \& Set & 83 & 1.53 & 33.43 & 10.36 & 6.54 \\
10 & Cook \& Shape & 50 & 2.20 & 105.87 & 17.82 & 17.57 \\
11 & Flip \& Fold & 96 & 0.70 & 82.07 & 15.99 & 15.67 \\
12 & Season Food & 125 & 0.87 & 38.93 & 6.82 & 4.27 \\
13 & Serve Food & 61 & 1.30 & 40.23 & 10.63 & 8.30 \\
14 & Cleanup & 213 & 0.40 & 112.93 & 5.03 & 9.51 \\
15 & Mistake & 2 & 7.83 & 12.33 & 10.08 & 2.25 \\
\bottomrule
\end{tabular}
}

\end{table}

\begin{table}[htbp]
\caption{Per-class stats for ``Cooking Tomato \& Eggs''}
\label{tab:cls_stats_Cooking_Tomato_and_Eggs}
\centering
\resizebox{\columnwidth}{!}{
\begin{tabular}{llccccc}
\toprule
Class ID & Class Name & Instances & Min (s) & Max (s) & Avg (s) & Std (s) \\
\midrule
0 & Background & 1797 & 0.03 & 126.73 & 2.49 & 6.72 \\
1 & Get Kitchenware & 266 & 0.10 & 23.70 & 4.44 & 3.48 \\
2 & Get Ingredients & 277 & 0.60 & 64.20 & 4.83 & 5.53 \\
3 & Wash Ingredients & 24 & 1.97 & 33.30 & 10.24 & 7.46 \\
4 & Peel Ingredients & 25 & 1.57 & 187.90 & 32.87 & 36.94 \\
5 & Cut Aromatics & 86 & 2.00 & 152.07 & 24.22 & 22.56 \\
6 & Cut Tomato & 68 & 2.00 & 162.00 & 34.44 & 37.32 \\
7 & Crack Eggs & 47 & 4.43 & 39.07 & 10.98 & 6.49 \\
8 & Whisk Eggs & 81 & 1.80 & 129.20 & 18.34 & 18.92 \\
9 & Heat Skillet & 459 & 0.50 & 57.53 & 6.02 & 7.22 \\
10 & Scramble Eggs & 110 & 2.63 & 91.47 & 19.50 & 17.86 \\
11 & Transfer Eggs (Out) & 16 & 5.60 & 46.93 & 18.22 & 11.88 \\
12 & Add \& Mush Tomatoes & 198 & 0.97 & 83.00 & 10.24 & 10.80 \\
13 & Stir-Fry Mixture & 224 & 0.53 & 184.23 & 19.39 & 20.15 \\
14 & Combine Eggs & 28 & 2.63 & 22.13 & 9.54 & 4.98 \\
15 & Season Food & 97 & 1.53 & 36.10 & 9.18 & 6.48 \\
16 & Serve Food & 50 & 4.70 & 123.50 & 20.48 & 18.24 \\
17 & Cleanup & 320 & 0.23 & 43.03 & 5.21 & 5.35 \\
\bottomrule
\end{tabular}
}

\end{table}

\begin{table}[htbp]
\caption{Per-class stats for ``Fix a Flat Tire - Replace a Bike Tube''}
\label{tab:cls_stats_Fix_a_Flat_Tire_-_Replace_a_Bike_Tube}
\centering
\resizebox{\columnwidth}{!}{
\begin{tabular}{llccccc}
\toprule
Class ID & Class Name & Instances & Min (s) & Max (s) & Avg (s) & Std (s) \\
\midrule
0 & Background & 894 & 0.03 & 163.57 & 2.39 & 7.15 \\
1 & Remove Valve Hardware & 86 & 0.80 & 26.53 & 8.09 & 4.33 \\
2 & Deflate Tube & 53 & 1.83 & 117.47 & 25.80 & 18.81 \\
3 & Detach Tire Bead & 106 & 0.53 & 101.03 & 15.80 & 16.84 \\
4 & Remove Old Tube & 96 & 1.10 & 80.03 & 15.59 & 13.72 \\
5 & Inspect Tire & 64 & 1.00 & 124.63 & 27.73 & 27.03 \\
6 & Prepare New Tube & 88 & 0.87 & 81.90 & 10.76 & 13.13 \\
7 & Install Inner Tube & 136 & 1.97 & 134.63 & 28.87 & 28.89 \\
8 & Mount Tire Bead & 98 & 5.60 & 184.13 & 30.89 & 26.28 \\
9 & Install Valve Hardware & 52 & 1.33 & 32.50 & 8.70 & 5.53 \\
10 & Inflate Tire & 166 & 3.93 & 139.13 & 25.38 & 24.10 \\
11 & Check Bead Seating & 59 & 2.23 & 75.77 & 19.92 & 17.89 \\
\bottomrule
\end{tabular}
}

\end{table}

\begin{table}[htbp]
\caption{Per-class stats for ``Install a Wheel''}
\label{tab:cls_stats_Install_a_Wheel}
\centering
\resizebox{\columnwidth}{!}{
\begin{tabular}{llccccc}
\toprule
Class ID & Class Name & Instances & Min (s) & Max (s) & Avg (s) & Std (s) \\
\midrule
0 & Background & 374 & 0.03 & 132.97 & 2.69 & 8.53 \\
1 & Get Tools & 37 & 0.97 & 89.60 & 8.17 & 18.45 \\
2 & Mount Wheel to Frame & 93 & 0.90 & 110.87 & 14.87 & 18.62 \\
3 & Tighten Axle Nuts & 106 & 1.13 & 106.73 & 25.40 & 22.67 \\
4 & Secure Quick Release & 32 & 1.63 & 38.57 & 16.49 & 9.98 \\
5 & Secure Brakes & 54 & 2.23 & 314.40 & 20.46 & 44.68 \\
6 & Adjust Derailleur & 14 & 2.97 & 18.63 & 7.50 & 4.60 \\
7 & Check Function (Spin/Brake) & 83 & 0.53 & 11.80 & 4.33 & 2.52 \\
\bottomrule
\end{tabular}
}

\end{table}

\begin{table}[htbp]
\caption{Per-class stats for ``Making Chai Tea''}
\label{tab:cls_stats_Making_Chai_Tea}
\centering
\resizebox{\columnwidth}{!}{
\begin{tabular}{llccccc}
\toprule
Class ID & Class Name & Instances & Min (s) & Max (s) & Avg (s) & Std (s) \\
\midrule
0 & Background & 317 & 0.03 & 86.73 & 3.64 & 9.60 \\
1 & Get Items & 81 & 0.87 & 32.57 & 5.45 & 4.84 \\
2 & Stove Operations & 70 & 0.77 & 28.50 & 7.04 & 6.46 \\
3 & Add Water \& Boil & 25 & 2.57 & 44.20 & 10.59 & 8.17 \\
4 & Add Spices & 22 & 1.83 & 40.97 & 11.68 & 9.98 \\
5 & Add Tea Leaves & 12 & 2.33 & 24.07 & 11.92 & 6.14 \\
6 & Simmering & 8 & 1.57 & 77.63 & 29.73 & 24.83 \\
7 & Add Milk & 14 & 3.90 & 39.00 & 21.07 & 11.66 \\
8 & Add Sugar & 8 & 8.70 & 54.00 & 24.33 & 13.53 \\
9 & Stir Mixture & 46 & 3.43 & 259.47 & 21.74 & 37.68 \\
10 & Strain \& Serve & 9 & 8.57 & 39.90 & 25.93 & 9.95 \\
11 & Wash Items & 11 & 3.50 & 19.53 & 10.18 & 4.22 \\
12 & Put Away Items & 51 & 1.20 & 16.33 & 4.65 & 3.03 \\
\bottomrule
\end{tabular}
}

\end{table}

\begin{table}[htbp]
\caption{Per-class stats for ``Making Coffee latte''}
\label{tab:cls_stats_Making_Coffee_latte}
\centering
\resizebox{\columnwidth}{!}{
\begin{tabular}{llccccc}
\toprule
Class ID & Class Name & Instances & Min (s) & Max (s) & Avg (s) & Std (s) \\
\midrule
0 & Background & 302 & 0.03 & 131.40 & 4.04 & 12.75 \\
1 & Get Kitchenware & 62 & 0.63 & 28.57 & 6.53 & 5.53 \\
2 & Get Ingredients & 30 & 1.17 & 25.37 & 5.96 & 5.13 \\
3 & Operate Heating Appliance & 53 & 0.67 & 31.20 & 8.68 & 6.62 \\
4 & Wait for Boiling & 26 & 3.50 & 91.37 & 26.07 & 18.25 \\
5 & Prepare Filter \& Grounds & 20 & 2.73 & 96.73 & 19.93 & 21.45 \\
6 & Pour Water for Brewing & 25 & 0.03 & 48.77 & 15.13 & 12.24 \\
7 & Process Milk (Heat/Stir/Froth) & 35 & 3.80 & 64.03 & 16.76 & 12.51 \\
8 & Pour Milk & 20 & 1.83 & 96.23 & 19.66 & 21.21 \\
9 & Extract \& Mix Coffee & 16 & 4.73 & 91.43 & 20.98 & 21.86 \\
10 & Pour Coffee to Serve & 6 & 8.57 & 42.37 & 20.05 & 12.38 \\
11 & Add Sweetener \& Spices & 26 & 3.17 & 20.07 & 9.92 & 5.23 \\
12 & Wash Items & 8 & 2.20 & 29.03 & 11.45 & 8.89 \\
13 & Put Away Items & 25 & 0.60 & 19.73 & 4.63 & 3.66 \\
14 & Wipe Clean & 9 & 1.33 & 29.67 & 7.85 & 9.11 \\
\bottomrule
\end{tabular}
}

\end{table}

\begin{table}[htbp]
\caption{Per-class stats for ``Making Cucumber \& Tomato Salad``}
\label{tab:cls_stats_Making_Cucumber_and_Tomato_Salad}
\centering
\resizebox{\columnwidth}{!}{
\begin{tabular}{llccccc}
\toprule
Class ID & Class Name & Instances & Min (s) & Max (s) & Avg (s) & Std (s) \\
\midrule
0 & Background & 801 & 0.03 & 33.27 & 1.24 & 3.81 \\
1 & Get Kitchenware & 121 & 0.47 & 27.97 & 4.43 & 4.16 \\
2 & Get Ingredients & 189 & 0.30 & 42.80 & 5.28 & 7.11 \\
3 & Wash Vegetables & 39 & 0.60 & 34.80 & 11.71 & 7.59 \\
4 & Peel Ingredients & 23 & 2.53 & 61.23 & 18.82 & 11.70 \\
5 & Cut Tomato & 97 & 1.40 & 89.77 & 16.62 & 18.05 \\
6 & Cut Cucumber & 86 & 0.03 & 87.43 & 24.34 & 22.57 \\
7 & Cut Other Ingredients & 25 & 1.97 & 61.00 & 14.55 & 15.28 \\
8 & Add Solid Ingredients & 131 & 0.47 & 36.67 & 6.93 & 7.56 \\
9 & Add Dressing \& Spices & 117 & 1.53 & 53.33 & 8.93 & 6.69 \\
10 & Mix Salad & 59 & 0.87 & 40.00 & 14.45 & 9.75 \\
11 & Serve Salad & 10 & 3.13 & 24.47 & 11.05 & 7.49 \\
12 & Wash Dishes & 22 & 3.20 & 65.63 & 11.68 & 12.90 \\
13 & Put Away Items & 72 & 0.33 & 30.87 & 5.07 & 6.74 \\
14 & Wipe \& Clean & 46 & 0.40 & 45.57 & 6.76 & 8.35 \\
\bottomrule
\end{tabular}
}

\end{table}

\begin{table}[htbp]
\caption{Per-class stats for ``Making Milk Tea''}
\label{tab:cls_stats_Making_Milk_Tea}
\centering

\resizebox{\columnwidth}{!}{
\begin{tabular}{llccccc}
\toprule
Class ID & Class Name & Instances & Min (s) & Max (s) & Avg (s) & Std (s) \\
\midrule
0 & Background & 736 & 0.03 & 171.13 & 3.90 & 12.12 \\
1 & Get Items & 249 & 0.57 & 104.83 & 7.19 & 9.09 \\
2 & Heat Water & 191 & 0.20 & 147.43 & 9.76 & 16.30 \\
3 & Brew Tea & 66 & 3.60 & 42.97 & 12.48 & 7.83 \\
4 & Steep \& Simmer & 53 & 1.20 & 151.03 & 20.93 & 25.73 \\
5 & Remove Tea & 10 & 1.17 & 15.67 & 7.05 & 3.66 \\
6 & Add Milk & 48 & 2.73 & 22.00 & 10.34 & 4.60 \\
7 & Add Sweetener \& Spices & 50 & 1.47 & 23.23 & 9.91 & 5.37 \\
8 & Stir Drink & 102 & 1.67 & 59.33 & 9.17 & 8.32 \\
9 & Pour to Serve & 29 & 2.27 & 41.17 & 16.66 & 9.30 \\
10 & Cleanup & 156 & 0.57 & 56.80 & 7.00 & 7.72 \\
11 & Mistake & 1 & 2.20 & 2.20 & 2.20 & 0.00 \\
\bottomrule
\end{tabular}
}

\end{table}

\begin{table}[htbp]
\caption{Per-class stats for ``Making Sesame-Ginger Asian Salad''}
\label{tab:cls_stats_Making_Sesame-Ginger_Asian_Salad}
\centering
\resizebox{\columnwidth}{!}{
\begin{tabular}{llccccc}
\toprule
Class ID & Class Name & Instances & Min (s) & Max (s) & Avg (s) & Std (s) \\
\midrule
0 & Background & 964 & 0.03 & 88.10 & 2.83 & 6.7 \\
1 & Get Items & 357 & 0.67 & 99.83 & 6.12 & 8.64 \\
2 & Wash Ingredients & 55 & 0.53 & 29.97 & 10.19 & 7.81 \\
3 & Peel Ingredients & 47 & 3.40 & 108.93 & 28.82 & 22.15 \\
4 & Grate Ingredients & 37 & 7.53 & 220.37 & 70.40 & 47.96 \\
5 & Cut Bell Peppers & 51 & 2.23 & 183.50 & 30.39 & 32.99 \\
6 & Cut Other Veg & 149 & 2.20 & 181.20 & 33.77 & 30.07 \\
7 & Add Solid Ingredients & 154 & 0.90 & 77.43 & 12.73 & 10.90 \\
8 & Make Dressing & 171 & 0.07 & 112.53 & 14.97 & 14.05 \\
9 & Whisk/Shake Dressing & 23 & 2.17 & 37.17 & 16.15 & 8.65 \\
10 & Pour Dressing & 22 & 3.07 & 50.63 & 13.22 & 9.55 \\
11 & Toss Salad & 50 & 2.13 & 61.00 & 18.88 & 10.80 \\
12 & Serve Salad & 16 & 1.53 & 44.77 & 15.28 & 12.05 \\
13 & Cleanup & 192 & 0.30 & 49.70 & 6.45 & 6.66 \\
\bottomrule
\end{tabular}
}

\end{table}

\begin{table}[htbp]
\caption{Per-class stats for ``Remove a Wheel''}
\label{tab:cls_stats_Remove_a_Wheel}
\centering
\resizebox{\columnwidth}{!}{
\begin{tabular}{llccccc}
\toprule
Class ID & Class Name & Instances & Min (s) & Max (s) & Avg (s) & Std (s) \\
\midrule
0 & Background & 241 & 0.03 & 98.07 & 6.85 & 13.00 \\
1 & Get Tools & 32 & 0.23 & 16.87 & 3.54 & 3.16 \\
2 & Adjust Derailleur & 14 & 3.37 & 38.57 & 9.47 & 8.72 \\
3 & Open Quick Release & 129 & 1.60 & 188.10 & 22.67 & 28.01 \\
4 & Remove Wheel & 24 & 1.70 & 21.03 & 6.83 & 4.73 \\
5 & Deflate Tube & 3 & 14.60 & 70.73 & 36.84 & 24.35 \\
\bottomrule
\end{tabular}
}

\end{table}

\begin{table}[htbp]
\caption{Per-class stats for ``First Aid - CPR''}
\label{tab:cls_stats_First_Aid_-_CPR}
\centering
\resizebox{\columnwidth}{!}{
\begin{tabular}{llccccc}
\toprule
Class ID & Class Name & Instances & Min (s) & Max (s) & Avg (s) & Std (s) \\
\midrule
0 & Background & 388 & 0.03 & 26.07 & 2.12 & 4.64 \\
1 & Have a conversation asking different qu & 24 & 0.53 & 16.13 & 3.34 & 4.08 \\
2 & Check if the patient is responding at a & 18 & 1.17 & 17.23 & 4.16 & 3.58 \\
3 & Tap patient to confirm consciousness & 75 & 0.83 & 12.70 & 3.82 & 2.35 \\
4 & Confirm patient consciousness from the & 98 & 0.73 & 13.07 & 3.82 & 2.13 \\
5 & Kneel on the side of the patient's neck & 20 & 1.27 & 16.07 & 4.77 & 3.30 \\
6 & Place the lower part of the palm (heel) & 45 & 0.17 & 43.47 & 1.88 & 6.33 \\
7 & Place the other hand on top of the firs & 65 & 0.17 & 8.97 & 1.38 & 1.39 \\
8 & Press hard at a rate of 100 to 120 comp & 109 & 0.23 & 57.40 & 18.39 & 8.98 \\
9 & Do artificial respirations two times af & 4 & 2.20 & 6.53 & 4.84 & 1.62 \\
10 & Call for help & 23 & 1.77 & 24.93 & 6.31 & 5.74 \\
\bottomrule
\end{tabular}
}

\end{table}

\clearpage
\begin{table*}[t]
\caption{{CMU-MMAC} Dataset Per-Scenario Statistics}
\centering

\resizebox{\textwidth}{!}{%
\begin{tabular}{lccccccccc}
& \multicolumn{3}{c}{N. Videos / duration (min)}& & &\\
\cmidrule(lr){2-4}
Scenario & Train & Val & Test & Classes & Total Time (h) & Avg Len (s) & Min Len (s) & Max Len (s) \\
\hline
Salad & 19/105.21 & 6/36.00 & 6/29.45 & 10 & 2.84 & 330.3 $\pm$ 79.8 & 129.6 & 527.6 \\
Sandwich & 20/60.08 & 4/10.06 & 3/9.94 & 5 & 1.33 & 177.9 $\pm$ 32.6 & 99.1 & 224.4 \\
Eggs & 19/97.65 & 3/16.53 & 7/38.21 & 9 & 2.54 & 315.3 $\pm$ 74.5 & 179.0 & 471.6 \\
Brownie & 15/95.61 & 5/37.11 & 11/77.41 & 9 & 3.50 & 406.7 $\pm$ 96.7 & 213.4 & 683.7 \\
Pizza & 20/167.29 & 5/34.97 & 5/47.17 & 6 & 4.16 & 498.9 $\pm$ 136.0 & 242.3 & 850.2 \\
\bottomrule
\end{tabular}
}%
\label{tab:cmu_stats}

\end{table*}

\begin{table}[h]
\caption{Per-Class Statistics for ``Salad''}
\centering

\label{tab:classes_1}
\resizebox{\columnwidth}{!}{%
\begin{tabular}{rlrrrrr}
\toprule
\textbf{ID} & \textbf{Class Name} & \textbf{Inst.} & \textbf{Min (s)} & \textbf{Max (s)} & \textbf{Avg (s)} & \textbf{Std (s)} \\
\midrule
0 & Background & 252 & 0.07 & 111.20 & 27.79 & 27.14 \\
1 & peel carrot & 32 & 2.83 & 34.43 & 15.51 & 7.12 \\
2 & cut carrot & 33 & 2.10 & 113.87 & 18.65 & 21.01 \\
3 & cut lettuce & 20 & 2.70 & 42.00 & 13.44 & 9.58 \\
4 & peel cucumber & 32 & 4.70 & 40.37 & 19.52 & 9.70 \\
5 & cut cucumber & 35 & 3.07 & 27.33 & 11.29 & 6.41 \\
6 & cut white onion & 33 & 2.87 & 43.03 & 17.49 & 10.29 \\
7 & add pepper & 15 & 2.23 & 10.93 & 6.30 & 2.41 \\
8 & add mayonnaise & 13 & 3.17 & 13.17 & 6.35 & 2.99 \\
9 & mix & 8 & 2.93 & 22.10 & 10.20 & 6.48 \\
\bottomrule
\end{tabular}%
}

\end{table}

\begin{table}[h]
\caption{Per-Class Statistics for ``Sandwich''}
\centering

\label{tab:classes_2}
\resizebox{\columnwidth}{!}{%
\begin{tabular}{rlrrrrr}
\toprule
\textbf{ID} & \textbf{Class Name} & \textbf{Inst.} & \textbf{Min (s)} & \textbf{Max (s)} & \textbf{Avg (s)} & \textbf{Std (s)} \\
\midrule
0 & Background & 194 & 0.10 & 89.30 & 18.58 & 19.30 \\
1 & keep bread on plate & 26 & 0.37 & 5.60 & 1.45 & 1.09 \\
2 & apply peanut butter & 62 & 1.53 & 28.00 & 8.72 & 6.16 \\
3 & apply jam & 52 & 0.70 & 43.80 & 10.13 & 8.94 \\
4 & press bread slices & 27 & 1.30 & 8.17 & 3.50 & 1.71 \\
\bottomrule
\end{tabular}%
}

\end{table}

\begin{table}[h]
\caption{Per-Class Statistics for ``Eggs''}
\centering

\label{tab:classes_3}
\resizebox{\columnwidth}{!}{%
\begin{tabular}{rlrrrrr}
\toprule
\textbf{ID} & \textbf{Class Name} & \textbf{Inst.} & \textbf{Min (s)} & \textbf{Max (s)} & \textbf{Avg (s)} & \textbf{Std (s)} \\
\midrule
0 & Background & 331 & 0.23 & 200.90 & 20.37 & 26.22 \\
1 & break egg & 47 & 2.67 & 12.73 & 5.25 & 1.72 \\
2 & mix & 70 & 1.13 & 51.57 & 12.64 & 10.06 \\
3 & add salt & 32 & 1.07 & 11.87 & 3.92 & 2.15 \\
4 & add pepper & 28 & 1.50 & 12.73 & 4.03 & 2.40 \\
5 & pour oil & 31 & 1.50 & 10.33 & 3.98 & 2.05 \\
6 & pour mixture & 29 & 3.57 & 28.33 & 10.13 & 4.82 \\
7 & flip & 38 & 1.47 & 33.47 & 11.10 & 6.97 \\
8 & plate omelette & 29 & 1.47 & 13.30 & 6.61 & 3.34 \\
\bottomrule
\end{tabular}%
}

\end{table}

\begin{table}[h]
\caption{Per-Class Statistics for ``Brownie''}
\centering

\label{tab:classes_4}
\resizebox{\columnwidth}{!}{%
\begin{tabular}{rlrrrrr}
\toprule
\textbf{ID} & \textbf{Class Name} & \textbf{Inst.} & \textbf{Min (s)} & \textbf{Max (s)} & \textbf{Avg (s)} & \textbf{Std (s)} \\
\midrule
0 & Background & 318 & 0.07 & 176.43 & 22.24 & 24.60 \\
1 & break egg & 45 & 3.43 & 27.83 & 9.51 & 5.18 \\
2 & mix eggs & 14 & 6.03 & 32.80 & 19.56 & 8.48 \\
3 & add water & 32 & 1.90 & 7.20 & 3.41 & 1.17 \\
4 & add oil & 33 & 2.07 & 10.10 & 4.60 & 1.91 \\
5 & add brownie mix & 45 & 2.47 & 33.63 & 13.15 & 7.68 \\
6 & mix contents & 67 & 0.80 & 133.03 & 36.96 & 28.80 \\
7 & spray oil & 28 & 3.57 & 12.13 & 6.77 & 2.05 \\
8 & pour mixture & 39 & 5.97 & 66.07 & 33.70 & 15.14 \\
\bottomrule
\end{tabular}%
}

\end{table}

\begin{table}[h]
\caption{Per-Class Statistics for ``Pizza''}
\centering
\label{tab:classes_5}
\resizebox{\columnwidth}{!}{%
\begin{tabular}{rlrrrrr}
\toprule
\textbf{ID} & \textbf{Class Name} & \textbf{Inst.} & \textbf{Min (s)} & \textbf{Max (s)} & \textbf{Avg (s)} & \textbf{Std (s)} \\
\midrule
0 & Background & 235 & 0.10 & 167.60 & 29.42 & 32.53 \\
1 & spread batter & 40 & 2.77 & 128.93 & 34.05 & 23.21 \\
2 & apply sauce & 36 & 1.73 & 106.23 & 38.44 & 21.64 \\
3 & add cheese & 50 & 11.57 & 214.13 & 53.76 & 40.18 \\
4 & add pepperoni & 49 & 0.70 & 228.13 & 51.61 & 48.61 \\
5 & to oven & 30 & 1.00 & 8.50 & 2.97 & 1.62 \\
\bottomrule
\end{tabular}%
}

\end{table}

\clearpage
\begin{table*}[t]
\caption{Captain Cook 4D: Videos per Scenario across Split Strategies}
\label{tab:split_comparison}
\centering

\resizebox{\textwidth}{!}{%
\begin{tabular}{lrrrrrrrrr}
\toprule
\multirow{2}{*}{\textbf{Scenario}} & \multicolumn{3}{c}{\textbf{Recordings}} & \multicolumn{3}{c}{\textbf{Person}} & \multicolumn{3}{c}{\textbf{Environment}} \\
\cmidrule(lr){2-4} \cmidrule(lr){5-7} \cmidrule(lr){8-10}
& Train & Val & Test & Train & Val & Test & Train & Val & Test \\
\midrule
Blender Banana Pancakes & 2 & 0 & 1 & 1 & 1 & 1 & 3 & 0 & 0 \\
Broccoli Stir Fry & 1 & 0 & 1 & 1 & 1 & 0 & 2 & 0 & 0 \\
Butter Corn Cup & 2 & 0 & 1 & 1 & 1 & 1 & 3 & 0 & 0 \\
Caprese Bruschetta & 3 & 0 & 2 & 4 & 0 & 1 & 2 & 0 & 3 \\
Cheese Pimiento & 3 & 1 & 0 & 2 & 0 & 2 & 2 & 2 & 0 \\
Coffee & 1 & 0 & 0 & 0 & 1 & 0 & 0 & 1 & 0 \\
Cucumber Raita & 5 & 1 & 3 & 3 & 3 & 3 & 5 & 4 & 0 \\
Dressed Up Meatballs & 1 & 1 & 2 & 3 & 1 & 0 & 3 & 0 & 1 \\
Herb Omelet with Fried Tomatoes & 0 & 1 & 2 & 2 & 0 & 1 & 3 & 0 & 0 \\
Microwave Egg Sandwich & 2 & 1 & 0 & 1 & 0 & 2 & 2 & 0 & 1 \\
Microwave Mug Pizza & 3 & 0 & 1 & 1 & 2 & 1 & 3 & 1 & 0 \\
Mug Cake & 3 & 0 & 2 & 4 & 0 & 1 & 1 & 0 & 4 \\
Pan Fried Tofu & 3 & 0 & 2 & 2 & 1 & 2 & 2 & 3 & 0 \\
Pinwheels & 1 & 0 & 0 & 0 & 1 & 0 & 0 & 1 & 0 \\
Ramen & 3 & 0 & 2 & 3 & 1 & 1 & 4 & 1 & 0 \\
Sauted Mushrooms & 2 & 0 & 1 & 2 & 1 & 0 & 3 & 0 & 0 \\
Scrambled Eggs & 1 & 1 & 2 & 2 & 1 & 1 & 1 & 0 & 3 \\
Spicy Tuna Avocado Wraps & 1 & 1 & 0 & 2 & 0 & 0 & 1 & 0 & 1 \\
Tomato Chutney & 1 & 0 & 1 & 2 & 0 & 0 & 1 & 1 & 0 \\
Tomato Mozzarella Salad & 4 & 2 & 1 & 4 & 0 & 3 & 5 & 2 & 0 \\
Zoodles & 2 & 0 & 1 & 2 & 0 & 1 & 1 & 0 & 2 \\
\midrule
\textbf{TOTAL} & \textbf{44} & \textbf{9} & \textbf{25} & \textbf{42} & \textbf{15} & \textbf{21} & \textbf{47} & \textbf{16} & \textbf{15} \\
\bottomrule
\end{tabular}
}%

\end{table*}

\begin{table*}[t]
\caption{Captain Cook 4D Dataset Per-Activity Statistics (Our Split)}
\label{tab:captain_stats}
\centering

\resizebox{\textwidth}{!}{%
\begin{tabular}{lccccccccc}
& \multicolumn{3}{c}{N. Videos / duration (min)}& & & &\\
\cmidrule(lr){2-4}
Activity & Train & Val & Test & Unique classes & Total (h) & Avg Len (s) & Min Len (s) & Max Len (s) \\
\hline
Pan Fried Tofu & 1/22.24 & 3/65.62 & 1/35.41 & 7 & 2.05 & 1479.4 & 1057.5 & 2124.8 \\
Ramen & 1/14.83 & 3/44.84 & 1/20.06 & 6 & 1.33 & 956.7 & 774.0 & 1203.4 \\
Cucumber Raita & 1/14.94 & 7/98.10 & 1/12.72 & 4 & 2.10 & 838.4 & 623.8 & 985.0 \\
Dressed Up Meatballs & 1/18.38 & 2/43.58 & 1/18.26 & 6 & 1.34 & 1203.3 & 1095.7 & 1324.1 \\
Microwave Mug Pizza & 1/9.70 & 2/20.98 & 1/15.90 & 6 & 0.78 & 698.5 & 581.8 & 953.8 \\
Scrambled Eggs & 1/32.15 & 2/50.41 & 1/41.10 & 5 & 2.06 & 1855.0 & 1289.2 & 2465.9 \\
\bottomrule
\end{tabular}
}%

\end{table*}

\end{bibunit}

\end{document}